\pdfoutput=1

\documentclass[11pt]{article}

\usepackage[preprint]{acl}

\usepackage{times}
\usepackage{latexsym}

\usepackage[T1]{fontenc}

\usepackage[utf8]{inputenc}

\usepackage{microtype}

\usepackage{inconsolata}

\usepackage{graphicx}

\usepackage{subcaption}
\usepackage{multirow}
\usepackage{dingbat}

\newcommand{\hochkomma}{$^{,}$}

\title{Language Models, Graph Searching, and Supervision Adulteration: 
\\ When More Supervision
is Less and How to Make More More}

\author{Arvid Frydenlund \\
  University of Toronto, Computer Science \\
  Vector Institute \\
  \texttt{arvie@cs.toronto.edu} 
  \\}

\usepackage[absolute,overlay]{textpos}

\begin{document}
\maketitle

\begin{textblock*}{18cm}(1.5cm,28.5cm) 
\centering
\footnotesize
   \href{https://aclanthology.org/TODO}{Proceedings of the 63nd Annual Meeting of the Association for Computational Linguistics (Volume 1: Long Papers), 
July 27–August 1st, 2025 ©2025 Association for Computational Linguistics}
\end{textblock*}

\begin{abstract}

This work concerns the path-star task, a minimal example of searching over a graph.  The graph, $G$, is star-shaped with $D$ arms radiating from a start node, $s$.
A language model (LM) is given $G$, $s$, and a target node $t$, which ends one of the arms and is tasked with generating the arm containing $t$. 
The minimal nature of this task means only a single choice needs to be made: which of the $D$ arms contains $t$?  

Decoder-only LMs fail to solve this 
elementary task above $1/D$ chance 
due to a learned shortcut that absorbs training supervision. We show how this pathology is caused by excess supervision and we present a series of solutions
demonstrating that the task is solvable via decoder-only LMs. {\bf We find that the task's minimal nature 
causes its difficulty, as it prevents task decomposition.} Our solutions provide
insight into the pathology and its implications for LMs trained via next-token prediction.   


\end{abstract}

\section{Introduction}

The path-star task is a seemingly simple, minimal graph search task intended to exhibit a flaw in the standard next-token prediction paradigm used to train 
decoder-only 
autoregressive LMs via teacher-forcing (TF) \citep{bachmann2024the}.

\begin{figure}
    \centering  
    \includegraphics[scale=0.68,trim={.65cm .65cm .65cm .65cm},clip]{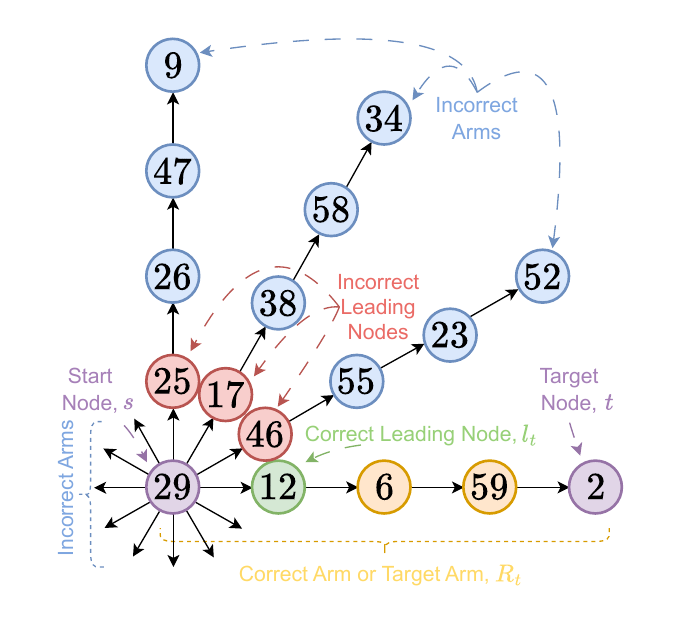}
    \vspace{-6pt} 
    \caption{An example path-star graph.  $D=12$, $M=5$, $s$ is `29', $t$ is `2', $R_t$ is `29 12 6 59 2', and $l_t$ is `12'.  We omit eight incorrect arms for space. 
    {\bf The task is to generate $R_t$ given a query, $Q=(s,\,t)$, and the graph, $G$, as a tokenized shuffled edge list} (See Fig.\@ \ref{fig:psg2-tokenized}).  
    }
    \label{fig:psg2}
\end{figure}

Each graph is star-shaped with $D$ arms rooted 
at a single start node, $s$.  The LM is given the complete graph (as a shuffled edge list) and a query, $(s,\, t)$, where $t$ is a target node that ends an arm.  The task is to generate the arm 
with $t$ 
from $s$ to $t$ 
(
Fig
\ref{fig:psg2}).  This requires the LM to choose an arm by initially generating one of the $D$ {\em leading} nodes adjacent to $s$, with the rest of the arm 
being dictated by following edges. Thus, the task's
difficulty lies in choosing the correct leading node, $l_t$, necessitating 
planning and reconstruction of the correct arm 
from $t$ to $l_t$.  

Training via TF conditions the LM on prior ground-truth tokens. 
This induces learning an undesired shortcut, 
the {\em Clever Hans Cheat} (CHC), which allows for trivial prediction of all non-leading nodes via a single edge look-up given the preceding node (given via TF). Thus, all the sequential supervision is absorbed into learning the CHC except for a single target token, $l_t$, which becomes the sole support for learning the required arm reconstruction subtask.\footnote{Why this itself is hard is an open question, see Sec.\@ \ref{sec:sensitivity}.}   
As a result, LMs fail to generate the correct arm above the random baseline of $1/D$ chance  \citep{bachmann2024the}.  While it has been shown that 
decoder-only LMs can 
express the task \citep{frydenlund-2024-mystery}, {\bf it remains an open question if decoder-only language models trained via teacher-forcing can learn the task}.

\subsection{Significance of the Failure}  

LMs are the ubiquitous 
model for NLP tasks \citep{NEURIPS2020_1457c0d6}, as well as for reasoning tasks \citep{bubeck2023sparks}.  These tasks often require planning, which LMs struggle with \citep[Ap.\@ \@\ref{appx:LLM-reasoning}]{valmeekam2023on, kambhampati2024position}. 

Potentially, this poor planning performance may be attributable to a fundamental problem with the next-token prediction paradigm. 
{\bf The path-star task is designed to support such a claim, where the minimal nature of the planning task is meant to isolate and highlight the failure; if standard LMs trained in standard ways fail to solve such a brutally simple task, it calls into question the sufficiency of the standard paradigm.}    

This motivated the use of alternative models. \citet{bachmann2024the} used a `teacher-less' model which foregoes TF by conditioning on fully masked input \citep{monea2023pass}. \citet{frydenlund-2024-mystery} generalized this to non- and iterative-autoregressive models and 
demonstrated learnability differences between models, where an encoder-only LM could 
solve the task (on small graphs). 

\citet{saparov2025transformers}  showed positive results on path-star graphs with encoder-only models and on more general graph topologies with both encoder- and decoder-only models. {\em They did not try path-star graphs with decoder-only LMs}.   
They found that the topology is critical to generalization, but that learning  
does not scale with graph size (and
using 
scratchpad that performs a depth-first search did not resolve this issue),
leading to the claim that `transformers struggle to learn to search'. 

\citet{yin-etal-2024-semformer} and \citet{hu2025learning} both 
proposed 
novel model architectures 
on the perceived deficiency of decoder-only models in solving the path-star task.
\citet{yin-etal-2024-semformer} trained an auxiliary autoencoder to form {\em planning latent-states} that encode future tokens and then trained an LM which regressed against these to learn special {\em planning tokens}.
\citet{hu2025learning} introduced a model trained on both forward and backward contexts using two separate forward and backward encoders. 
\citet{ahn2025efficient} continued with this motivation but tried to minimize the required architecture changes.

 \citet{wu2024can} put forth a related argument that next-token prediction is potentially problematic for planning tasks due to the cross-entropy loss leading to spurious correlations (see Appx.\@ \ref{appx:graph-learnability}).


\subsection{Disproving Prior Claim and Conjecture}

The claim that failure to learn the 
task demonstrates an insufficiency of the standard paradigm is empirically supported. Hence, showing the task to be learnable with standard methods would refute this claim.
\citet{bachmann2024the} also conjectured that this failure will apply to more complex planning tasks.  They gave story generation as an example, but without any empirical testing.   

{\bf This work will show that this claim and conjecture are unfounded, or at least extremely limited, by demonstrating that the task is fragile to minor modifications which make the task learnable via standard methods.}  We show that the CHC is not the sole cause of the task's difficulty, and that preventing it is unnecessary for learning the task.  We instead show that subtask decomposition is necessary for learning the task, and explain that the failure is caused by excessive or adulterated supervision that prevents this decomposition.

\citet{saparov2025transformers} showed strong success on the task with encoder-only models and on a similar-looking task with decoder-only models.  This success came without discussion with \citet{bachmann2024the}. We will explain this gap 
as also being attributable to subtle changes in the task that induce subtask decomposition.  Like \citet{saparov2025transformers}, we find the task becomes harder to learn with scale.  However, it is unnecessary to show scalability to larger graphs to refute prior claims.

\section{
Task, Data, and Tokenization}

\begin{figure*}
    \centering  
    \includegraphics[scale=0.272, trim={.65cm .15cm .65cm .65cm},clip]{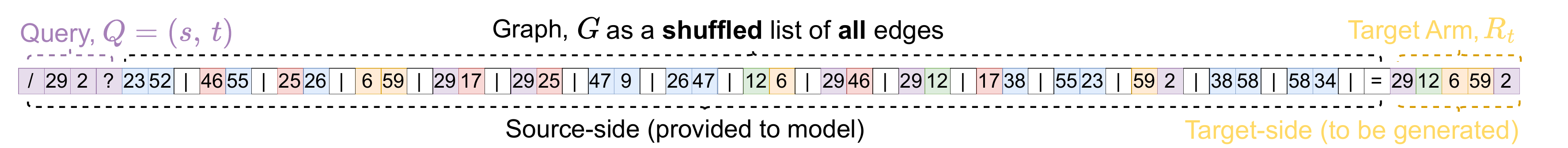}
    \vspace{-18pt}
    \caption{{\em A tokenization} corresponding to Fig.\@ \ref{fig:psg2}. We omit any edges belonging to the omitted incorrect arms.    
    }
    \label{fig:psg2-tokenized}
\end{figure*}

Each graph, $G$, has $D$ arms of the same length $M$ (inclusive of $s$) and is constructed by sampling nodes from a set of possible nodes, $V$, without replacement, making the graph size $|G| = D(M-1) + 1$.
The edges are determined by this sample order.  Thus, all nodes are unique and {\em semanticless} as they only relate via randomly sampled edges.  

The task is tokenized as a sequence consisting of the query, $Q = (s,\, t)$, with start- and end-of-query markers (`/ s t ?').  Each edge, $(u,\, v)$, is followed by the end-of-edge marker (`u v |').  See Fig.\@ \ref{fig:psg2-tokenized}. {\bf The entire graph is provided to the language model as a series of edges, which are randomly shuffled.}  This destroys any higher-order structural information about $G$, meaning that the task must be solved via planning and edge-following.   The source-side input into the model is $Q$ followed by $G$ and the end-of-graph marker (`=').  We place $Q$ before $G$ as it is better for decoder-only models \citep{frydenlund-2024-mystery}.  Let $R_t=x_1,\, \dots,\, x_M$ be the series of nodes 
from $s$ to $t$ forming the target arm and the sequential target-side supervision. 

Each experiment uses a model trained from scratch on graphs with static $D$ and $M$, so different-sized graphs are not mixed during training (except in Sec.\@ \ref{sec:multi-len}). We avoid uncontrollable biases from natural data by not using pretrained models.

\citet{frydenlund-2024-mystery} identified that the original experimental design leads to spurious correlations and overfitting due to the task's large sample space,  
\begin{equation}\label{eq:bi-sample-size}
    Z = \frac{|V|!}{(|V| - D(M-1) -1)!} \times D.  
\end{equation}
 To this end, they proposed using `structured samples'. While this helped, 
it did not resolve overfitting and increased training time. Instead, we use an online dataset that generates new samples during training. 
\citet{saparov2025transformers} also used an online dataset.  \citet{pmlr-v235-sanford24a} found better learnability with online training for the $k$-hop task.  We also minimize the space by using $|V| = |G|$.

Information can only be routed into the future due to the decoder's causal constraint. 
This increases the task's difficulty as the LM must learn two separate routing rules subject to edge $(u,\, v)$ proceeding or succeeding $(v,\, w)$.  
To avoid 
this, \citet{frydenlund-2024-mystery} introduced an `arm-wise shuffle' which 
only shuffles arms relative to each other. 
However, this 
allows for a trivial solution by looking back $M-1$ positions from $t$ to predict $l_t$.  Instead, we present a `causal-wise shuffle'  
where each arm is in sequential order but not contiguous. 
This alternative setup acts as a 
control to indicate if the causal constraint is causing 
difficulties.

\subsection{Supervision Adulteration}

We discussed how the models will overfit due to spurious correlations in the data.  The CHC is also a shortcut learnt due to overfitting; however, this is a different kind of overfitting, as it is not caused by the data, but rather by the way the task is constructed.  
In particular, the CHC is a shortcut caused by providing excess supervision or {\em adulteration}.   

Consider the various ways the task is supervised. In a supervised learning framework, we generally regard the target labels as `the supervision' as they can be human-annotated.  With the next-token prediction paradigm, we forego human annotation by using 
a self-supervised rule for generating the targets.  In both cases, 
the targets are a function of the input and thus the choice of input is just as much a form of supervision as the targets themselves.  Under this view, the model is provided with three types of supervision during training: the target-side labels, target-side inputs, and source-side inputs.

The path-star task (PST) is designed to induce a bad interaction between these three types of supervision under standard training.  For a given step $i$, the LM is trained to predict the target-side label $x_i \in R_t$.  However, it will be given $x_{<i}$, including $x_{i-1}$ as target-side input due to TF.  This induces learning a trivial single-edge lookup as the edge $(x_{i-1},\, x_i)$ is provided in the source-side input.\footnote{To foreshadow Sec.\@ \ref{sec:trees}, this would not be possible if that edge was not provided in the source-side inputs.}  Fig.\@ \ref{fig:alg1} illustrates the CHC as a single-edge lookup.

Thus 
excessive supervision partitions the sequential target-label supervision into supporting two tasks: the desired  PST, supported by a single target label, and the undesired single-edge lookup task, supported by the remaining labels.  This indicates that the task is not constructed properly to induce learning the PST. {\bf We will demonstrate that 1) this bad interaction, and thus the CHC, can be avoided in various ways 
and, 2) avoiding the CHC is not actually critical for learning the task if we consider other ways the task is supervised.}   

Task construction is a form of supervision encompassing multiple design decisions.  We considered the target-side above, however, the source-side representation is also supervised. For example, how the $G$ is shuffled matters (edge-wise vs.\@ arm-wise \citep{frydenlund-2024-mystery} or causal-wise as in Tbl.\@ \ref{tbl:baseline-results-tiny}), the decision to place $Q$ after or before $G$ or which tokens to include in the query (Sec.\@ \ref{sec:alt-queries}). 

There are also non-representational forms of supervision in creating the training data and training procedure.  \citet{bachmann2024the} consider training and evaluating each model of graphs of the exact same size.  This is done to 
explicitly dismiss out-of-domain effects.\footnote{``For each experiment, we generate the training and test graphs from the same distribution 
... with fixed [$D$], [$M$] and [$|V|$]. Thus, any failure we demonstrate is an in-distribution failure, and does not arise from the inability to generalize to different problem lengths''  \citep{bachmann2024the}.}
Alternatively, we can train the models on various sizes (Sec.\@ \ref{sec:multi-len}).  To elucidate why this is supervision, 
we could supervise the order of 
data to guide training from easy to hard via curriculum learning \citep{bengio2009curriculum}.  

Our choices of supervision to include $s$ and $t$ in $Q$ and only considering same-sized graphs leads to learning shortcuts for trivially predicting $s$ and $t$.  These are not bigram-based like the CHC, but positional as they are given on the source-side and always appear in the same place on the target-side.  


\subsection{Sensitivity Conjecture}\label{sec:sensitivity}

Why learning $l_t$ from a single target token is difficult is an open question.
\citet{frydenlund-2024-mystery} conjectured it relates to the task being sensitive to a single token, $t$.  \citet{hu2025learning} provided a construction of parity as a 
path-star task that only generates $l_t$, implying it is at least as hard to solve as parity. 
Parity is maximally sensitive and known to be extremely difficult to learn with transformers \citep{bhattamishra-etal-2023-simplicity, hahn2024sensitive}. 
This conjecture motivates some methodology, however, we find little empirical support for it (Sec.\@ \ref{sec:alt-queries}).

\section{Methods and Experiments}

We use decoder-only models with 2 heads, 64 dim.\@ embeddings, 256 dim.\@ feed-forward layers, and learned positional embeddings.  We use $L=8$ layers for most experiments.  Having $M < L$ allows for the linear graph reconstruction alg.\@ to be learnt.
\citet{frydenlund-2024-mystery} proved $\mathcal{O}(\log(M))$ layers are sufficient for this in theory. 
This was empirically demonstrated by \citet{yin-etal-2024-semformer, saparov2025transformers} who use $L < M$. 
\citet{pmlr-v235-sanford24a} demonstrated that this $\mathcal{O}(\log(M))$ alg.\@ can be learnt for a related task.
We use a learning-rate of 
$5 \times 10^{-4}$, 
a batch-size of 1024, and do not use dropout  
or a scheduler.
We train for 
100M samples. 

We 
use  
$D \in \{2, 3, 4, 5\}$, $M \in \{5, 7, 9, 12, 15\}$ but only up until we observe unsuccessful trials. 

\begin{figure}
    \centering  
    \includegraphics[scale=0.48, trim={.55cm .6cm .55cm .6cm},clip]{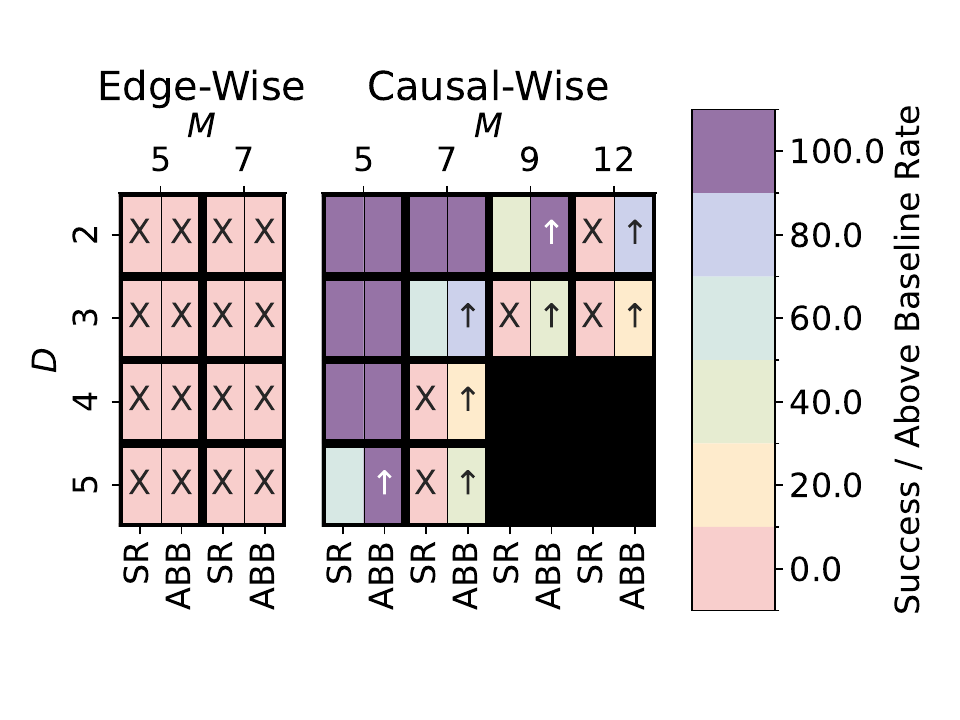}
    \vspace{-16pt}
\caption{Baseline results.  We report the {\em Success Rate} (SR) where the model predicts $> 95$\% sequential accuracy 
over $n=5$ seeded trials and {\em Above-Baseline} (ABB) where the model predicts $> (100/D +10)$\% sequential accuracy.  This happens when the model can predict $l_t$ above $1/D$ chance.  As such, when ABB $>$ SR ($\uparrow$), it implies that the model has overcome the main challenge of the PST and would have learnt the task had it been provided with more training time in these cases.  An `x' further indicates no trials learnt the task. 
} 
\label{tbl:baseline-results-tiny}
\end{figure}

We modify the task setting from \citet{bachmann2024the} by a) placing $Q$ before $G$, b) using an online dataset to avoid overfitting, and c) setting $|V| = |G|$ instead of 100.  {\bf These changes are immaterial to any conjecture regarding an inability to plan and do not prevent learning the CHC or its apparent effect on the task.}  We confirm the PST is still unlearnable (even on minimal graphs with only $|G|=9$ nodes) in this setting in Fig.\@ \ref{tbl:baseline-results-tiny} (full results are in Tbl.\@ \ref{tbl:baseline-results} in Appx.\@ \ref{appx:baseline-results}).

Fig.\@ \ref{tbl:baseline-results-tiny} shows the PST is learnable with causal-wise shuffling, {\bf indicating that the causal constraint accounts for some of the task's difficulty.}  All further experiments use `edge-wise' shuffling.

{\bf In the following sections, we will introduce a method or a slight modification to the task which allows the task to be learnt.  In general, these will use standard teacher-forcing and next-token prediction.  They will also be orthogonal to each other.  Importantly, the success of each method can be explained as avoiding supervision adulteration and so inducing subtask decomposition} (one example of this is illustrated in Fig. \ref{fig:psg2-algs}).

\subsection{Token Masking}

We first consider token masking to address the adulteration.  This will discourage learning the CHC by preventing conditioning on fully observed prior ground-truths during training, thus breaking the bad supervision interaction by modifying the target-side inputs.  This is motivated by the limited successes of `teacher-less',  iterative-, and non-autoregressive models \citep{frydenlund-2024-mystery}.

A main innovation from these models is that we do not need to employ full masking, unlike the `teacher-less' and non-autoregressive models and we can keep the causal parameterization of the model, unlike the iterative- and non-autoregressive models.  
Importantly, this can be achieved via ubiquitous data-noising methods used with standard TFed training.  In particular, we can employ either token dropout/masking \citep{NIPS2016_076a0c97, bowman-etal-2016-generating} or token replacements via scheduled sampling\footnote{As we know how the model generates, we skip implementing scheduled sampling and just sample from $V$ instead.} \citep{bengio2015scheduled} (or a mix). 
Replacement has the benefit of providing an anti-CHC learning signal as the model can not trust edge look-ups, but it introduces more complex noise. We try both uniform sampling of the sequence length and contiguous span sampling to shun consecutive ground-truths \citep{joshi-etal-2020-spanbert}.



\subsubsection{Results and Discussion}


\begin{figure}
    \centering  
    \includegraphics[scale=0.48, trim={.50cm .50cm .50cm .50cm},clip]{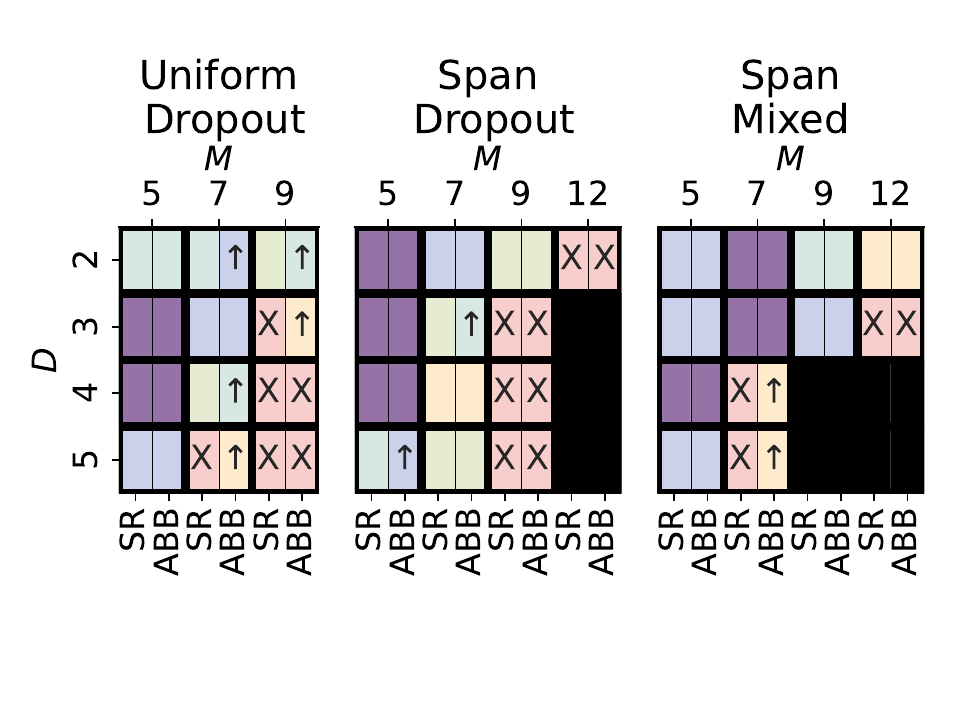}
    \vspace{-10pt}
\caption{Masking results (full Tbl.\@ \ref{tbl:masking-results} in Appx.\@ \ref{appx:masking-results}).}
\label{tbl:masking-results-tiny}
\end{figure}

Fig.\@ \ref{tbl:masking-results-tiny} shows that masking makes the task learnable but struggles as $D$ and $M$ scale.
We find minor differences in the two masking types and try mixing them, as they may provide different benefits (token replacement tells the model not to trust single-edge lookups while a masked token prevents these).    

 Finding that a given method makes the PST learnable but does not scale will be a consistent pattern across methods.  {\bf Our focus for these methods is, a), showing that the task becomes learnable and, b), explaining why.}  We conjecture about limitations in scalability in Sec.\@ \ref{sec:limits} 
 which were also observed by \citet[see Appx.\@ \ref{appx:graph-learnability}]{saparov2025transformers}.

\subsubsection{Unadulterated Task Decomposition}\label{sec:masking-decomp}

\begin{figure}
\centering 
\begin{subfigure}[b]{0.47\textwidth}  
\centering  
   \includegraphics[width=1.\linewidth, trim={.85cm .15cm .72cm .3cm},clip]{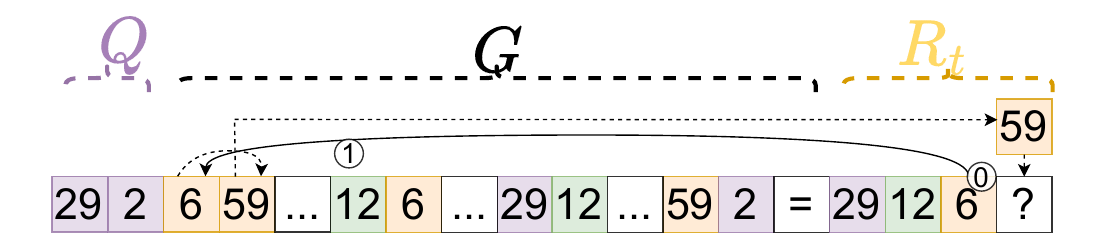}
   \caption{The CHC as a single-edge lookup when `6' is TFed.}
   \label{fig:alg1} 
\end{subfigure}
\begin{subfigure}[b]{0.47\textwidth}
\centering
   \includegraphics[width=1.\linewidth, trim={.65cm 0.cm .7cm 0.cm},clip]{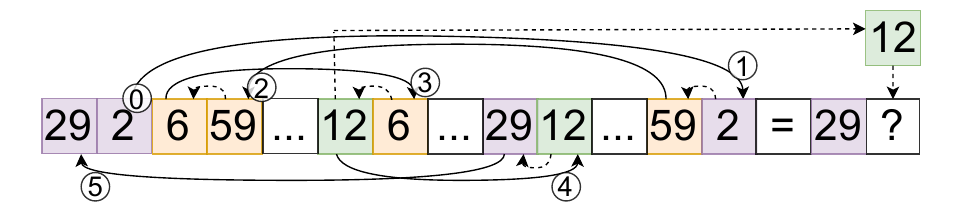}
   \caption{Arm reconstruction needed for predicting $l_t=$`12'.  Note how the algorithm must work backward from $t=$`2'.  Steps 4 and 5 are verification steps to match $l_t$ being adjacent to $s$.  These are not actually learnt due to positional shortcuts.}
   \label{fig:alg2}
\end{subfigure}
\begin{subfigure}[b]{0.47\textwidth}
\centering
   \includegraphics[width=1.\linewidth, trim={.65cm 0.cm .7cm .1cm},clip]{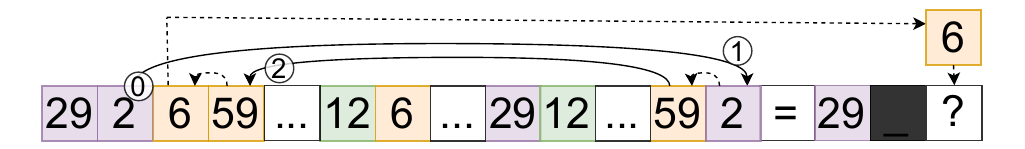}
   \caption{Predicting `6' with $l_t$ masked.  {\bf This avoids the CHC and induces learning a decomposed subtask which mirrors the steps needed to predict $l_t$ in the core task while also simplifying the task since it is only a subset of the steps.} This exact decomposition requires a front-spanning mask that disallows conditioning on any prior ground-truths and explains the (limited) success of the `teacher-less' model.}  
   \label{fig:alg3}
\end{subfigure}
\begin{subfigure}[b]{0.47\textwidth}
\centering
   \includegraphics[width=1.\linewidth, trim={.65cm 0.cm .7cm .1cm},clip]{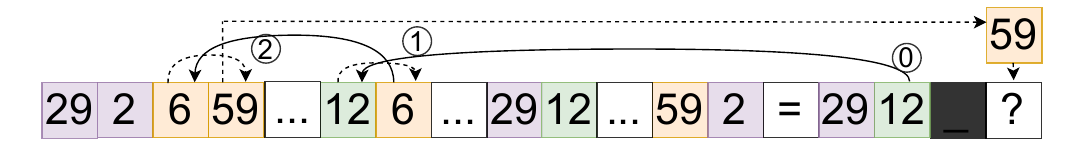}
   \caption{Providing TFed input before masking induces multi-edge lookup but potentially with the forward algorithm.}
   \label{fig:alg4}
\end{subfigure}
\vspace{-3pt}
\caption{Algorithmic steps performed in the CHC and arm reconstruction, also with masking (blacked-out).}
\label{fig:psg2-algs}
\end{figure}

We 
show how masking prevents the CHC in Fig.\@ \ref{fig:psg2-algs}.  First, consider the CHC in Fig.\@ \ref{fig:alg1} and the needed algorithm for predicting $l_t$ in Fig.\@ \ref{fig:alg2}.  The CHC learns a forward alg.\@ from the prior token, while the required alg.\@ must work backward from the target query. Figs.\@ \ref{fig:alg3} and \ref{fig:alg4} show how masking induces multi-edge lookups.  In Fig.\@ \ref{fig:alg3}, when all prior tokens are masked, it induces learning a subset of steps for the required alg.  {\bf This provides a deeper explanation for why masking works beyond preventing the CHC; it induces task decomposition}.  This also explains why unadulterated sequential supervision is important.  Decomposition can occur because arm reconstruction is inherently recursive. 

Fig.\@ \ref{fig:alg4} shows that having unmasked prior tokens may lead to learning a forward alg.  While these seem similar to a human, they are different algs.\@ and it's unclear if they mutually support each other in terms of learning to predict $l_t$, i.e., does a subtask need to exactly mirror the core task for subtask decomposition to help learning the core task. 

\vspace{-3pt}


\subsection{Alternative Sequential Distributions}\label{sec:alt-distros}

We consider learning a distribution over the next \underline{tokens} instead of the next \underline{token}, i.e., multi-token prediction.  This is done via learning a belief-state, $B$, which is a hidden-state that supports making future predictions via some linear function of $B$ i.e.\@,  $P_B(x_{i:M}\,|\, f(B=x_{<i}))$ \citep{hu2025learning}.

We present three simple methods for learning this future distribution, $P_B$: bag-of-words (BoW), label-smoothing (LS), 
and ranking.  \citet{yin-etal-2024-semformer} used a BoW baseline with $R_t$ as the bag. 
This performed nearly as well as their proposed model 
and solved the task in the majority of cases.\footnote{They used pretrained GPT2 models in their experiments.} 
We exclude prior tokens ($<i$) from the bag so as to only contain future tokens at each step.  
Note BoW is equivalent to LS with uniform smoothing. 

BoW is based on the inductive bias that nodes in $R_t$ are more important than nodes in the other arms.  
We can extend this with another inductive bias which assumes that near-present tokens are more important than far-future tokens, i.e.\@, that the order matters. This can be achieved using monotonically decreasing label weights.   Thus, LS requires hand-crafted weights to form a hand-crafted distribution that the model tries to match.  We can avoid this ad-hoc approach via an equivalence between LS and ranking   \citep{Frydenlund_Singh_Rudzicz_2022}.

\subsubsection{Ranking-into-the-Future (RITF)}

We generalize from LS to ranking-into-the-future by constructing rank targets and training with a rank-based loss, providing a structured loss over multiple tokens at each time-step.  As we are using future tokens, the structure is over the sequence, and the sequential order is used for the rank-order.  

Let the future distribution {\em at a single step $i$} be $P_{B,\, i}(x_i \succ x_{i+1} \succ \dots \succ x_{M})$ such that the scores of sequential tokens decreases monotonically from time-step $i$.  Let $\sigma_i = f(B=x_{<i})$ be a vector of these scores or logits. Then we use a pair-wise hinge loss over the entire sequence in $R_t$ s.t.\@ 
\begin{equation}\label{eq:pairwise-loss}
L_B = \sum_{i=1}^{M}\sum_{j=i}^M\sum_{k=j+1}^M \max(0,\, 1 - 
(\sigma_{i}[j] - \sigma_{i}[k])) 
\end{equation}

We incorporate another bias that ranks tokens in $R_t$ above all others, encoding the concept that the correct arm is more important than the others.\footnote{This is not included
in Eq.\@ \ref{eq:pairwise-loss}. See Appx.\@ \ref{appx:alt-distro-results}.}  
This creates very
dense supervision with $M(M-1)/2$ intra- and $M^2(|V|-M)$ inner-arm pairs.  Initial experimentation found this was crucial.  This makes sense as, when the model fails to learn to predict $l_t$, it learns a uniform distribution over the set of leading nodes, meaning they have equal scores.  This secondary bias creates supervision that $l_t$ is more important than the other leading nodes.  Note that this is already implicitly done in any cross-entropy loss, including BoW and LS, as cross-entropy works by promoting the singular ground-truth while demoting all other nodes.  


\subsubsection{Results and Discussion}


\begin{figure}
    \centering  
    \includegraphics[scale=0.48, trim={.50cm 3.80cm .50cm 1.90cm},clip]{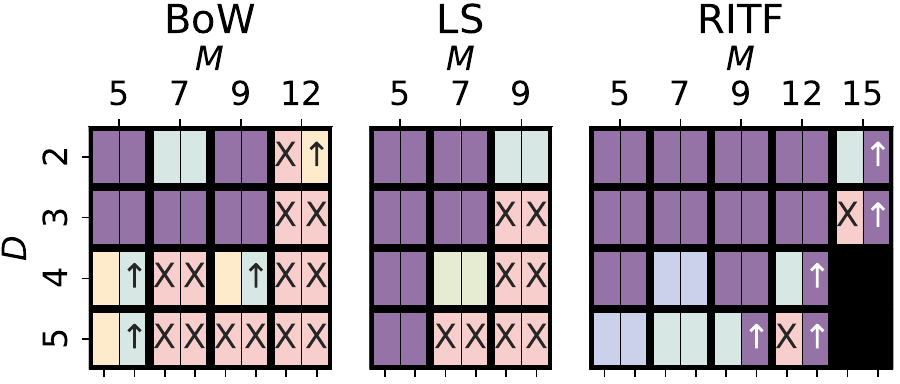}
    \vspace{-8pt}
\caption{Alt.\@ distro.\@ results (full Tbl.\@ \ref{tbl:alt-distro-results} in Appx.\@ \ref{appx:alt-distro-results}).}
\label{tbl:alt-distro-results-tiny}
\end{figure}

Fig. \ref{tbl:alt-distro-results-tiny} shows that RITF is superior to both BoW and LS. 
For LS we use a stepped monotonically decreasing weight \citep{Frydenlund_Singh_Rudzicz_2022}.  We believe this does not work as it couples the inductive bias with loss scaling (so future tokens have tiny weights).  We know the inductive bias is not at fault, as it is the same as in RITF.  {\bf This shows that it is easier to specify rules than specific weights.}

More importantly, future predictions make the task learnable for multiple related reasons.  First, the multi-token  loss requires multi-edge lookups and so induces decomposition.  Second, it avoids adulteration by skipping adjacent inputs for all targets at $> i + 1$.  {\bf This is the same as masking, except instead of 
using noised input to cause the skip, it is implicitly defined as part of the loss.}  The first step is also fully masked, thus inducing the desired backward alg. for all tokens. 
Third, by applying this at each time-step, {\bf the loss induces novel dense decomposition across the sequence} where learning $P_{B,\,i+1}$ is a sub-problem of learning $P_{B,\,i}$ (see Appx.\@ \ref{appx:masking-and-alt} for an expanded discussion). 


\subsection{Scratchpads (SP) to Increase Supervision}\label{sec::scratchpads}

SPs predict an intermediate sequence before the target sequence, providing auxiliary input and target supervision \citep{nye2022show}.  Both the reverse arm order and the arm-wise graph shuffle 
make the task trivial, and so would be obvious SPs.  These are problem-specific and do not prevent adulteration and, so, are not insightful. We present alternatives.    

Instead, for arm reconstruction (AR-SP), we generalize the reverse 
order to generate the arm nodes as a BoW in any order.  As there are $M!$ 
orderings, we use LS over the choices.  This unifies the auxiliary BoW and single next-token distributions in Sec.\@ \ref{sec:alt-distros} since, 
{\em the next $M$ tokens are the BoW}.
We can avoid LS by determining a canonical ordering via sorting by node values. {\bf This introduces node semantics}. 
This may provide strong supervision as nodes in $R_t$ need to be identified and then ordered, which requires making comparisons that will not match the source-side edges, thus avoiding the bad interaction that causes the CHC. 

We also use SPs that reconstruct the entire graph by ordering the arms (GR-SP).  Full reconstruction would cause adulteration, so we just match leading- and target-node pairs. 
We order the arms by leading- or target-node value, again, introducing semantics (see Fig.\@ \ref{fig2:sp-graph-recon} in Appx.\@ \ref{appx:sp-graph-recon}). 



\subsubsection{Results and Discussion}

AR-SP results can be seen in Tbl.\@ \ref{tbl:sp-arm-recon} in Appx.\@ \ref{appx:sp-arm-recon}.  We report the sequential accuracy for $R_t$ and the SP separately to isolate where any errors occur.  The reverse SP is trivially learnt as expected.  Some BoW trials are successful.   Importantly, this can be explained by inducing subtask decomposition (Appx.\@ \ref{appx:bow-ls-and-scratchpad}).  While the BoW SP make the PST learnable (on small scales), its performance is disappointing, since an obvious solution (to a human) exists, which would be to generate the BoW in the reverse order, however, this solution is not found.  As Tbl.\@ \ref{tbl:sp-arm-recon}  shows, the models fail to correctly predict the SP and then fails to predict $R_t$ when conditioning on the incorrect SP. This failure is informative as it shows {\bf the reverse solution is only trivial when it is provided with direct supervision} (the same sequence supervision that causes the CHC).

The performance of the sorted SP is harder to explain.  Sorting naturally decomposes, but, by design, is agnostic to graph edges (except for the identification step).  It may be that this subtask does not mutually support learning the PST task.  However, the issue is that the model fails to learn to sort at all at scale, so we suspect that this the same scaling issue affecting the other methods.     

GR-SP results can be seen in Tbl.\@ \ref{tbl:sp-graph-recon} in Appx.\@ \ref{appx:sp-graph-recon}.  These only learn to solve the task in 4/80 trials and  this is exceedingly informative.  We have four variants of the GR-SP, as we can go from leading- to target-nodes or vice versa, and then either sort by leading- or target-node values.  Fig.\@ \ref{fig2:sp-graph-recon-plots} in Appx.\@ \ref{appx:sp-graph-recon}  plots the accuracy of each SP token across training.  The models learn to correctly identify the needed sets of leading and target nodes.  This is done by single-edge shortcuts; leading nodes by adjacency to $s$ and targets nodes by counting degree.  The models also correctly learn to sort either the leading- or target-nodes.  This means that for the SP where we go from leading-to-targets, sorted by leading nodes, the model can correctly identify the first leading-node but fails to connect it with its paired target. The same thing happens using targets-to-leading, sorted by target nodes. 

{\bf  In both cases, the model knows which arm to reconstruct, and can condition on either the correct leading or target node,  but still does not learn the actual reconstruction!  Here, all the model needs to do is deterministic path-following with no planning to choose the correct arm.  This begs the question: if the solution is deterministic and does not require choices, is this actually a planning problem? } These and the BoW SP results indicate that arm reconstruction is what makes the task hard -- not planning.  
Thus, these negative results are consistent with and indirectly support the theory that supervised task decomposition is necessary, since the scratchpads the fail are those that do not induce decomposition.





\subsection{From Path-Star to Tree-Star}\label{sec:trees}

Here, we partially prevent the CHC and induce task decomposition via the source-side modification of changing the graph topology from paths to trees.   

One way of preventing the CHC is to remove or modify edges in $G$ to prevent single-edge lookups. We achieve this by generalizing the task to consider arms as trees instead of paths.  In particular, we train on trees but evaluate on path-star graphs.  Training on trees slightly changes the training objective as we are not generating the arm -- which is more generally the shortest path from $s$ to $t$ -- but rather an equivalent pre-order traversal of the tree.  This introduces a problem in that such a traversal requires a planner tree to determine the order of multiple child nodes in the traversal.   Encoding it as a planner tree will induce new undesired shortcuts.    

\begin{figure}
    \centering
    \includegraphics[scale=0.42, trim={.85cm .42cm .95cm .65cm},clip]{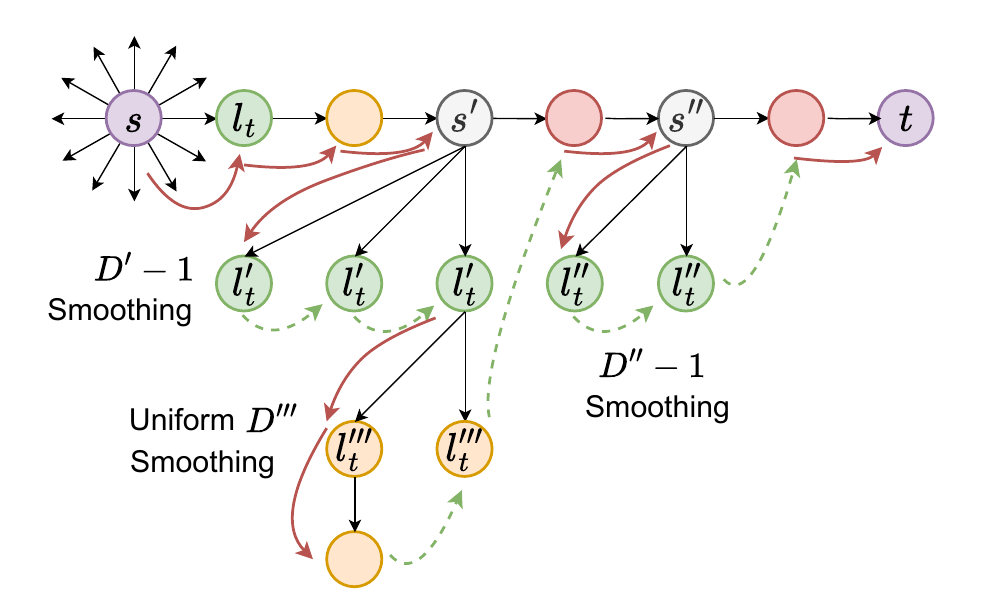}  
    \caption{Red and green-dashed arrows form the desired pre-order traversal; red edges are included in $G$, while green edges are not, and hence are immune to the CHC. 
    }
    \label{fig:ps2-tree1}
    \vspace{-4pt}
\end{figure}

We employ a trick to avoid this.   By task definition, $t$ must be the last generated token. This precludes any subtree containing $t$ from being generated before the others.  Then, given a subtree containing $D'$ child nodes (including $t$), the distribution over these children being valid continuations of a traversal is asymmetric in that $t$ is excluded but uniform over the remaining 
$D'-1$ choices. However, this only works for subtrees containing $t$.  
This is achieved via LS over valid child nodes during training. 
Call these {\em $D$-ary trees}.  See Fig.\@ \ref{fig:ps2-tree1}.  


\begin{figure}
    \centering
    \includegraphics[scale=0.42, trim={.7cm .42cm .7cm .35cm},clip]{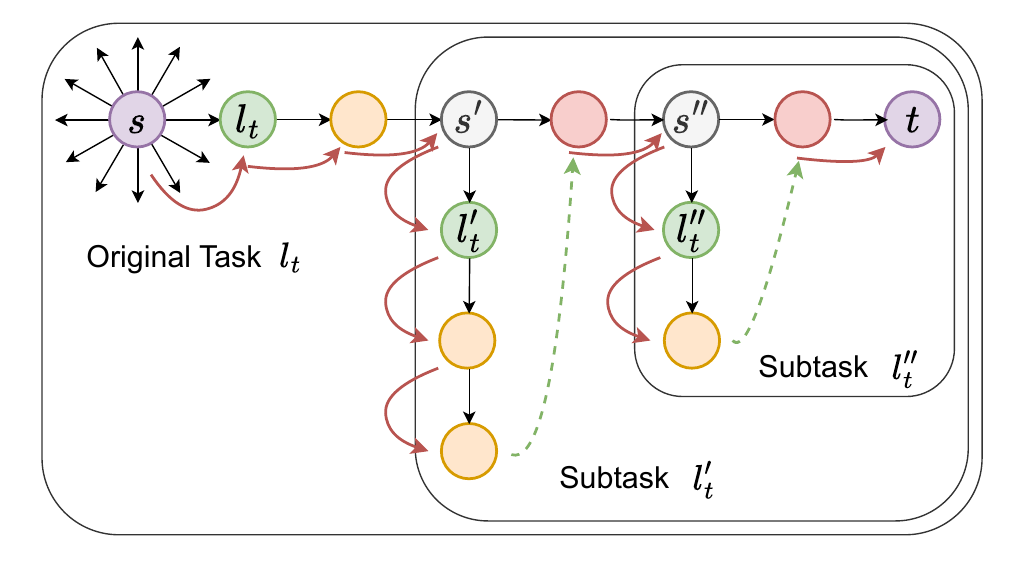}  
    \caption{A split tree.  Each split induces a
    decomposition and there is only a single valid pre-order traversal. 
    }
    \label{fig:ps2-tree2}
    \vspace{-4pt}
\end{figure}

Following this logic, we can design the tree to resolve any ordering ambiguity and avoid needing LS with a deterministic traversal by only allowing subtrees containing $t$ to have a max of 2 children and all others one child, i.e.\@, structurally-lopsided binary trees. Call these {\em split trees}.  See Fig.\@ \ref{fig:ps2-tree2}. 

\subsubsection{Results and Discussion}

\begin{figure}
    \centering  
    \includegraphics[scale=0.36, trim={.50cm 1.9cm .50cm .0cm},clip]{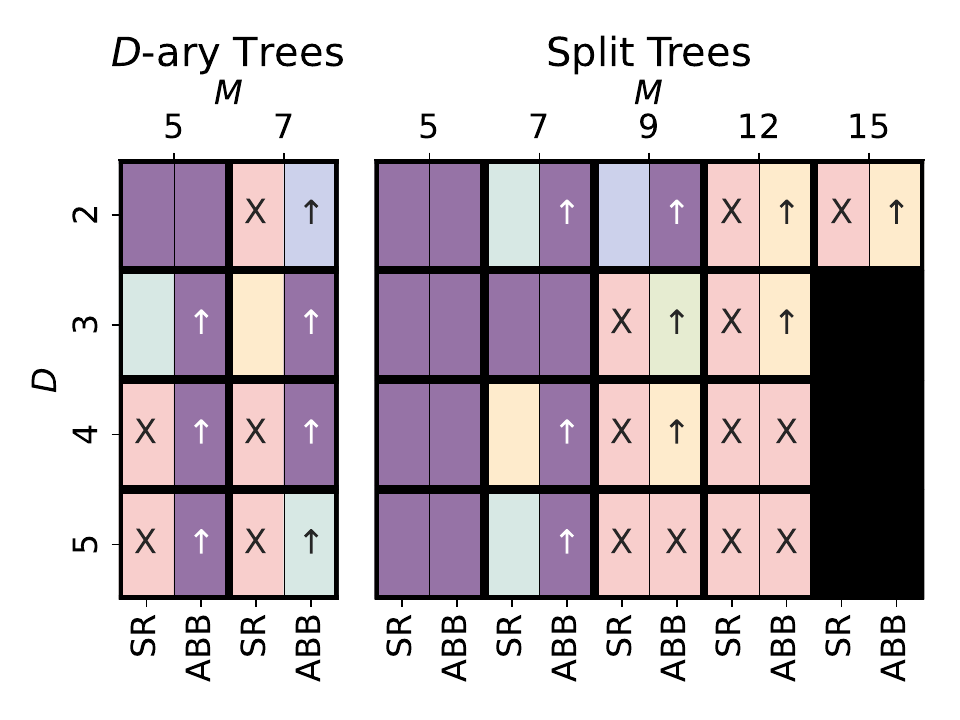}
    \vspace{-8pt}
\caption{Split tree results (Tbl.\@ \ref{tbl:tree-results} in Appx.\@ \ref{appx:tree-results}) }
\label{tbl:tree-results-tiny}
\end{figure}

Fig.\@ \ref{tbl:tree-results-tiny} shows that training on split trees makes the task learnable.  This can again be explained as inducing task decomposition.  In Fig.\@ \ref{fig:ps2-tree2}, each subtask (marked as primes) is similar to a path-star graph with $D'=2$ where a new start node, $s'$, is any node with two children and the correct leading node, $l_t'$, is the first node of the subtree not containing $t$.  It is interesting that experiments where $D > 2$ work since the induced subtask is restricted to $D'=2$ and thus does not match the original task.  

We find that the $D$-ary trees do not learn the task outside of a few cases, despite introducing a similar decomposition.  We conjecture that smoothing over subtrees with $D$ choices creates a deficient subtask that mirrors the adulterated PST task since we are explicitly forcing the model to learn a uniform distribution over leading nodes and this matches the undesired learnt behaviour of models that fail the task, i.e., we are directly teaching the model to learn 
the exact behaviour we are trying to avoid.


Training on trees and evaluating path-star graphs may look like an exotic solution, but we stress this is actually a generalization of graph topology and one that does not go far enough.  We suspect that the best graph topology would be one that allows for perfect decomposition, where each subtask exactly mirrors the core task except for a change in the number of recurrent steps needed to solve the task.  That is, the choice of graph topology will affect subtask homology.   This conjecture would explain the necessity of `balanced' graphs beyond preventing shortcuts    \citep{saparov2025transformers}.\footnote{Shortcuts are probably the symptom, not the illness.}


Training on trees and then evaluating path-star graphs is also interesting since it is counterintuitive from the perspective of in-domain learning; we require training on trees to generalize to paths when direct training on paths fails.  Ironically, the PST is defined as it is to rule out out-of-domain effects \citep{bachmann2024the}. From the perspective of adulteration, paths are an excessively informative graph structure.  Also, 
training on paths to evaluate paths is more direct task supervision than training on trees to evaluate paths.

\subsection{Generalized Queries}\label{sec:alt-queries}

Given the sensitivity conjecture, we consider if providing more than one node from $R_t$ in $Q$ will decrease sensitivity.  We do this by sampling a subset of $R_t$ (in any order to avoid adulteration).  This is similar to token masking in that we are supporting prediction via providing multiple tokens from $R_t$ to condition on, except this is being applied on the source-side.    
During inference, only $t$ is given.  

\subsubsection{Results and Discussion}


\begin{figure}
    \centering  
    \includegraphics[scale=0.36, trim={.50cm 1.25cm .50cm .0cm},clip]{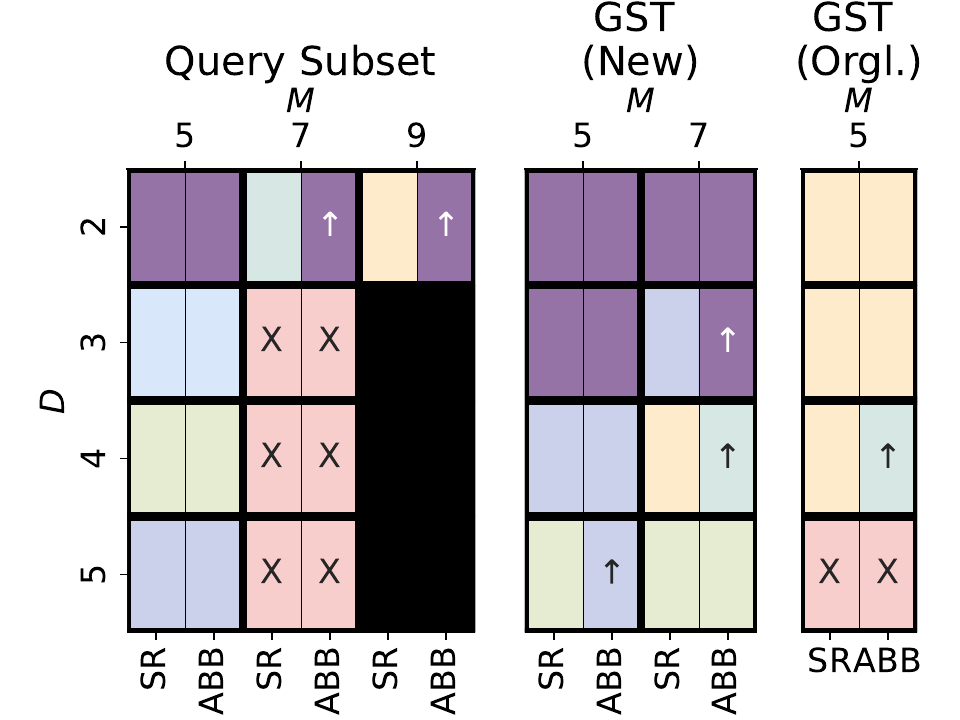}
    \vspace{-8pt}
\caption{Results for general query methods.  Full Tbl.\@ \ref{tbl:query-results} in Appx.\@ \ref{appx:query-results}, along with original-setting experiments.}
\label{tbl:query-results-tiny}
\end{figure}

\begin{figure}
    \centering  
    \includegraphics[scale=0.4, trim={.65cm .1cm .45cm .1cm},clip]{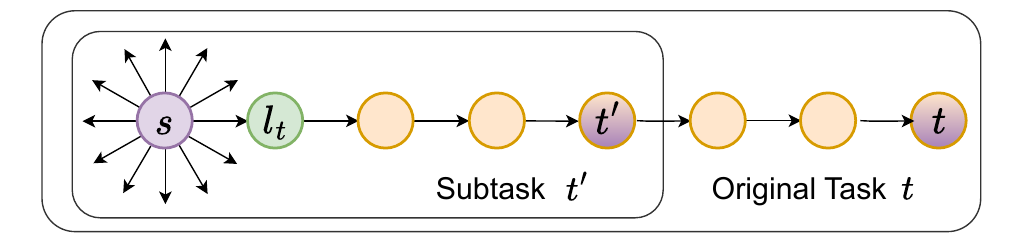}  
    \caption{Sampling $t$ as task decomposition.  $t$ is the original target at position $M$ and  $t'$ is a sampled target. 
    }
    \label{fig:ps2-general-q}
    \vspace{-16pt}
\end{figure}

Fig.\@ \ref{tbl:query-results-tiny} shows that using a subset of $R_t$ makes the task learnable for small graphs.  To verify that that is due to decreasing sensitivity, we tried sampling a single node as a general single target (GST) from $R_t$. 
We find that not only does this make the task learnable, it performs better than using a subset.  We expect this is because a subset introduces too much noise. However, if this does not work because of reduced sensitivity, why does it work? Because it induces task decomposition (illustrated in Fig.\@ \ref{fig:ps2-general-q}).    

\citet{hu2025learning} performed this same experiment and found it did not learn the task.\footnote{And they performed this experiment explicitly to determine if subtask decomposition makes the task learnable.}  To explain this contradiction, we experiment
using the original task settings (using an offline dataset, $|V|=100$, and $Q$ after $G$). 
Here we find that the task is much harder to learn, with only 3/20 trials succeeding (Tbl.\@ \ref{tbl:query-results} in Appx.\@ \ref{appx:query-results}).  This implies that it would be easy to find only negative results, especially if seeded trials were not used.  We argue further about issues with hyperparameters in Appx.\@ \ref{appx:query-results}.  {\bf This highlights the importance of using an online dataset, and how other issues -- not related to planning -- contribute to the PST being unlearnable in the original experimental setting.}

\subsection{Generalized Length Decomposition}\label{sec:multi-len}

Given the above, it should be obvious now that a direct way to induce subtask decomposition would be to supervise the training process by sampling different-sized graphs.  Tbl.\@ \ref{tbl:multi-results} in Appx.\@ \ref{appx:multi-results}  shows this makes the task learnable as expected.  Training on various values of $D$ does not seem to help. Combining general length and target sampling improves performance over just doing the former.  We expect these results would be better if given more training time, as this introduces a lot of noise.  

{\bf These results show that training on various lengths is not just for out-of-domain generalization, but also promotes in-domain learning. 
This also prevents positional shortcuts.} 


\section{Conclusion}

We have demonstrated that the path-star task, which was seemingly unlearnable, is actually learnable with decoder-only models.
We have shown how the original task is designed with adulterated supervision, explained why this makes it unlearnable due to obscuring or absorbing decomposition supervision, and shown that preventing the CHC is not critical for learning the task given some decomposition supervision.  We developed multiple 
methods to overcome this lack of supervision, with all retaining the next-token prediction paradigm with standard training via teacher-forcing.\footnote{Except for the future distributions.}


{\bf We have empirically demonstrated that decomposition is critical for learning to search over graphs.}  This is strongly supported by the fact that the methods we have developed are all orthogonal to each other but can all be explained as inducing subtask decomposition in some form.\footnote{See Appx.\@ \ref{appx:summary} for a summary and comparison of methods.}   

A few informative negative results suggest that the core difficulty of the task does not concern planning at all but rather graph reconstruction and that seemingly trivial solutions will not be found unless directly supervised. 
We also show that graph reconstruction is made more difficult for decoder-only models due to causal constraints via the causal-wise shuffling experiments.

Our work serves as a bridge between \citet{bachmann2024the} and \citet{saparov2025transformers} by providing explanation for why the graph search task presented by the former is seemingly unlearnable, while the very similar graph search task presented by the latter is learnable; the latter provided decomposition supervision (specifically they incorporated general graph topologies and lengths), while the former did not due to adulteration.  
Note, these still permit the CHC. While this is not critical for learning the task, preventing all shortcuts, say by masking, may be important for generalization.

As searching is recursively defined, 
decomposition is inherent in the original task.  Thus, we can induce 
it using the original supervision -- provided we are careful not to adulterate it.  This contrasts with other tasks that require introducing scratchpads or extra supervisory information 
 -- done by modifying the task itself or learning a secondary subtask that is decompositional \citep{wies2023subtask}.

{\bf If one were concerned about the implications that the empirical results of the path-star task had on the sufficiency of the next-token prediction paradigm for planning tasks, this work alleviates those concerns.  If one is skeptical about such conjectures, this validates their beliefs with an explanation of why the PST in its original form is unlearnable.  Our findings show that the task is fragile, where minor changes induce decomposition and make it leanrable, which indicates these issues will not apply to  -- or be nearly as potent to -- complex tasks as conjectured.}


\section{Limitations}\label{sec:limits}

{\bf Scaling issues:} The major limitation of our work is that each method fails to scale with either $D$ or $M$.  We believe that using graph topologies that allow for stronger and more consistent decomposition, where each subtask mirrors the main task, will be key for scalability.  Thus, we think that path-star experimental setting is not a suitable environment to consider how the methods we have developed will scale to larger graphs.  Instead, we believe using more general graphs would be a better environment and leave this to future work. However, \citet{saparov2025transformers} observed similar scaling issues using more general graphs, which suggests that the issue does not stem from topology by itself (see Appx. \ref{appx:graph-learnability} for a comparison of our works). Still, this would let us avoid or isolate issues related to using topology when considering scale.  

We conjecture that different issues may affect the scaling of $D$ and $M$.  $M$ scales with the number of model layers \citep{frydenlund-2024-mystery}.  It is unclear if learning the log alg.\@ is harder than the linear alg.  

{\bf Model parameterization:}  \citet{frydenlund-2024-mystery} empirically demonstrated differences in performance on the PST between various model parameterizations.  The difference between decoder-only models and the others is due to the causal constraint.  We show that overcoming the constraint makes the task learnable, but not exactly what induces learning to overcome it.  We do, however, make a connection to the target-side encoder models by showing that masking, the same method used to train these models, makes the task learnable due to decomposition.  
Now that we have successfully shown that decoder-only models can learn the task, we can better explore the learnability conditions between the models in future work.  Differences in parameterizations have been shown to be important in other symbolic tasks \citep{ye2025beyond, ye2025implicit}.



{\bf The value of the path-star task:}   Given our findings, we argue that the PST is not suitable for evaluating the performance of novel  methods or architectures for planning \citep[among others]{yin-etal-2024-semformer, hu2025learning}.\footnote{They also both perform additional non-PST experiments.}  
We also argue that graph search is not a stand-in for general search.  
Given the finesse required to learn the PST and issues with scalability, we suggest that graph search may be hard due to issues that only apply to these experimental settings and caution against making claims based on these tasks being used as surrogates for other search tasks on natural data (see below for positive uses of the task). 
While our goal of this work is to show that next-token prediction is sufficient for learning the task, we do not consider what method is best.
Given the above, we have introduced the novel ranking-into-the-future method, but have not evaluated its potential for planning on a suitable task.  We frame RITF as a change in distribution from next-token prediction.  This is important to the discussion of the paradigm in this work, but may not be important for other uses.

{\bf Semantics as high-order graph structure:} The path-star task is semanticless. This often applies to other minimal graph search tasks.  While we do little experimentation in this direction, we conjecture that this plays an important role in learnability, where the right semantics would probably make graph reconstruction significantly easier.  This relates to not using pretrained models, which would inherit natural language semantics.

{\bf Why is task decomposition necessary:}  We have empirically shown that task decomposition is necessary for learning the task.  While it is intuitive why supervising decomposition will help learning, we do not explain why it is necessary.  Thus, characterizing the core underlying difficulty is still an open question.  We expect solving this will be insightful for learnability theory. 
{\bf Thus we believe the original adulterated form of the path-star task is of scientific value and hope that research into it will continue.}  We did not find positive results trying to decrease sensitivity with multiple query nodes, but our results are inconclusive and may be due to an increase in noise.  Thus, the sensitivity conjecture is still a possible explanation.

{\bf Shortcuts:} The PST is a great framework for exploring shortcut learning.  Different variations result in different shortcuts, especially when using the SPs.  One question we have is if shortcuts, once learnt, actively harm learning the desired task or if they are just benign symptoms of other issues?

{\bf Alternative examples of adulteration:}  We describe adulteration as a generic issue but only consider the context of graph searching.  We believe it is a useful term for identifying similar issues which can be solved via similar methods.  For example, \citet{chang2025language} considered trying to learn to count by providing a model with a contiguous sequence of numbers and training using next-token prediction.  As they point out, this will be ineffective due to trivial bigram shortcuts.  This is because they have presented the task in an adulterated form.

\section*{Acknowledgments}

We thank Frank Rudzicz and Gagandeep Singh for helping to proofread, and Rich Zemel for suggesting some clarifying experiments.  

Resources used in preparing this research were provided, in part, by the Province of Ontario, the Government of Canada through CIFAR, and companies sponsoring the Vector Institute.

\bibliography{acl_2025_post_submission}

\appendix

\section{Relationships Between Methods}\label{appx:summary}


This section provides a summary of the methods introduced and the various relations between the methods, focusing on how each induces subtask decomposition.  We also outline how each method either prevents, alleviates, or just ignores the CHC.  Note that, except for the alternative sequential distributions, all methods use standard teacher-forcing and next-token prediction.

\subsection{The Different Forms of Masking}

\begin{figure}
    \centering
        \begin{subfigure}[b]{0.47\textwidth}
\centering
        \includegraphics[scale=.73, trim={.5cm .2cm .2cm .2cm},clip]{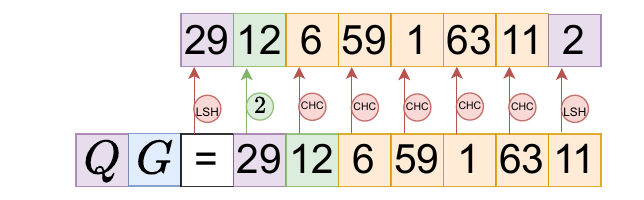}
        \caption{The original task with \textbf{fully observed} teacher-forced input induces learning the CHC for all but the leading node.  Note, we have extended the length of the arm from Fig.\@ \ref{fig:psg2}. }\label{fig2:sum-m0}
    \end{subfigure}
    
    \begin{subfigure}[b]{0.47\textwidth}
\centering
        \includegraphics[scale=.73, trim={.5cm .2cm .2cm .2cm},clip]{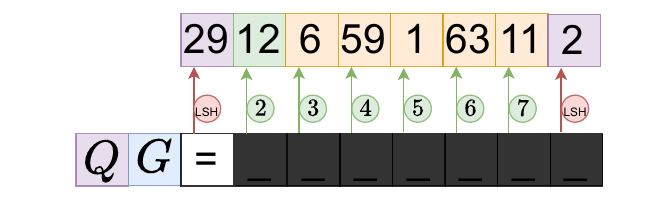}
        \caption{An example of a \textbf{fully masked} sequence (equivalent to the teacher-less model or NAR in terms of masking).  All five predictions are conditioned via different masked tokens (at unique positions).  Each prediction must use the backward algorithm to determine the correct target.}\label{fig2:sum-m1}
    \end{subfigure}
    
    \begin{subfigure}[b]{0.47\textwidth}
\centering
        \includegraphics[scale=.73, trim={.5cm .2cm .2cm .2cm},clip]{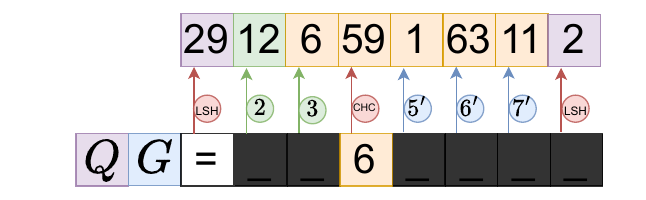}
        \caption{A \textbf{partially masked} sequence. Unmasked tokens cause the CHC to be learnt for the corresponding prediction.   Predictions after an unmasked token \textit{may} use the forward algorithm to determine the correct target (indicated with primes).}\label{fig2:sum-m2}
    \end{subfigure}

        \begin{subfigure}[b]{0.47\textwidth}
\centering
        \includegraphics[scale=1.3, trim={.5cm .5cm .5cm .5cm},clip]{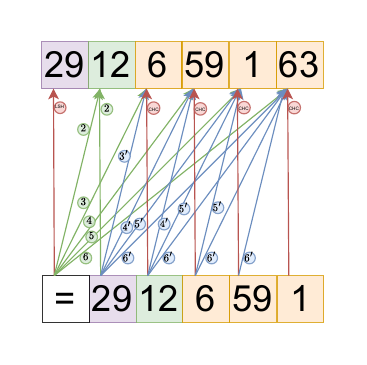}
        \caption{\textbf{Alternative sequential distributions} create dense decomposition supervision.
        The CHC is learnt for one prediction at all time-steps, except for the leading node (same as in the original task with no masking).  In the first time-step, all predictions corresponding to the fully masked must be learnt; the difference between this and Fig.\@ \ref{fig2:sum-m1} is that all predictions are conditioned from a single hidden-state instead of being separated into five temporally distinct hidden-states created via different masked token inputs.  Then, for all other future predictions, the forward algorithm \textit{may} be used to determine the correct targets since the model conditions on ground-truth tokens on those time-steps.  Also, some of these predictions may be repeated at different time-steps. Nodes `11' and `2' are omitted for visibility.}\label{fig2:sum-m3}
    \end{subfigure}
\caption{A comparison of subtask decomposition between masking and alternative sequential distributions.}\label{fig2:sum-m}
\end{figure}

A comparison between fully observed, fully masked, partially masked, and alternative sequential distributions is given in Fig.\@ \ref{fig2:sum-m}.  These are all target-side modifications to the task.  Masking modifies the target-side input, while the alternative sequential distributions modify the target-side targets.  However, this can also be viewed as inducing the same masking modification on the target-side inputs, except that this is done implicitly instead of explicitly via masked tokens.


We illustrate teacher-forced inputs and corresponding targets while also labeling how the model achieves each prediction in Fig.\@ \ref{fig2:sum-m}. Predictions learnt by the CHC are indicated with `CHC' red labels. 
Because the path-star task uses graphs of the same size, the start and target nodes can be trivially predicted via {\em length shortcuts} as they are both included in $Q$ (see Fig.\@ \ref{fig:baseline1}).   These are indicated with `LSH' red labels.
Predictions for the second through seventh target tokens that use the desired backward algorithm are indicated with green labels `2' through `7'.  Those that may use the forward algorithm are indicated with primes and blue labels.  Consider again Fig.\@ \ref{fig:psg2-algs} in Sec.\@ \ref{sec:masking-decomp} for a more detailed illustration of how subtask decomposition is achieved via masking.

We illustrate the original adulterated task in Fig.\@ \ref{fig2:sum-m0}, where the CHC is learnt for all predictions except for the leading node.  
The CHC was the mechanism originally attributed to causing the task to be unlearnable by \citet{bachmann2024the}.
This motivated the use of a `teacher-less' model, which fully masks the target-side inputs.\footnote{This has a technical difference with our masking method.  Our masking keeps true autoregressive conditioning at all steps while `teacher-less' models only condition on a single mask token per time-step and, hence, can predict all masked positions in parallel (like a non-autoregressive model).  This difference only matters in that our masking method is a standard method to use with language models, while the teacher-less method is not, and this then helps to support our argument that standard methods are sufficient to solve the task and that alternative models are not needed.} We illustrate the effect of full masking in Fig.\@ \ref{fig2:sum-m1}.  This prevents the CHC and induces learning of the desired backward algorithm for all predictions.\footnote{Thus, the only methods that completely prevent learning the CHC are full masking, the teacher-less model, and `fully' non-autoregressive models \citep{frydenlund-2024-mystery}.} This is in contrast to partial or sampled masking, illustrated in Fig.\@ \ref{fig2:sum-m2}, which still allows the CHC to be learnt, hence only alleviating the issue.\footnote{This also means that the iterative autoregressive models and the discrete diffusion models only alleviating the issue (see Appx.\@ \ref{appx:iar}).}\hochkomma\footnote{\citet{bachmann2024the} introduced the teacher-less model with masking as a direct anti-CHC mechanism on the assumption that it was the mechanism causing the task to be unlearnable and that its prevention is necessary to make the task learnable.  However, we show that masking also introduces subtask decomposition.  {\bf This raises the question: what is the critical aspect of this method that allows for the task to be learnt?  We argue that decomposition is the critical aspect, not CHC-prevention.}  This is held up by the fact that, as mentioned above, partial masking doesn't actually fully prevent the CHC and still learns the task, and that other methods like alternative sequence distributions, general queries, and general lengths, a), learn the task, b), do not prevent the CHC at all, but, c), do induce decomposition.  Hence, decomposition is necessary to learn the task, but preventing the CHC is not, and so, this must be the critical aspect of why masking works.} 

Whereas full masking induces learning of the backward algorithm for predictions, partial masking means that predictions after any unmasked ground-truth inputs may use the forward algorithm.  {\bf The differences in algorithms may be important since, while they are very similar, the backward algorithm is required to be learnt to correctly predict the leading node, and as the forward algorithm is different, this means it is not an exact subtask decomposition.  Assuming that learning the forward algorithm supports learning the backward algorithm, this suggests that subtask decomposition does not need to be exact mirror of the original or core task.}   We also discuss this idea in relation to how tree topology induces decomposition in Sec.\@ \ref{sec:trees}.  Given the success of these methods, it seems like subtask decomposition does not need to be exact.  This makes a formal definition of decomposition difficult (whereas in the exact case, we could just define a task recursively and treat each recursive step as a decomposition).   

\subsection{Masking and Alternative Sequential Distributions}\label{appx:masking-and-alt}

Comparing masking to alternative sequential distributions described in Sec.\@ \ref{sec:alt-distros}, 
{\bf alternative sequential distributions act like a combination of all three of the described masking types -- fully observed (no mask), fully masked, and partially masked -- all at once.} This is achieved due to having multiple targets per time-step (i.e., being multi-token prediction methods), which leads to a greater density of target supervision (and hence predictions).  Or to put another way, more targets are given, and a subset of these targets induce predictions like the fully observed task, a subset induces predictions like the fully masked task, and the remaining subset induces predictions like the partially masked task. 

This is illustrated in Fig.\@ \ref{fig2:sum-m3}.  Here, teacher-forcing with fully observed input is used.  This means that the same predictions in Fig.\@ \ref{fig2:sum-m0} are also learnt, thus, the CHC is still learnt.  However, by predicting all future tokens from the initial generic task token `=' on the first time-step, no actual instance-specific supervision is being provided to these predictions via teacher-forcing, i.e., there is no conditioning on the correct arm.  Thus, these predictions behave exactly like the fully masked predictions in Fig.\@ \ref{fig2:sum-m1}.  Then, partial ground-truths are observed for future predictions for all other time-steps, and these predictions behave exactly like the partially masked prediction in Fig.\@ \ref{fig2:sum-m2}.  Note that these are repeated multiple times.  

Because alternative sequential distributions act like a combination of masking methods, they induce multihop arm reconstruction and subtask decomposition in the same way as masking.  The only difference here is that multiple predictions are conditioned from a single hidden-state instead of being conditioned on five temporally distinct hidden-states (each of which is made unique by using a new masked token and positional embeddings), however, this technical difference does not affect the induced decomposition.  This is also why one can view this as implicitly modifying the target-side input with masked tokens (and the galaxy-brain view is that we have converted target-side input supervision to target-side target supervision.).  

See \citet{gerontopoulos2025multi} for an example of a method which helps bridge the masking method that we use with the alternative sequential distributions to help demonstrate this idea.  They introduced extra tokens as input into the model to create new hidden-states from which they can make future predictions.  These extra tokens can be compared to masked tokens, except they do not replace exiting tokens.  Then this method can be compared to our alternative sequential distributions, where the former creates new hidden-states via modifying the input to the model, whereas the latter reuses hidden-states to make multi-token predictions.  This then shows how one can convert input supervision into target supervision using a structured loss.\footnote{See Appx.\@ \ref{appx:future} for a discussion of their work -- we highly recommend it to readers who have made it this.}  

Finally, alternative sequential distributions also create a unique or novel form of subtask decomposition since the same target is reused at different time-steps.  For example, it requires a subset of the steps needed to predict a target at step 6 from step 3, compared to predicting the target at step 6 from step 2 (one less step to be precise).  Consider the four (teacher-forced) predictions for target `63' indicated with the label $6'$ in Fig.\@ \ref{fig2:sum-m3}.  To predict `63' from `59' requires one hop through `1'. To predict `63' from  `6' requires two hops; one through `59' and then the same one through `1'. To predict `63' from `12' requires three hops; one through `6' and then the same two through `1' and `59' etc.

{\bf Because teacher-forcing is still being used for the alternative sequential distributions, the only non-standard aspect of these methods is that they are not strictly using the next-token prediction paradigm but rather a generalization of it -- next-\textit{tokens} prediction, if you will.}\footnote{This is more commonly called multi-token prediction, but that also includes NAR/IAR/discrete diffusion models that can predict anywhere in the sequence.}   This distinction is important for this work since the core research question of this work is whether next-token prediction is sufficient to learn the task.  Thus, these methods are not directly relevant to that research question.  However, they do provide another example of how subtask decomposition makes the task solvable, and this ties back to the core research question, as we show that next-token prediction is sufficient to learn the task, so long as some kind of subtask decomposition is given as supervision and not adulterated.\footnote{We also included these methods in this work to explain the success of other multi-token works which consider the path-star task.}

\subsection{BoW Label Smoothing and Scratchpad}\label{appx:bow-ls-and-scratchpad}

\begin{figure}
    \centering
        \begin{subfigure}[b]{0.47\textwidth}
\centering
        \includegraphics[scale=.68, trim={1.cm .2cm .2cm .2cm},clip]{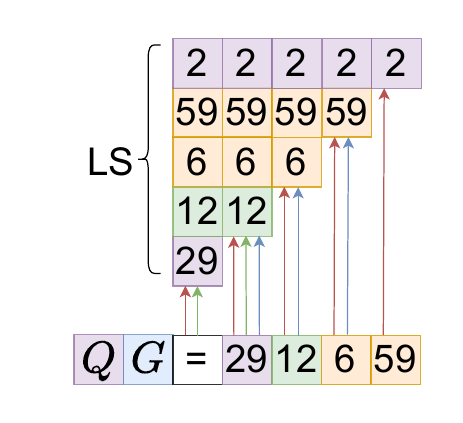}
        \caption{\textbf{Alternative sequential distribution using {\em uniform} label smoothing}.  This is the same as Fig.\@ \ref{fig2:sum-m3} except the targets are stacked and only prediction types, but not individual predictions, are illustrated for visibility.  In each step, `2' and `29' are trivially predicted via the length shortcuts and a token is trivially predicted by the CHC for all but the second time-step. The prediction of the leading node `12' must always be achieved using the forward algorithm.}\label{fig2:sum-bow1}
    \end{subfigure}
    
    \begin{subfigure}[b]{0.47\textwidth}
\centering
        \includegraphics[scale=.6, trim={1.2cm .2cm .6cm .2cm},clip]{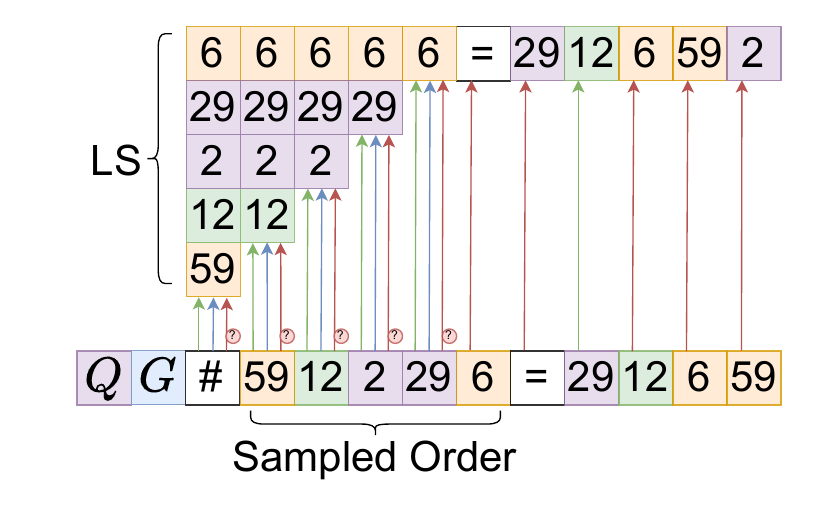}
        \caption{We sample a BoW order to be used as sequential teacher-forced input into the model.  Multiple targets are valid since any unsampled node in the arm can be the next token. Dispite having multiple targets, this is a true next-token distribution. }\label{fig2:sum-bow2}
    \end{subfigure}
\caption{BoW alternative sequential distribution and an equivalent BoW scratchpad.}\label{fig2:sum-bow}
\end{figure}

As mentioned in Sec.\@ \ref{sec::scratchpads}, the BoW scratchpad unifies the alternative sequential BoW distribution with a single next-token distribution since the next $M$ scratchpad tokens are the BoW.  {\bf That is, we can treat the multi-target BoW prediction task as a next-token sequential task via intermediate scratchpad predictions.} This is done by sampling a permutation of the target arm sequence and treating that as the ground-truth sequence for both the inputs and the targets of the scratchpad.  

We apply label smoothing to the scratchpad targets.  In theory, this does not need to be done, provided enough BoW orders are sampled.  However, given the issue concerning the large sample space of graphs, this would be a bad strategy in practice (especially since the sample space is made even larger when considering sampling a scratchpad in addition to the graph and target arm).
By sampling an input order for the scratchpad, we have incorporated order permutation into learning the task.\footnote{This relates to encoder-only non- and iterative-autoregressive models, except that these also take it one step further to create true permutation invariance via removing the causal constraint, where this attempts to learn permutation invariance via sampling permuted input sequences but keeps the causal constraint of decoder-only models.}    
{\bf Even though the scratchpad has multiple targets, this is not an alternative sequential distribution but a true next-token distribution since having multiple targets is a consequence of the order permutation, which allows for any unsampled node in the arm to be the next token.}  One can see how this becomes less unified if we weighted the labels like we do with the other alternative sequential distributions, which would result in preferring specific permutations over others.  

This is illustrated in  Fig.\@ \ref{fig2:sum-bow}, which shows the BoW label smoothing alternative distribution in Fig.\@ \ref{fig2:sum-bow1} and how this is equivalent to a BoW scratchpad in Fig.\@ \ref{fig2:sum-bow2}.  Because we allow the BoW scratchpad to be generated in any order, an easy shortcut the model could learn would be to just learn to generate the BoW as the arm in reverse order. We know this shortcut is not learnt since it would allow the model to trivially solve the task (see Tbl.\@ \ref{tbl:sp-arm-recon}).

\begin{figure}
\centering
    \includegraphics[scale=.5, trim={.6cm .2cm .6cm .2cm},clip]{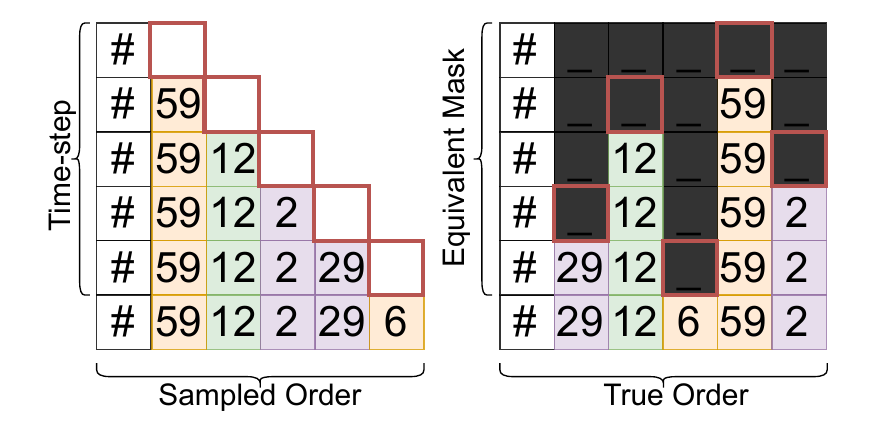}
    \caption{On the left, we show a sampled BoW order.  This is a permutation of the target sequence, and we show this across all time-steps when using autoregressive next-token prediction.  On the right, we show the sampled masks over the ground-truth sequence that are equivalent to a permuted sequence with next-token prediction at each step. Single-target predictions are indicated with red boxes, since this is true of either single- or multi-target predictions.}
    \label{fig:masking-as-perm}
\end{figure}

Now we explain how the BoW scratchpad induces subtask decomposition.  First, determining which nodes can be in the BoW requires selecting which nodes belong to the target arm, and this necessitates arm reconstruction.  Second, since the trivial reverse order is not learnt, this means that the CHC can not be solely used to achieve this, and that multihop arm reconstruction must be learnt.  To see why this is,  consider when two nodes, $u$ and $v$ are adjacent in the BoW but not in the arm, then there is no edge, $(u,\,v)$, and to predict $v$ as the next node after $u$ requires going though at least one intermediate node.   Then this will induce subtask decomposition in the same way as masking.  Fig.\@ \ref{fig:masking-as-perm} illustrates how the next token prediction of the BoW scratchpad is related to masked predictions, where, like the alternative sequence distributions, the masked input is implicit.

\subsection{Sorting Scratchpad and Graph-Reconstruction Scratchpads}\label{appx:other-scratchpads}

A sorted scratchpad induces subtask decomposition in the same way as the BoW scratchpad as it is also a randomly ordered scratchpad like the BoW scratchpad.  This is because the node IDs/values are randomly sampled; therefore, even though the input is in a canonical sorted order, it is still random with respect to the arm's true order.  {\bf Thus, masking (including non-autoregressive, iterative-autoregressive, and discrete diffusion models), alternative sequential distributions, the BoW scratchpad, and the sorted scratchpad methods all induce subtask decomposition dispute in the same way, being very different methods in terms of implementation and motivations.}

Sorting by itself does not induce subtask decomposition.  This is demonstrated with the graph reconstruction scratchpad which learn to either sort the leading or target nodes but fail to learn the task.  Here, the scratchpads do not require learning to select which nodes belong in the target arm, and hence there is no decomposition of learning to reconstruct the target arm.  Also, the sorting of either the leading or target nodes only works because these can be trivially identified using shortcuts (the leading nodes by their adjacency to the start node and the target nodes by their degree of 1).

Sorting induces node semantics.  This is because the model now learns a relationship between all the nodes ($1 < 2 < \dots < |V|$) based solely on their values and not the graph structure. This is in contrast with the other versions of the task, where there is no relationship between nodes outside of the graph structure, making the nodes -- and overall graphs -- for these versions of the task {\em semanticless}.

\subsection{Source-Side Methods}

So far, the above methods, masking, alternative distributions, and scratchpads, have all been target-side modifications.  Trees (or in graph topology in general), general queries, and general lengths\footnote{Note that general lengths are actually a graph topology modification as well, just a minor one.} are all source-side modifications.  

We illustrated how trees induce subtask decomposition in Fig.\@ \ref{fig:ps2-tree2} and how tree topology alleviates the CHC by creating new positions in the sequence which avoid the CHC in  Fig.\@ \ref{fig:ps2-tree1}.  Here, the CHC is not completely prevented. This makes sense since, in the original task, the leading node is immune to the CHC, but all others are not.  Then, by inducing subtasks that mirror the original task, we are creating new subtask leading nodes (indicated with primes) that are also immune to the CHC, but any non-leading node in the subtask is still affected by the CHC.  The general query and length methods still only have the leading node being immune, thus alleviating the CHC even less.

\begin{figure}
\centering
    \includegraphics[scale=.28, trim={.6cm .0cm .6cm 0.cm},clip]{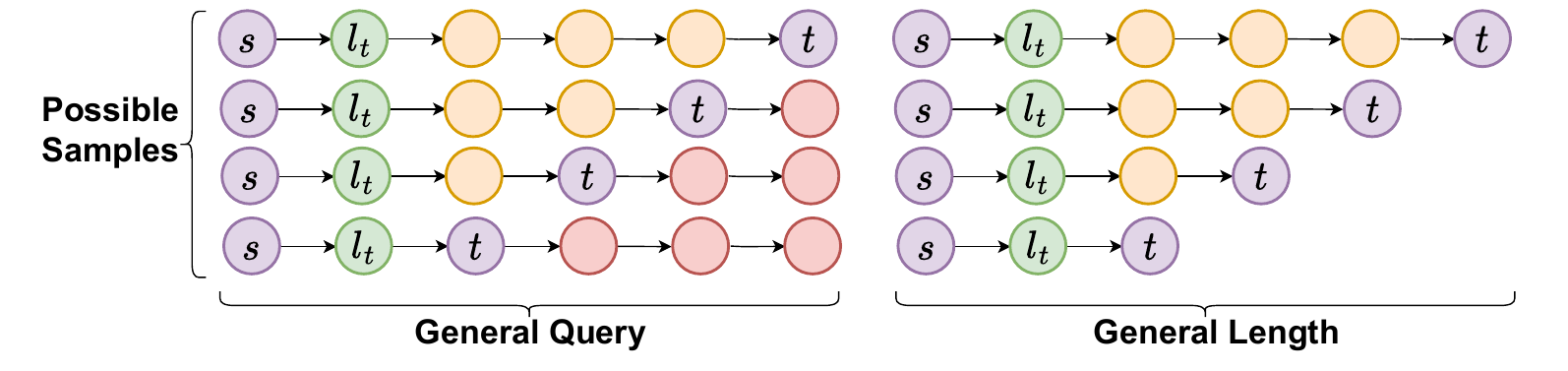}
    \caption{We show the possible general queries that can be sampled on the right and the corresponding general lengths on the left.  Red nodes succeed the given target node and present the only difference between the two versions of the task.}
    \label{fig:gen-queries-lengths}
\end{figure}

The success of autoregressive language models is due to their sample-efficient training procedure, which use the chain-rule to factorize the probability of a $T$-length sequence into $T$ conditional distributions, which are efficiently trained together in parallel for transformers.  One could imagine a poor training procedure for language models, which samples a single conditional distribution to train on per training sample.  {\bf Like masking, the subtask decomposition of trees is achieved across the sequential dimension, i.e., within a given training sample.  This contrasts with general queries and general lengths, which achieve decomposition across training samples, i.e., intra vs.\@ inter sample decomposition.  Since the CHC applies to all targets but the leading node, this is effectively sampling a single conditional distribution per training sample -- exactly like our imagined poor training procedure.  This will make general queries and general lengths less sample efficient than the other methods and may explain their limited success.}  

We illustrate the possible samples in Fig.\@ \ref{fig:gen-queries-lengths}.  {\bf General queries and general induce the same possible samples, and hence, the same subtask decomposition}.  The only difference between them is that the task using general queries also requires the model to generate the full arm with tokens coming after the target token, while general lengths do not.  However, this is irrelevant since tokens after the target can trivially be predicted via the CHC.

\section{Experiments}

We report the results of $n=5$ differently seeded trials for each experiment (except for a few where some trials prematurely stopped due to issues with our GPU cluster). 
{\bf We find a high variance for the number of iterations needed to solve the task between trials.}\footnote{This was also independently observed by \citet{saparov2025transformers}, see Appx. \ref{appx:graph-learnability}.}  This makes considering multiple trials important when considering if a given experiment is learnable or not.  Note that when we say `unlearnable', this does not mean the task is provably unlearnable, but rather a shorthand for `not found to be empirically learnt given 100 epochs'.  We abuse the term `epoch' to mean 1M samples for reporting results as there are no true epochs when using online datasets.  In the original setting using an offline dataset, there are 1M sampled examples, and the models are trained for 100 true epochs.  

Each trial can be run on a single small GPU (a 12GB Titan X works) and, at max, takes about a day to finish (if the model does not converge and runs for all 100 epochs, and depending on the specific GPU used). The main computational cost is reporting results for 1220 trials. So, while each experiment is cheap and easy to run on moderately sized hardware, the entire set of experiments is not. 

We implemented our experiments using Fairseq \citep{ott2019fairseq}.  We perform greedy decoding via Fairseq's beamsearch with a single beam.  Prefixes up to and including the special start-of-targets, `=' (or start-of-scratchpad, `\#') are force-generated. Temperature is set to 1.0, and no length penalty is applied.  We generate for a max length of 20 tokens over the ground-truth sequence length.  Despite common knowledge that beamsearch with a single beam is equivalent to greedy search, we were unsure of this due to implementation details about beamsearch, which is complex \citep{kasai-etal-2024-call}.  Both the vanilla and first-come-first-serve (used by Fairseq) variants should be equivalent with a beamsize of 1 and hence equivalent to greedy search.\footnote{As a side note, interestingly, this no longer becomes true if employing the patience hyperparameter ($> 1$) proposed in \citet{kasai-etal-2024-call}.}  This was important to verify, as we did not want the model to `cheat' by using post-hoc inference search methods in place of reasoning.





\subsection{Baseline Results}\label{appx:baseline-results}

\begin{figure}
\centering
    \includegraphics[width=1.\linewidth]{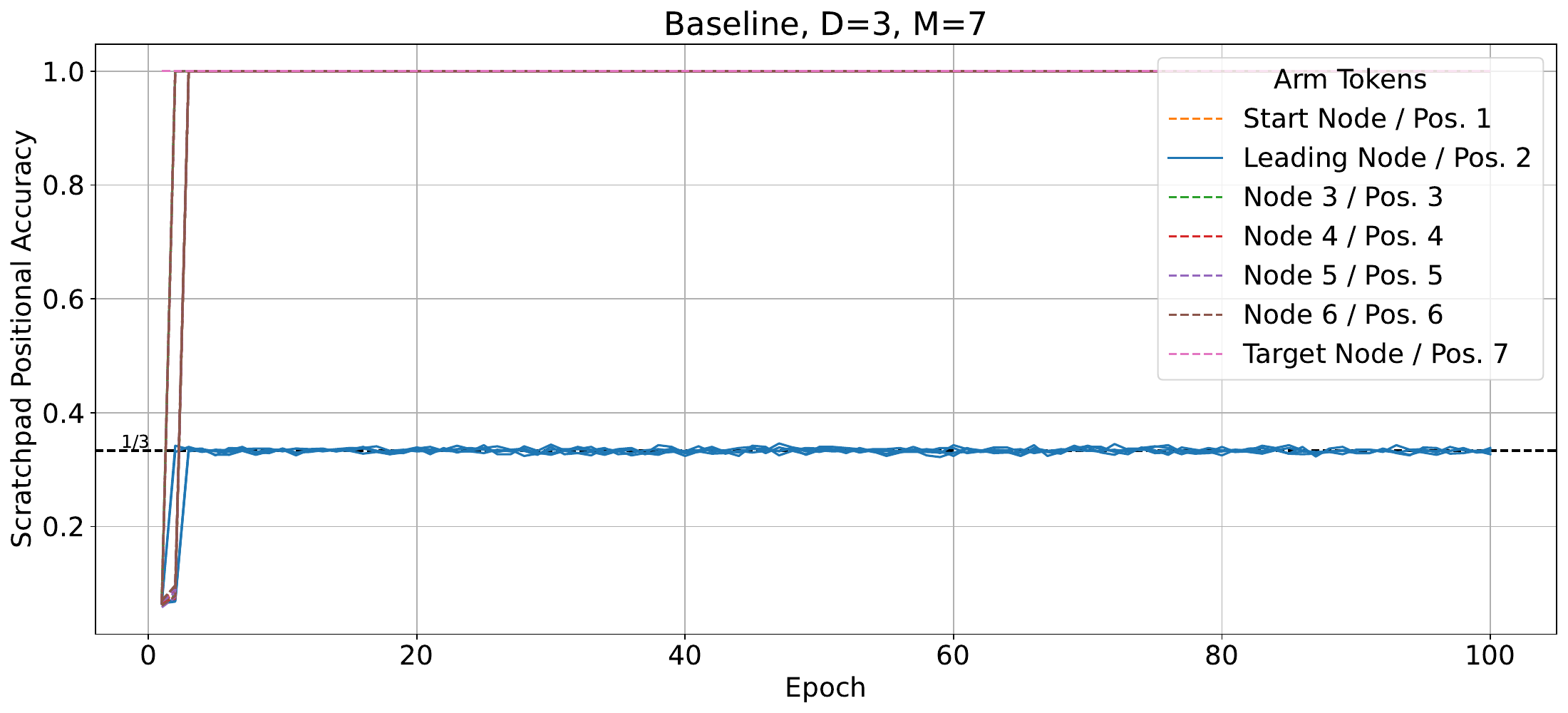}
    \caption{A baseline demonstrating multiple shortcuts used to learn all nodes except the leading node.  The start and target nodes can be immediately learnt by positional shortcuts, while nodes 3-6 are learnt by the bigram CHC.  The leading node is only predicted at chance accuracy of $1/D$.  These consider  `teacher-forced' inference which conditions on the correct sequence regardless of past inaccuracies.  We use online training so each `epoch' is 1M sampled examples.  It is over five seeded trials.}
    \label{fig:baseline1}
\end{figure}

Tbl.\@ \ref{tbl:baseline-results} provides baseline results of the path-star task (PST) using both edge- and causal-wise shuffling of $G$. We find that the PST is unlearnable, even when using online training, reducing the sample space by setting the vocabulary size to the graph size ($|V|=|G|$), and tokenizing $Q$ before $G$ \citep{frydenlund-2024-mystery}.  This is consistent with the results of \citet{bachmann2024the}.  Tbl.\@ \ref{tbl:sbaseline-results-per-run} provides a more fine-grained breakdown of the accuracy of the positional token accuracy for $l_t$ and the following two nodes for each run.  This shows that $l_t$ is predicted at $1/D$ chance, while the next two nodes are predicted with 100\% accuracy due to the CHC.  This behaviour is illustrated in Fig.\@ \ref{fig:baseline1}.  One exception is Run 3 of the exp.\@ where $D=5,\, M=7$ which fails to learn the CHC,  
{\bf demonstrating the CHC is not guaranteed to be learnt, despite its seeming simplicity.}  The start and target nodes are learnt immediately due to positional shortcuts and not because of the CHC.

\begin{figure}
\centering
    \includegraphics[width=1.\linewidth]{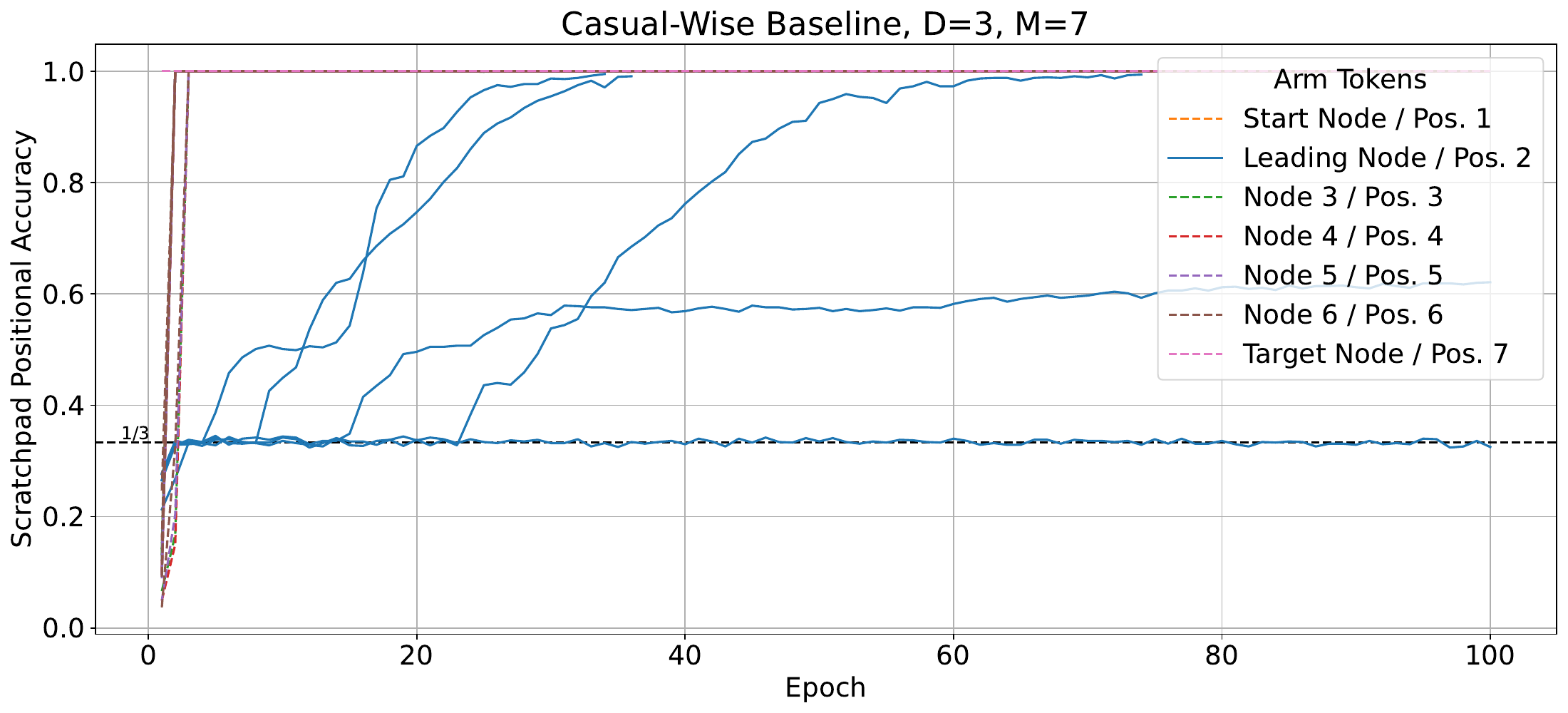}
    \caption{Using the causal-wise ordering of edges allows the task to be learnt on 3/5 trials with one run exceeding the $1/D$ baseline but not learning the task to 95\% sequential accuracy.  {\bf This is also an example showing the large variability on task success depending on the initial seed.}  Note that for this experiment the only source of randomness is in graph generation.}
    \label{fig:baseline3}
\end{figure}

\begin{figure*}
    \centering
    \includegraphics[width=.98\linewidth]{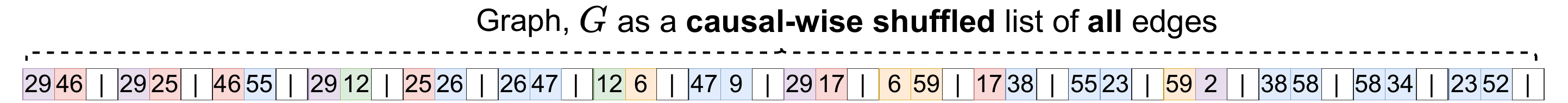}
    \caption{One potential causal-wise shuffle of the path-star graph of Fig.\@ \ref{fig:psg2}.  The arms are not contiguous, but, the order of the edges is such that a given one is always further or of equal distance from $s$ compared to all prior edges.}
    \label{fig:causal-wise-shuffle}
\end{figure*}

Casual-wise shuffling enforces that, given two edges, $(u,\, v)$ and $(v,\, w)$, the former always proceeds the latter.  This avoids the issue of learning two separate routing rules for decoder-only models.  See Fig.\@ \ref{fig:causal-wise-shuffle} for an illustration. This is achieved via a sampling procedure where an arm is sampled and the edge closest to $s$ is taken without replacement.  This sampling is done until no edges remain.

We find that using a causal-wise shuffle makes the PST learnable.  {\bf This indicates that the causal constraint accounts for some of the task's difficulty.}  However, once we consider arms with moderate length ($M=9$), the task is no longer perfectly learnt (at least within the 100 epochs provided).  Fig.\@ \ref{fig:baseline3} shows $l_t$ being fitted by in the causal-wise version of the task. This also shows that, even when the task is learnt, the CHC is still employed when solving the task, otherwise we would expect a change in the overall accuracy of the other nodes as they become predicted by the new algorithm.  

\begin{table*}[htp]
\begin{center}
\begin{tabular}{lcc|rr|rr}
                                    &         &     &  \multicolumn{2}{c}{Test-Force $R_t$} & \multicolumn{2}{|c}{Test-Gen $R_t$}  \\
 Experiment Description                           &     $D$ & $M$ &                               SR &ABB & SR & ABB  \\ \hline
    \multirow{8}{*}{\shortstack[l]{Edge-Wise\\ $|V|=|G|$, Online Training, $Q$ Before $G$}}  & 2 & 5 &  0\% &  0\%  &  0\% &  0\%  \\
                                             & 3 & 5 &  0\% &  0\%  &  0\% &  0\%  \\
                                             & 4 & 5 &  0\% &  0\%  &  0\% &  0\%  \\
                                             & 5 & 5 &  0\% &  0\%  &  0\% &  0\%  \\ \cline{2-7}
                                             & 2 & 7 &  0\% &  0\%  &  0\% &  0\%  \\
                                             & 3 & 7 &  0\% &  0\%  &  0\% &  0\%  \\
                                             & 4 & 7 &  0\% &  0\%  &  0\% &  0\%  \\
                                             & 5 & 7 &  0\% &  0\%  &  0\% &  0\%  \\  \hline
    \multirow{ 12}{*}{Causal-Wise}  & 2 & 5 & 100\% & 100\%  & 100\% & 100\%  \\
                                             & 3 & 5 & 100\% & 100\%  & 100\% & 100\%  \\
                                             & 4 & 5 & 100\% & 100\%  & 100\% & 100\%  \\
                                             & 5 & 5 & 60\% & 100\%  & 60\% & 100\%  \\ \cline{2-7}
                                             & 2 & 7 & 100\% & 100\%  & 100\% & 100\%  \\
                                             & 3 & 7 & 60\% & 80\%  & 60\% & 80\%  \\
                                             & 4 & 7 & 0\% & 20\%  & 0\% & 20\%  \\
                                             & 5 & 7 & 0\% & 40\%  & 0\% & 40\%  \\ \cline{2-7}
                                             & 2 & 9 &   40\% & 100\%  &   40\% & 100\%  \\
                                             & 3 & 9 &   0\% & 40\%  &   0\% & 40\%  \\ \cline{2-7}
                                             & 2 & 12 &   0\% & 80\%  &   0\% & 80\%  \\
                                             & 3 & 12 &   0\% & 20\%  &   0\% & 20\%  \\
\end{tabular}
\end{center}
\caption{Full baseline experiment results.  We report the {\em Success Rate} (SR) where the model predicts $> 95$\% sequential accuracy and {\em Above-Baseline} (ABB) where the model predicts $> (100/D +10)$\% sequential accuracy.  When this happens it indicates that the model can predict $l_t$ in some cases.  As such, when ABB $>$ SR, it implies that the model would have learnt the task had it been provided with more training time in these cases.  We report on the test partition using both `teacher-forced inference' which conditions on the correct sequence regardless of past inaccuracies (Test-Force $R_t$) as well as true auto-regressive generation (Test-Gen $R_t$).  In general, these provide the same results, since $l_t$ will either be learnt at $> 95$\% accuracy or not in both cases, leading to the same overall sequential accuracy.  Results are reported after 100 epochs i.e.\@ 100M training samples.}
\label{tbl:baseline-results}
\end{table*}

\begin{table*}[htp]
\begin{center}
\begin{tabular}{cc|rrr|rrr|rrr|rrr|rrr}
& & \multicolumn{3}{c}{Run 1} & \multicolumn{3}{c}{Run 2} & \multicolumn{3}{c}{Run 3} & \multicolumn{3}{c}{Run 4} & \multicolumn{3}{c}{Run 5}  \\
$D$ & $M$ &  $l_t$ & pos.\@ 2 & pos.\@ 3 &  $l_t$ & 2 & 3 &  $l_t$ & 2 & 3 &  $l_t$ & 2 & 3 &  $l_t$ & 2 & 3    \\ \hline
   2 & 5 & 50\% & {\tiny \checkmark} & {\tiny \checkmark} & 50\% & {\tiny \checkmark}& {\tiny \checkmark}  & 50\% & {\tiny \checkmark} & {\tiny \checkmark} & 50\% & {\tiny \checkmark}& {\tiny \checkmark}  & 50\% & {\tiny \checkmark} & {\tiny \checkmark}\\
   3 & 5 & 33\% & {\tiny \checkmark} & {\tiny \checkmark} & 33\% & {\tiny \checkmark}& {\tiny \checkmark}  & 33\% & {\tiny \checkmark} & {\tiny \checkmark} & 33\% & {\tiny \checkmark}& {\tiny \checkmark}  & 33\% & {\tiny \checkmark} & {\tiny \checkmark}\\
   4 & 5 & 25\% & {\tiny \checkmark} & {\tiny \checkmark} & 25\% & {\tiny \checkmark}& {\tiny \checkmark}  & 25\% & {\tiny \checkmark} & {\tiny \checkmark} & 25\% & {\tiny \checkmark}& {\tiny \checkmark}  & 25\% & {\tiny \checkmark} & {\tiny \checkmark}\\
   5 & 5 & 20\% & {\tiny \checkmark} & {\tiny \checkmark} & 20\% & {\tiny \checkmark}& {\tiny \checkmark}  & 20\% & {\tiny \checkmark} & {\tiny \checkmark} & 20\% & {\tiny \checkmark}& {\tiny \checkmark}  & 20\% & {\tiny \checkmark} & {\tiny \checkmark}\\ \hline

   2 & 7 & 50\% & {\tiny \checkmark} & {\tiny \checkmark} & 50\% & {\tiny \checkmark}& {\tiny \checkmark}  & 50\% & {\tiny \checkmark} & {\tiny \checkmark} & 50\% & {\tiny \checkmark}& {\tiny \checkmark}  & 50\% & {\tiny \checkmark} & {\tiny \checkmark}\\
   3 & 7 & 33\% & {\tiny \checkmark} & {\tiny \checkmark} & 33\% & {\tiny \checkmark}& {\tiny \checkmark}  & 33\% & {\tiny \checkmark} & {\tiny \checkmark} & 33\% & {\tiny \checkmark}& {\tiny \checkmark}  & 33\% & {\tiny \checkmark} & {\tiny \checkmark}\\
   4 & 7 & 25\% & {\tiny \checkmark} & {\tiny \checkmark} & 25\% & {\tiny \checkmark}& {\tiny \checkmark}  & 25\% & {\tiny \checkmark} & {\tiny \checkmark} & 25\% & {\tiny \checkmark}& {\tiny \checkmark}  & 25\% & {\tiny \checkmark} & {\tiny \checkmark}\\
   5 & 7 & 20\% & {\tiny \checkmark} & {\tiny \checkmark} & 20\% & {\tiny \checkmark}& {\tiny \checkmark}  & 4\% & 4\% & 4\% & 20\% & {\tiny \checkmark}& {\tiny \checkmark}  & 20\% & {\tiny \checkmark} & {\tiny \checkmark}\\
\end{tabular}
\end{center}
\caption{Training positional accuracy for $l_t$, pos.\@ 2, and pos.\@ 3 for the edge-wise baseline results in Tbl. \ref{tbl:baseline-results}. {\tiny \checkmark} indicates 100\% accuracy.  In all but a single run, the CHC is learnt for pos.\@ 2 and 3 (and all other non-leading nodes) and $l_t$ is predicted at $1/D$ chance.  
`pos.' is redacted for space for trials 2-5.}
\label{tbl:sbaseline-results-per-run}
\end{table*}

Because the PST in its original form is unlearnable (for reasons not due to regular hyperparamaters), it is impossible to hyperparameter tune it.  {\bf As such, we use the causal-wise version to determine valid hyperparameters.}  The reported results for all experiments are after having found hyperparameters using the causal-wise version of the task and the baseline results which were redone for consistency and to rule-out improper hyperparamaters causing the task to be unlearnable.


\subsection{Masking Results}\label{appx:masking-results}

We sample which tokens are masked or replaced using a either a uniform distribution over the sequence length or spanning masks. Using uniform sampled masks introduces no inductive bias since it is not dependent on any task-specific information.  Uniform sampling is also consistent with the sampling method using in the iterative-autoregressive models used by \citet{frydenlund-2024-mystery} and most discrete diffusion models.  Spanning masks sample multiple contiguous tokens to be masked. This is achieved via sampling a span length from a geometric distribution parameterized by $p$ \citep{joshi-etal-2020-spanbert}. We sample two different spanning distributions to discourage contiguous ground-truth tokens with 
$p=\{.4,\ .5\}$ for the mask spans and $p=.8$ for the ground-truth spans.  The latter means that the majority of ground-truth spans will only be a single token, which means that the CHC will not be supported in these cases.  We randomize if we start with a masking- or a ground-truth span so $l_t$ is not always masked (which we found was important).   As we are sampling based on task specific information, this is an inductive bias.  But note, if we sampled enough masks, we would expect to sample these spanning masks from the uniform distribution.  
In addition to which positions get masked, there is also how they get masked where we try both masking via a special `mask-token' (dropout) or replaced by another node in $G$ (where the choice of node is uniformly sampled with replacement).

Tbl.\@ \ref{tbl:masking-results} provides results using uniform masking via token dropout and span masking via token dropout, token replacement, and a mixture of both.  The results show that the difference between these methods are not very consistent.  We suspect this is due to the amount of noise being added to the training procedure (and we do not ensure that the same nodes are noised in the same place across the different noise types to control for this).  {\bf We also show that mixing causal-wise shuffling with masking improves the results over using either in isolation, implying that they are helping to solve different underlying issues} (the causal constraint issue with the former and subtask decomposition issue with the latter).

\begin{table*}[htp]
\begin{center}
\begin{tabular}{lcc|rr|rr}
                                    &         &     &  \multicolumn{2}{c}{Test-Force $R_t$} & \multicolumn{2}{|c}{Test-Gen $R_t$}  \\
Experiment Description                           &     $D$ & $M$ &                               SR &ABB & SR & ABB  \\ \hline
    \multirow{12}{*}{\shortstack[l]{Uniform Token Dropout}}  
                                             & 2 & 5 &  60\% &  60\%  &  60\% &  60\%  \\
                                             & 3 & 5 &  100\% &  100\%  &  100\% &  100\%  \\
                                             & 4 & 5 &  100\% &  100\%  &  100\% &  100\%  \\
                                             & 5 & 5 &  80\% &  80\%  &  80\% &  80\%  \\ \cline{2-7}
                                             & 2 & 7 &  60\% &  80\%  &  60\% &  80\%  \\
                                             & 3 & 7 &  80\% &  80\%  &  80\% &  80\%  \\
                                             & 4 & 7 &  40\% &  60\%  &  40\% &  60\%  \\
                                             & 5 & 7 &  0\% &  20\%  &  0\% &  20\%  \\  \cline{2-7}
                                             & 2 & 9 &  40\% &  60\%  &  40\% &  60\%  \\
                                             & 3 & 9 &  0\% &  20\%  &  0\% &  20\%  \\
                                             & 4 & 9 &  0\% & 0\%  &  0\% &  0\%  \\
                                             & 5 & 9 &  0\% &  0\%  &  0\% &  0\%  \\  \hline
   \multirow{13}{*}{\shortstack[l]{Span Token Dropout}}  & 2 & 5 &  100\% &  100\%  &  100\% &  100\%  \\
                                             & 3 & 5 &  100\% &  100\%  &  100\% &  100\%  \\
                                             & 4 & 5 &  100\% &  100\%  &  100\% &  100\%  \\
                                             & 5 & 5 &  60\% &  80\%  &  60\% &  80\%  \\ \cline{2-7}
                                             & 2 & 7 &  80\% &  80\%  &  80\% &  80\%  \\
                                             & 3 & 7 &  40\% &  60\%  &  40\% &  60\%  \\
                                             & 4 & 7 &  20\% &  20\%  &  20\% &  20\%  \\
                                             & 5 & 7 &  40\% &  40\%  &  40\% &  40\%  \\  \cline{2-7}
                                             & 2 & 9 &  40\% &  40\%  &  40\% &  40\%  \\
                                             & 3 & 9 &  0\% &  0\%  &  0\% &  0\%  \\
                                             & 4 & 9 &  0\% &  0\%  &  0\% &  0\%  \\
                                             & 5 & 9 &  0\% &  0\%  &  0\% &  0\%  \\  \cline{2-7}
                                             & 2 & 12 &  0\% &  0\%  &  0\% &  0\%  \\  \hline
    \multirow{2}{*}{\shortstack[l]{Causal-Wise\\ Span Token Dropout}}  & 5 & 9 & 60\% & 60\%  & 60\% & 100\%  \\
                                             & 5 & 12 & 0\% & 66\%  & 0\% & 66\%  \\ \hline
    \multirow{4}{*}{\shortstack[l]{Span Token\\ Replacement}}  & 2 & 5 &  60\% &  60\%  &  60\% &  60\%  \\
                                             & 3 & 5 &  80\% &  80\%  &  80\% &  80\%  \\
                                             & 2 & 7 &  100\% &  100\%  &  100\% &  100\%  \\
                                             & 3 & 7 &  20\% &  40\%  &  20\% &  40\%  \\ \hline
    \multirow{12}{*}{\shortstack[l]{Span Mixed Token\\ Dropout and Replacement}}  & 2 & 5 &  80\% &  80\%  &  80\% &  80\%  \\
                                             & 3 & 5 &  80\% &  80\%  &  80\% &  80\%  \\ 
                                             & 4 & 5 &  100\% &  100\%  &  100\% &  100\%  \\
                                             & 5 & 5 &  80\% &  80\%  &  80\% &  80\%  \\ \cline{2-7}
                                             & 2 & 7 &  100\% &  100\%  &  100\% &  100\%  \\
                                             & 3 & 7 &  100\% &  100\%  &  100\% &  100\%  \\ 
                                             & 4 & 7 &  0\% &  20\%  &  0\% &  20\%  \\ 
                                             & 5 & 7 &  0\% &  20\%  &  0\% &  20\%  \\ \cline{2-7}
                                             & 2 & 9 &  60\% &  60\%  &  60\% &  60\%  \\
                                             & 3 & 9 &  80\% &  80\%  &  80\% &  80\%  \\ \cline{2-7}
                                             & 2 & 12 &  20\% &  20\%  &  20\% &  20\%  \\  
                                             & 3 & 12 &  0\% &  0\%  &  0\% &  0\%  \\
   
\end{tabular}
\end{center}
\caption{Full masking experiment results.}
\label{tbl:masking-results}
\end{table*}

\subsection{Alternative Distributions Results}\label{appx:alt-distro-results}

The alternative sequential distributions
have different semantics from next-token distributions 
and break the
`distributional' semantics of natural language 
\citep{mikolov2013efficient, emerson-2020-goals}. 
Thus they may not apply to non-planning tasks.

\begin{figure}
    \centering
    \includegraphics[width=1.\linewidth]{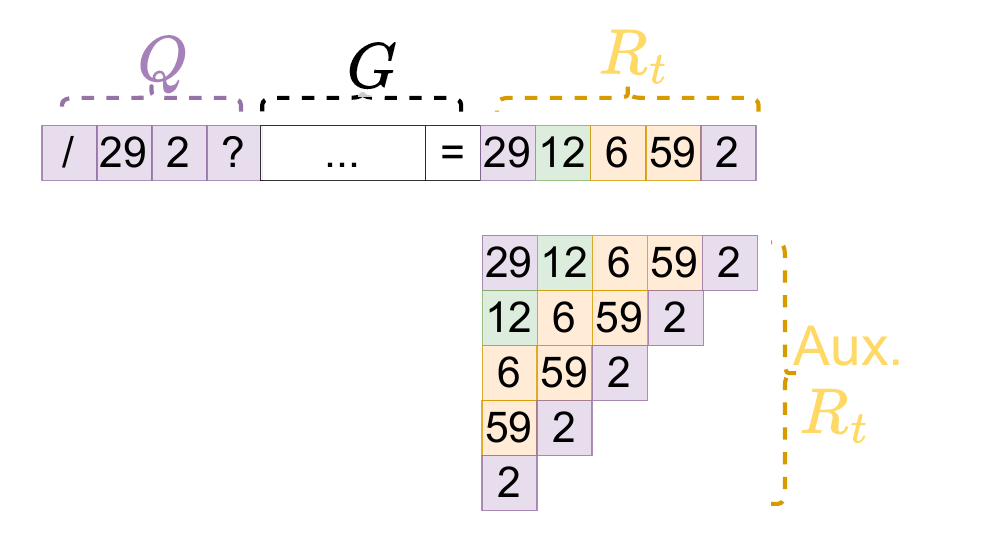}
    \caption{Auxiliary targets, Aux.\@ $R_t$, provided for training the BoW, LS, and RITF auxiliary losses.  Tokens from prior steps are removed from consideration.  Here $R_t$ provides a singular ground-truth at each step, while aux.\@ $R_t$ provides multiple for each step but the last.} 
    \label{fig:alt-distro-targets}
\end{figure}

For each alternative sequential distribution, we employ an auxiliary loss that is only used during training. Fig.\@ \ref{fig:alt-distro-targets} illustrates the extra target-side label supervision given to the model during training.  {\bf Note how this is just a replication of the original target labels with an alternative structure, thus no new information is given, but rather it is just provided in an alternative way.}  The auxiliary loss is trained in conjunction with the main loss.  Because of the change in semantics, we do not want to interfere with the main loss and the true next-token distribution.  As such, we use an interior hidden-state as $B$ instead of the final hidden-state, which supports the main loss as usual (we use the second last hidden-state).  We increase the number of layers from $L=8$ to $9$ to account for this.  This allows the auxiliary distributions to perform the same number of hops as the baseline models (see RASP constructions in \citet{frydenlund-2024-mystery}).  We do not believe that the extra computation affects comparisons between these results and other methods that use $L=8$.  

We use a monotonically decreasing stepped weighting for LS where the value between each consecutive weight is the same.  Thus the actual weighting dynamically changes depending on $M$ and the current step.  In Fig.\@ \ref{fig:alt-distro-targets} this is applied on each column of aux.\@ $R_t$ (at each step) individually.       

For RITF, we provided a partial implementation of the hinge-wise loss in Eq.\@ \ref{eq:pairwise-loss}.  This corresponds with ranking the elements in each column in Fig.\@ \ref{fig:alt-distro-targets} from highest to lowest (equivalent to the sequential order of the arm).\footnote{By removing the prior tokens from consideration, there are superficial similarities to the exclusionary procedure of Plackett-Luce, however, this is only superficial because the logits or scores change at every step here.} We used a hinge of $h=1$ and did not experiment with other values.  Note that Eq.\@ \ref{eq:pairwise-loss} is slightly ill-defined since we used score indices over the sequence length where these need to be translated to vocabulary indices in the range of $|V|$ plus the number of special tokens.  Including this would have complicated the equation to little benefit to the reader.  

In addition ranking nodes into the future, we also rank any node in $G$ not in $R_t$ lower than any node in $R_t$. We consider the entire vocabulary in practice because it is easier to calculate, however, the intuition of the inductive bias concerns nodes in $G$.  This can be done using the same calculation, 
\begin{equation}\label{eq:pairwise-loss-2}
L_{B} = \sum_{i=1}^{M}\sum_{j}\sum_{k} \max(0,\, 1 - 
(\sigma_{i}[j] - \sigma_{i}[k])), 
\end{equation}
except where $j\in R_t$ and $k \in V - R_t$ i.e.\@ over differently selected pairs from Eq.\@ \ref{eq:pairwise-loss}.  Thus the overall ranking loss factorizes as two disjoint losses, one for each inductive bias being modelled.  

Tbl.\@ \ref{tbl:alt-distro-results} provides results using the alternative distributions.  We find poor results for LS in particular.  We strongly suspect this is because each LS weight also functions as a weight on the corresponding loss term.  This means that far-future tokens will have tiny contributions to the overall loss.  Although we allow for scaling of the monotonically decreasing terms (via a temperature hyperparameter), we do not experiment with this.  It may be that doing so will result in better performance, but, we argue that using RITF instead avoids this complication.  

\begin{table*}[htp]
\begin{center}
\begin{tabular}{lcc|rr|rr}
                                    &         &     &  \multicolumn{2}{c}{Test-Force $R_t$} & \multicolumn{2}{|c}{Test-Gen $R_t$}  \\
 Experiment Description                           &     $D$ & $M$ &                               SR &ABB & SR & ABB  \\ \hline
    \multirow{16}{*}{\shortstack[l]{Uniform Label Smoothing\\ (BoW)}}  & 2 & 5 &  100\% &  100\%  &  100\% &  100\%  \\
                                             & 3 & 5 &  100\% &  100\%  &  100\% &  100\%  \\
                                             & 4 & 5 &  20\% &  60\%  &  20\% &  60\%  \\
                                             & 5 & 5 &  20\% &  60\%  &  20\% &  60\%  \\ \cline{2-7}
                                             & 2 & 7 &  60\% &  60\%  &  60\% &  60\%  \\
                                             & 3 & 7 &  100\% &  100\%  &  100\% &  100\%  \\
                                             & 4 & 7 &  0\% &  0\%  &  0\% &  0\%  \\
                                             & 5 & 7 &  0\% &  0\%  &  0\% &  0\%  \\  \cline{2-7}
                                             & 2 & 9 &  100\% &  100\%  &  100\% &  100\%  \\
                                             & 3 & 9 &  100\% &  100\%  &  100\% &  100\%  \\
                                             & 4 & 9 &  20\% &  60\%  &  20\% &  60\%  \\
                                             & 5 & 9 &  0\% &  0\%  &  0\% &  0\%  \\  \cline{2-7}
                                             & 2 & 12 &  0\% &  20\%  &  0\% &  20\%  \\  
                                             & 3 & 12 &  0\% &  0\%  &  0\% &  0\%  \\
                                             & 4 & 12 &  0\% &  0\%  &  0\% &  0\%  \\
                                             & 5 & 12 &  0\% &  0\%  &  0\% &  0\%  \\  \hline
    \multirow{12}{*}{\shortstack[l]{Monotonically Decreasing\\ Label Smoothing\\ (LS)}}  
                                            & 2 & 5 &  100\% &  100\%  &  100\% &  100\%  \\
                                             & 3 & 5 &  100\% &  100\%  &  100\% &  100\%  \\
                                             & 4 & 5 &  100\% &  100\%  &  100\% &  100\%  \\
                                             & 5 & 5 &  100\% &  100\%  &  100\% &  100\%  \\ \cline{2-7}
                                             & 2 & 7 &  100\% &  100\%  &  100\% &  100\%  \\
                                             & 3 & 7 &  100\% &  100\%  &  100\% &  100\%  \\
                                             & 4 & 7 &  40\% &  40\%  &  40\% &  40\%  \\
                                             & 5 & 7 &  0\% &  0\%  &  0\% &  0\%  \\ \cline{2-7}
                                             & 2 & 9 &  60\% &  60\%  &  60\% &  60\%  \\
                                             & 3 & 9 &  0\% &  0\%  &  0\% &  0\%  \\  
                                             & 4 & 9 &  0\% &  0\%  &  0\% &  0\%  \\
                                             & 5 & 9 &  0\% &  0\%  &  0\% &  0\%  \\ \hline
    \multirow{16}{*}{\shortstack[l]{Ranking into the Future \\ (RITF)}}  & 2 & 5 &  100\% &  100\%  &  100\% &  100\%  \\
                                             & 3 & 5 &  100\% &  100\%  &  100\% &  100\%  \\
                                             & 4 & 5 &  100\% &  100\%  &  100\% &  100\%  \\
                                             & 5 & 5 &  80\% &  80\%  &  80\% &  80\%  \\ \cline{2-7}
                                             & 2 & 7 &  100\% &  100\%  &  100\% &  100\%  \\
                                             & 3 & 7 &  100\% &  100\%  &  100\% &  100\%  \\
                                             & 4 & 7 &  80\% &  80\%  &  80\% &  80\%  \\
                                             & 5 & 7 &  60\% &  60\%  &  60\% &  60\%  \\  \cline{2-7}
                                             & 2 & 9 &  100\% &  100\%  &  100\% &  100\%  \\
                                             & 3 & 9 &  100\% &  100\%  &  100\% &  100\%  \\
                                             & 4 & 9 &  100\% &  100\%  &  100\% &  100\%  \\
                                             & 5 & 9 &  60\% &  100\%  &  60\% &  100\%  \\  \cline{2-7}
                                             & 2 & 12 &  100\% &  100\%  &  100\% &  100\%  \\
                                             & 3 & 12 &  100\% &  100\%  &  100\% &  100\%  \\
                                             & 4 & 12 &  60\% &  100\%  &  60\% &  100\%  \\
                                             & 5 & 12 &  0\% &  100\%  &  0\% &  100\%  \\  \cline{2-7}
                                             & 2 & 15 &  60\% &  100\%  &  60\% &  100\%  \\
                                             & 3 & 15 &  0\% &  100\%  &  0\% &  100\%  \\
\end{tabular}
\end{center}
\caption{Alternative sequential (future) distribution results.}
\label{tbl:alt-distro-results}
\end{table*}

\subsection{Arm Reconstruction Scratchpads Results}\label{appx:sp-arm-recon}

Tbl.\@ \ref{tbl:sp-arm-recon} shows the results for arm reconstruction scratchpads.  These are the first results that see differences between autoregressive inference and teacher-forced inference.  Sequence accuracy is evaluated independently for $R_t$ and the scratchpad ($S$) in order to support better analysis of what the model is learning.  This is why the model can get 100\% sequence accuracy for $R_t$ but less than that for the SP in the teacher-forced setting and why the autoregressive inference can differ from teacher-forced inference results.  As such, `Test-Gen $R_t$' is the statistic one should consider when determining if a given experiment actually learnt the task.  

{\bf Note that while the reverse `solution' is always 100\% accurate, it is only so because of the use of shortcuts were we can learn the reverse output via CHC, which then can be reversed.  However, this just allows the model to bypass learning any planning or graph reconstruction.  Thus while the task is `solved', it is for the wrong reasons.} 

In addition to the SPs described above, we tried one that predicts $R_t$ twice in a row to compare with the reverse SP i.e.\@ forward-forward instead of reverse-forward.  To do this we also used masking with token replacement on the SP.  This is the only experiment where we consider adding masking noise to the SP and, hence, is slightly incompatible with the others.  

\begin{table*}[htp]
\begin{center}
\begin{tabular}{lcc|rrrr|rrrr}
                                    &            & &  \multicolumn{2}{c}{Test-Force $R_t$} & \multicolumn{2}{c}{Test-Force $S$} & \multicolumn{2}{|c}{Test-Gen $R_t$} & \multicolumn{2}{c}{Test-Gen $S$} \\
    Exp.\@ Desc.\@                      & $D$ & $M$ & SR &ABB & SR & ABB & SR & ABB & SR & ABB \\ \hline
    \multirow{ 4}{*}{Reverse}  & 5 & 5 & 100\% & 100\%  & 100\% & 100\% & 100\% & 100\% & 100\% & 100\%  \\
                               & 5 & 7 & 100\% & 100\%  & 100\% & 100\% & 100\% & 100\% & 100\% & 100\%  \\
                               & 5 & 9 & 100\% & 100\%  & 100\% & 100\% & 100\% & 100\% & 100\% & 100\%  \\
                               & 5 & 12 & 100\% & 100\%  & 100\% & 100\% & 100\% & 100\% & 100\% & 100\%  \\ \hline
    \multirow{ 12}{*}{BoW}  & 2 & 5 & 100\% & 100\% & 100\% & 100\% & 100\% & 100\% & NA & NA  \\
                                     & 3 & 5 & 100\% & 100\% & 100\% & 100\% & 100\% & 100\% & NA & NA  \\
                                     & 4 & 5 & 100\% & 100\% & 100\% & 100\% & 100\% & 100\% & NA & NA  \\
                                     & 5 & 5 & 100\% & 100\% & 100\% & 100\% & 100\% & 100\% & NA & NA  \\ \cline{2-11}
                                     & 2 & 7 & 100\% & 100\% & 100\% & 100\% & 100\% & 100\% & NA & NA  \\
                                     & 3 & 7 & 100\% & 100\% & 100\% & 100\% & 100\% & 100\% & NA & NA  \\
                                     & 4 & 7 & 100\% & 100\% & 100\% & 100\% & 100\% & 100\% & NA & NA  \\
                                     & 5 & 7 & 60\% & 60\%  & 60\% & 60\% & 60\% & 60\% & NA & NA  \\ \cline{2-11}
                                     & 2 & 9 & 100\% & 100\% & 100\% & 100\% & 100\% & 100\% & NA & NA  \\
                                     & 3 & 9 & 80\% & 80\% & 80\% & 80\% & 40\% & 80\% & NA & NA  \\
                                     & 4 & 9 & 40\% & 60\% & 60\% & 60\% & 0\% & 40\% & NA & NA  \\
                                     & 5 & 9 & 40\% & 40\% & 40\% & 40\% & 0\% & 40\% & NA & NA  \\ \hline
    \multirow{16}{*}{\shortstack[l]{Sorted\\ Arm}}  & 2 & 5 & 100\% & 100\%  & 100\% & 100\% & 100\% & 100\% & 100\% & 100\%  \\
                               & 3 & 5 & 100\% & 100\%  & 100\% & 100\% & 100\% & 100\% & 100\% & 100\%  \\
                               & 4 & 5 & 100\% & 100\%  & 100\% & 100\% & 100\% & 100\% & 100\% & 100\%  \\
                               & 5 & 5 & 100\% & 100\%  & 100\% & 100\% & 100\% & 100\% & 100\% & 100\%  \\ \cline{2-11}
                               & 2 & 7 & 100\% & 100\%  & 100\% & 100\% & 100\% & 100\% & 100\% & 100\%  \\
                               & 3 & 7 & 100\% & 100\%  & 100\% & 100\% & 100\% & 100\% & 100\% & 100\%  \\
                               & 4 & 7 & 100\% & 100\%  &  60\% & 100\% &  60\% & 100\% &  60\% & 100\%  \\
                               & 5 & 7 & 100\% & 100\%  &  40\% & 100\% &  40\% & 100\% &  40\% & 100\%  \\ \cline{2-11}
                               & 2 & 9 & 100\% & 100\%  & 100\% & 100\% & 100\% & 100\% & 100\% & 100\%  \\
                               & 3 & 9 & 100\% & 100\%  &  80\% & 100\% & 100\% & 100\% & 80\% & 100\%  \\
                               & 4 & 9 & 100\% & 100\%  &   0\% & 100\% &   0\% & 100\% &  0\% & 100\%  \\
                               & 5 & 9 &  60\% &  60\%  &  20\% &  20\% &  20\% &  20\% &  20\% & 20\%  \\ \cline{2-11}
                               & 2 & 12 & 100\% & 100\%  &  40\% &  80\% &  40\% & 80\% &  40\% &  80\%  \\
                               & 3 & 12 & 100\% & 100\%  &   0\% &  40\% &   0\% & 40\% &   0\% &  40\%  \\
                               & 4 & 12 &   0\% &  20\%  &   0\% &   0\% &   0\% &  0\% &   0\% &   0\%  \\
                               & 5 & 12 &   0\% &   0\%  &   0\% &   0\% &   0\% &  0\% &   0\% &   0\%  \\ \hline

    \multirow{12}{*}{\shortstack[l]{Forward}}  & 2 & 5 & 100\% & 100\%  & 80\% & 80\% & 100\% & 80\% & 100\% & 80\%  \\
                               & 3 & 5 & 100\% & 100\%  & 100\% & 100\% & 100\% & 100\% & 100\% & 100\%  \\
                               & 4 & 5 & 100\% & 100\%  & 100\% & 100\% & 100\% & 100\% & 100\% & 100\%  \\
                               & 5 & 5 & 100\% & 100\%  & 100\% & 100\% & 100\% & 100\% & 100\% & 100\%  \\ \cline{2-11}
                               & 2 & 7 & 100\% & 100\%  & 75\% & 75\% & 75\% & 75\% & 75\% & 75\%  \\
                               & 3 & 7 & 20\% & 100\%  & 20\% & 20\% & 20\% & 20\% & 20\% & 20\%  \\
                               & 4 & 7 & 0\% & 0\%  &  0\% & 0\% &  0\% & 0\% &  0\% & 0\%  \\
                               & 5 & 7 & 0\% & 0\%  &  0\% & 0\% &  0\% & 0\% &  0\% & 0\%  \\ \cline{2-11}
                               & 2 & 9 & 60\% & 100\%  & 40\% & 60\% & 40\% & 60\% & 40\% & 80\%  \\
                               & 3 & 9 & 0\% & 0\%  &  0\% & 0\% & 0\% & 0\% & 0\% & 0\%  \\
                               & 4 & 9 & 0\% & 0\%  &   0\% & 0\% &   0\% & 0\% &  0\% & 0\%  \\
                               & 5 & 9 &  0\% &  0\%  &  0\% &  0\% &  0\% &  0\% &  0\% & 0\%  
\end{tabular}
\end{center}
\caption{Results for arm reconstruction scratchpad.  Sequence accuracy is evaluated independently for $R_t$ and the scratchpad ($S$)  We also achieved the same results for the reverse scratchpad for $D \in \{2,\,3,\,4\} \times M \in \{5,\,7,\, 9\}$.  For the BoW experiments, there are multiple correct values for each predictive scratchpad step (except for the last) and we did not implement multi-value accuracy for scratchpad generation (hence reporting `NA').}
\label{tbl:sp-arm-recon}
\end{table*}


\subsection{Graph Reconstruction Scratchpads Results}\label{appx:sp-graph-recon}

Given the path-star graph in Fig.\@ \ref{fig2:sp-gr-ps}, Figs.\@, \ref{fig2:sp-gr1}, \ref{fig2:sp-gr2}, \ref{fig2:sp-gr3}, and \ref{fig2:sp-gr4} illustrate the tokenization of the graph reconstruction scratchpads for all four combinations of leading-to-target or target-to-leading and either sorting by leading or target node values.  

\begin{figure}
\centering
    \includegraphics[width=1.\linewidth]{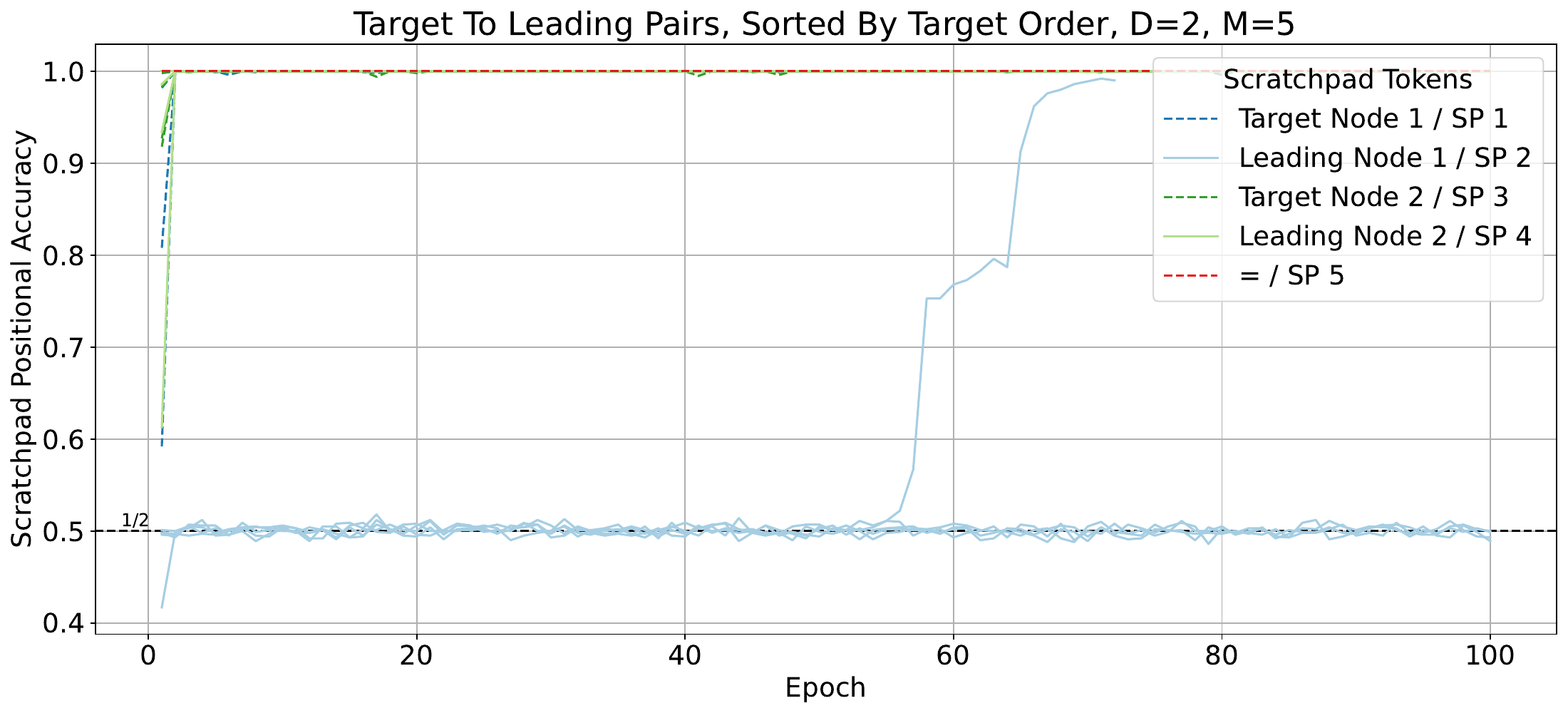}
    \caption{A $D=2$ graph-reconstruction experiment where one of the trials successfully learnt the task.}
    \label{fig:graph-recon-sp-5}
\end{figure}


\begin{figure*}[htp]
    \centering
    \begin{minipage}[s][7.87cm]{.9\columnwidth}
\vspace*{\fill}
\vspace*{\fill}
    \begin{subfigure}{1.\columnwidth}
        \includegraphics[scale=.6, trim={.65cm 0.cm .7cm 0.cm},clip]{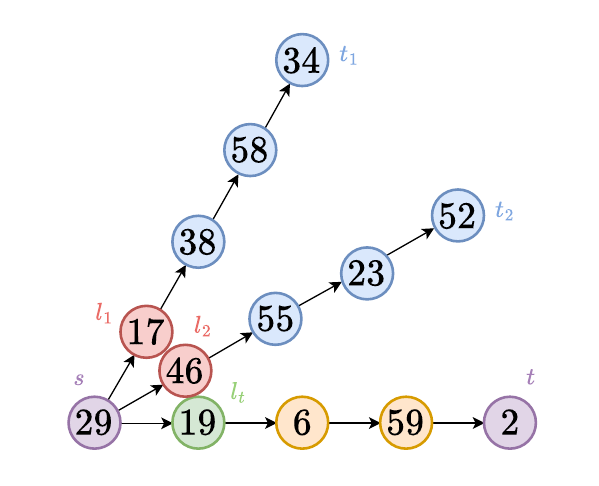}
        \caption{The path-star graph with $D=3$ and $M=5$ used when constructing  the graph reconstruction scratchpads in Figs \ref{fig2:sp-gr1}, \ref{fig2:sp-gr2}, \ref{fig2:sp-gr3}, and \ref{fig2:sp-gr4}.  }\label{fig2:sp-gr-ps}
    \end{subfigure}
\vspace*{\fill}
\end{minipage}
\begin{minipage}[s][7.87cm]{.9\columnwidth}
    \begin{subfigure}{1.\columnwidth}
        \includegraphics[scale=.35, trim={.65cm 0.cm .7cm 1.5cm},clip]{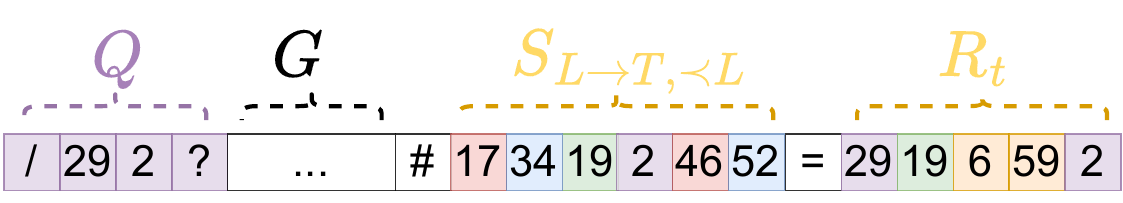}
        \caption{Leading to target pairs, sorted by leading order.}\label{fig2:sp-gr1}
    \end{subfigure}
    \begin{subfigure}{\columnwidth}
        \includegraphics[scale=.35, trim={.65cm 0.cm .7cm 1.5cm},clip]{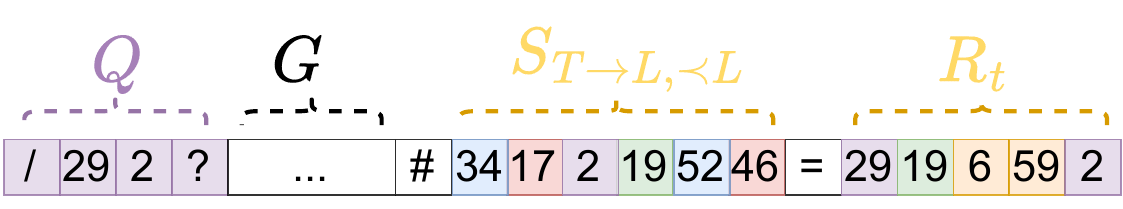}
        \caption{Target to leading pairs, sorted by leading order.}\label{fig2:sp-gr3}
    \end{subfigure}
    \begin{subfigure}{\columnwidth}
        \includegraphics[scale=.35, trim={.65cm 0.cm .7cm 1.5cm},clip]{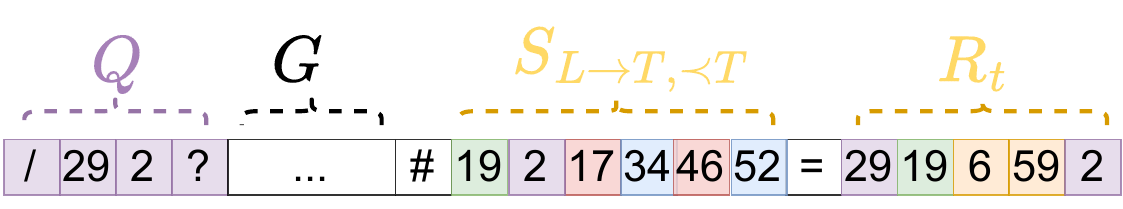}
        \caption{Leading to target pairs, sorted by target order.}\label{fig2:sp-gr2}
    \end{subfigure}
    \begin{subfigure}{\columnwidth}
        \includegraphics[scale=.35, trim={.65cm 0.cm .7cm 1.5cm},clip]{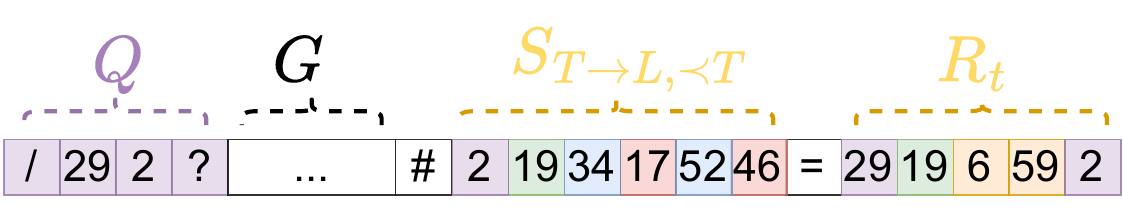}
        \caption{Target to leading pairs, sorted by target order.}\label{fig2:sp-gr4}
    \end{subfigure}
\end{minipage}
\caption{Illustration of graph reconstruction scratchpad.  Note this is slightly different from the above graph.  This is done to have more illustrative combinations of leading and target nodes after sorting.}\label{fig2:sp-graph-recon}
\end{figure*}

\begin{table*}[htp]
\begin{center}
\begin{tabular}{lcc|rrrr|rrrr}
                                    &            & &  \multicolumn{2}{c}{Test-Force $R_t$} & \multicolumn{2}{c}{Test-Force $S$} & \multicolumn{2}{|c}{Test-Gen $R_t$} & \multicolumn{2}{c}{Test-Gen $S$} \\
     Exp.\@ Desc.\@                   & $D$ & $M$ & SR &ABB & SR & ABB & SR & ABB & SR & ABB \\ \hline
    \multirow{ 4}{*}{ $S_{L \rightarrow T, \prec L}$}  & 2 & 5 & 100\% & 100\%  & 40\% & 40\% & 40\% & 40\% & 40\% & 40\%  \\
                                                       & 3 & 5 & 100\% & 100\%  & 0\% & 0\% & 0\% & 0\% & 0\% & 0\%  \\
                                                       & 4 & 5 & 100\% & 100\%  & 0\% & 0\% & 0\% & 0\% & 0\% & 0\%  \\
                                                       & 5 & 5 & 100\% & 100\%  & 0\% & 0\% & 0\% & 0\% & 0\% & 0\%  \\ \hline
    \multirow{ 4}{*}{ $S_{T \rightarrow L, \prec L}$}  & 2 & 5 & 100\% & 100\%  & 20\% & 20\% & 20\% & 20\% & 20\% & 20\%  \\
                                                       & 3 & 5 & 100\% & 100\%  & 0\% & 0\% & 0\% & 0\% & 0\% & 0\%  \\
                                                       & 4 & 5 & 100\% & 100\%  & 0\% & 0\% & 0\% & 0\% & 0\% & 0\%  \\
                                                       & 5 & 5 & 100\% & 100\%  & 0\% & 0\% & 0\% & 0\% & 0\% & 0\%  \\ \hline
    \multirow{ 4}{*}{ $S_{L \rightarrow T, \prec T}$}  & 2 & 5 & 100\% & 100\%  & 0\% & 0\% & 0\% & 0\% & 0\% & 0\%  \\
                                                       & 3 & 5 & 100\% & 100\%  & 0\% & 0\% & 0\% & 0\% & 0\% & 0\%  \\
                                                       & 4 & 5 & 100\% & 100\%  & 0\% & 0\% & 0\% & 0\% & 0\% & 0\%  \\
                                                       & 5 & 5 & 100\% & 100\%  & 0\% & 0\% & 0\% & 0\% & 0\% & 0\%  \\ \hline
    \multirow{ 4}{*}{ $S_{T \rightarrow L, \prec T}$}  & 2 & 5 & 100\% & 100\%  & 20\% & 20\% & 20\% & 20\% & 20\% & 20\%  \\
                                                       & 3 & 5 & 100\% & 100\%  & 0\% & 0\% & 0\% & 0\% & 0\% & 0\%  \\
                                                       & 4 & 5 & 100\% & 100\%  & 0\% & 0\% & 0\% & 0\% & 0\% & 0\%  \\
                                                       & 5 & 5 & 100\% & 100\%  & 0\% & 0\% & 0\% & 0\% & 0\% & 0\%  
\end{tabular}
\end{center}
\caption{Results for graph-reconstruction scratchpads.}
\label{tbl:sp-graph-recon}
\end{table*}

\begin{figure*}
\centering
\begin{subfigure}[b]{1.\textwidth}
\centering
    \includegraphics[width=.65\textwidth]{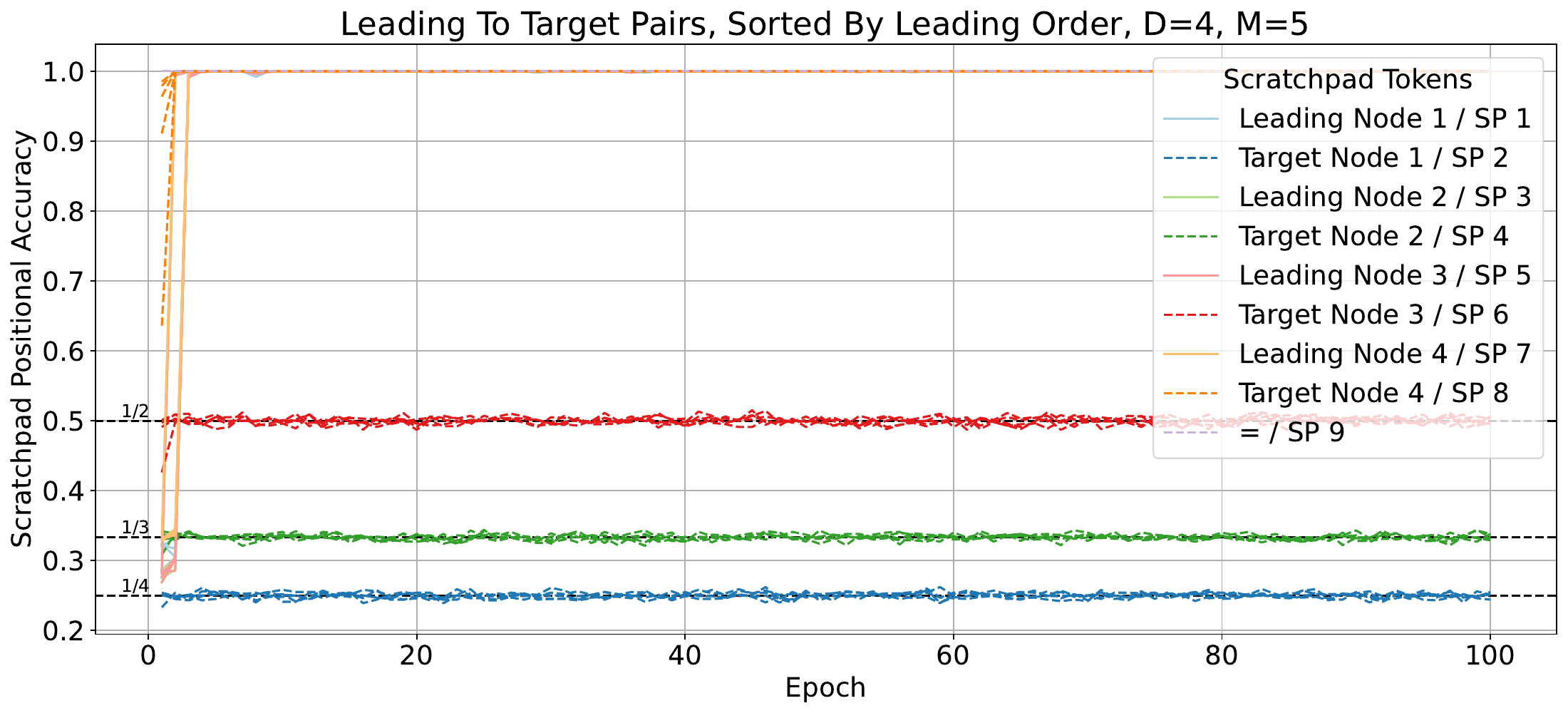}
    \caption{Leading nodes are predictable when sorting by leading order.  However, the targets corresponding to leading nodes can not be predicted even when conditioning on the correct corresponding leading node.  These then get guessed at $1/4$, $1/3$, and, $1/2$ accuracy, with the last being correctly predicted as the only remaining target.  Each plot consists of 5 differently seeded experiments.  Note that colours correspond to leading/target index and not scratchpad (SP) index i.e.\@ the sort order not the sequential order.  Thus the colours are consistent across figures.}
    \label{fig:graph-recon-sp-1}
\end{subfigure}
\begin{subfigure}[b]{1.\textwidth}
\centering
    \includegraphics[width=.65\linewidth]{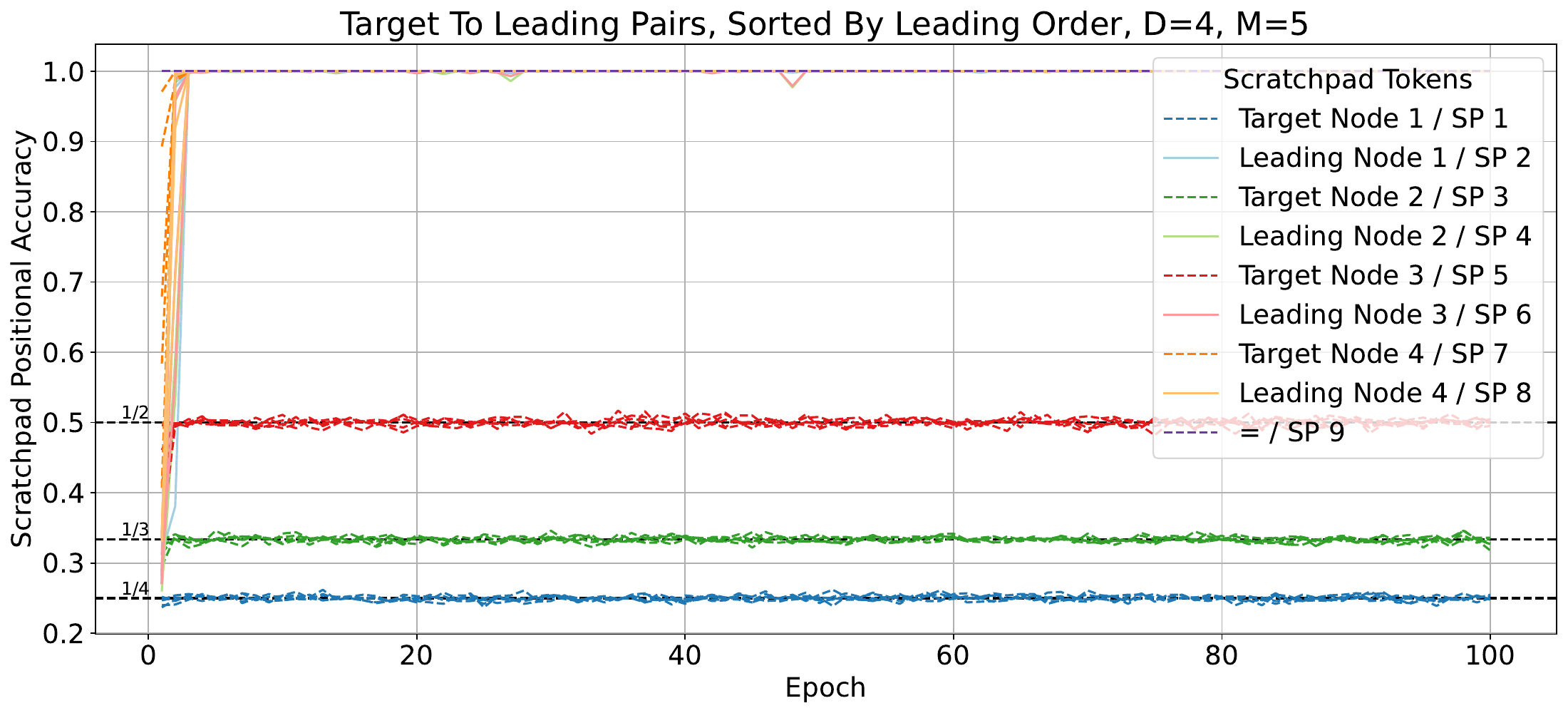}
    \caption{Leading nodes are still predictable when sorting by leading order, even when following incorrect target nodes in the scratchpad.}
    \label{fig:graph-recon-sp-2}
\end{subfigure}
\begin{subfigure}[b]{1.\textwidth}
\centering
    \includegraphics[width=.65\linewidth]{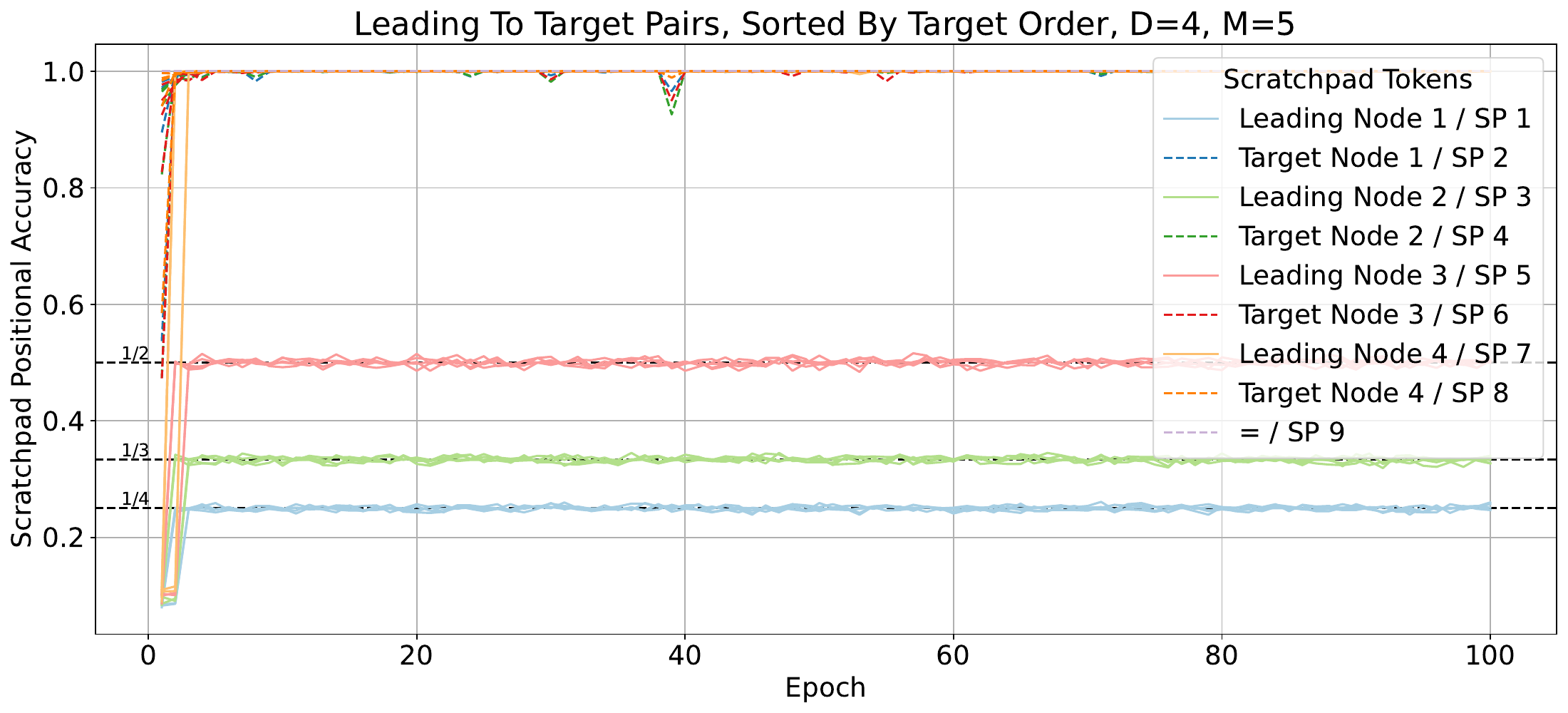}
    \caption{Target nodes are predictable when sorting by target order, even when following incorrect leading nodes in the scratchpad.}
    \label{fig:graph-recon-sp-3}
\end{subfigure}
\begin{subfigure}[b]{1.\textwidth}
\centering
    \includegraphics[width=.65\linewidth]{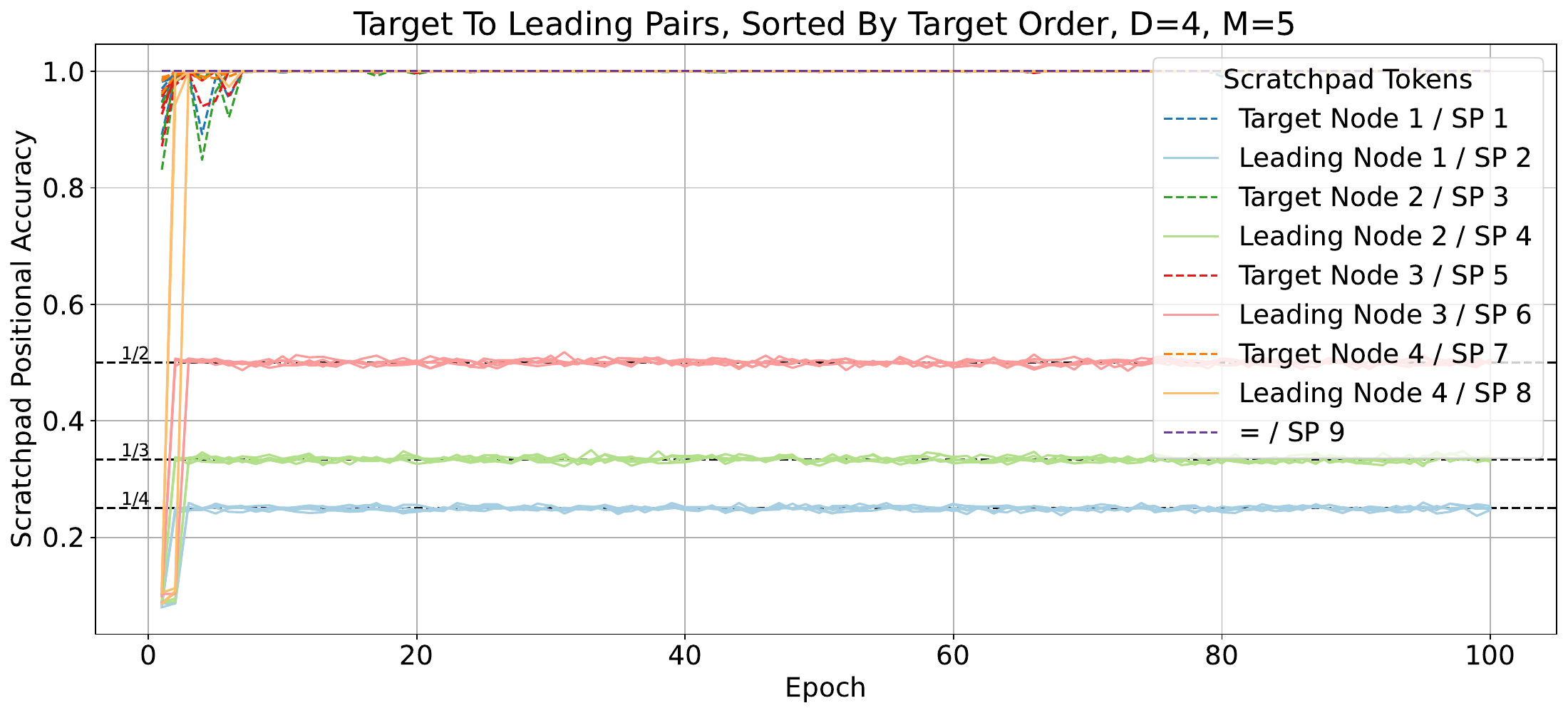}
    \caption{Target nodes are still predictable when sorting by target order, however, the correct leading node can not be predicted even when conditioning on the correct corresponding target node.}
    \label{fig:graph-recon-sp-4}
\end{subfigure}
\caption{Validation set accuracy of the scratch pad tokens across training.  These results consider `teacher-forced' inference which conditions on the correct sequence regardless of past inaccuracies.}\label{fig2:sp-graph-recon-plots} 
\end{figure*}

Tbl.\@ \ref{tbl:sp-graph-recon} shows the results for graph reconstruction scratchpads. This shows that only 4 trials succeeded in learning the task (again, `Test-Gen $R_t$' is the statistic to consider for success).  

Results for the experiments where $D=4$ and $M=5$ are plotted across training in Figs.\@ \ref{fig:graph-recon-sp-1}, \ref{fig:graph-recon-sp-2}, \ref{fig:graph-recon-sp-3}, and \ref{fig:graph-recon-sp-4}.  Figs.\@ \ref{fig:graph-recon-sp-1} and \ref{fig:graph-recon-sp-2}  demonstrate that sorting by leading node leads to all leading nodes being correctly predicted, regardless of whether the leading node precedes or succeeds the corresponding target node.  Figs.\@ \ref{fig:graph-recon-sp-3} and \ref{fig:graph-recon-sp-4} show the inverse of this where sorting on the target node leads to all target nodes being correctly predicted.  Here we see a consistent pattern where the corresponding non-sorted nodes fail to be predicted above chance. Note how once the model conditions on one of these nodes, it removes it from consideration in the next prediction, hence the first one fails at $1/D=1/4$ chance, and the next at $1/(D-1) = 1/3$ chance etc.  

Consider the two cases where the leading node precedes the target node and the arms are sorted by leading node value in Fig.\@ \ref{fig:graph-recon-sp-1} and where the target node precedes the leading node and the arms are sorted by target node value in Fig.\@ \ref{fig:graph-recon-sp-4}. {\bf This indicates that the model can correctly predict and thus condition on the correct preceding node but fails to predict the corresponding target or leading node in the next prediction even though the path between them is deterministic.  This is also an instance where, in trying to solve the problem, we introduce alternative shortcuts which also (seem to) prevent learning and shows that one needs to be careful when adding extra supervision via scratchpads to avoid adulteration.}    

Note that had we given the full arms instead of just pairs of nodes as the scratchpad, the task should become learnale, though for incorrect reasons, where instead of learning planning or arm reconstruction, the model should be able to use the sorting shortcuts in conjunction with the CHC to get 100\% accuracy.  That is, the version we present here is designed to avoid adulteration, even if that would actually lead to `succeeding' on the task.

\subsection{Query Results}\label{appx:query-results}

For the query subsets method, we use a random subset of $R_t$ as the query nodes in addition to the start node.  These are in random order.  The query is then padded out to be of length $M$ with the padding tokens coming after the observed query nodes.  This is to avoid introducing dynamic sequence lengths which would be a confounding factor when comparing against the original single target query results.  During evaluation, only the final node $t$ is given but the query is still padded to length $M$.   

For the general single target method, all nodes in $R_t$, with the exception of $s$, are considered with uniform probability.  Again only the final node is used during evaluation.

\begin{table*}[htp]
\begin{center}
\begin{tabular}{lcc|rr|rr}
                                    &         &     &  \multicolumn{2}{c}{Test-Force $R_t$} & \multicolumn{2}{|c}{Test-Gen $R_t$}  \\
 Experiment Description                           &     $D$ & $M$ &                               SR &ABB & SR & ABB  \\ \hline
    \multirow{ 9}{*}{Query Subset (Padded)}  & 2 & 5 & 100\% & 100\%  & 100\% & 100\%  \\
                                             & 3 & 5 &  75\% &  75\%  &  75\% &  75\%  \\
                                             & 4 & 5 &  40\% &  40\%  &  40\% &  40\%  \\
                                             & 5 & 5 &  80\% &  80\%  &  80\% &  80\%  \\ \cline{2-7}
                                             & 2 & 7 &  60\% & 100\%  &  60\% &  100\%  \\
                                             & 3 & 7 &   0\% &   0\%  &   0\% &    0\%  \\
                                             & 4 & 7 &   0\% &   0\%  &   0\% &    0\%  \\
                                             & 5 & 7 &   0\% &   0\%  &   0\% &    0\%  \\ \cline{2-7}
                                             & 2 & 9 &  20\% & 100\%  &  20\% &  100\%  \\ \hline
    \multirow{ 8}{*}{General Single Target}  & 2 & 5 & 100\% & 100\%  & 100\% & 100\%  \\
                                             & 3 & 5 & 100\% & 100\%  & 100\% & 100\%  \\
                                             & 4 & 5 &  80\% &  80\%  &  80\% &  80\%  \\
                                             & 5 & 5 &  40\% &  80\%  &  40\% &  80\%  \\ \cline{2-7}
                                             & 2 & 7 & 100\% & 100\%  & 100\% & 100\%  \\
                                             & 3 & 7 &  80\% & 100\%  &  80\% & 100\%  \\
                                             & 4 & 7 &  20\% &  60\%  &  20\% &  60\%  \\
                                             & 5 & 7 &  40\% &  40\%  &  40\% &  40\%  \\ \hline
    \multirow{ 4}{*}{\shortstack[l]{General Single Target (Original Setting)\\ $|V|$=100, Offline Training, $Q$ After $G$ }}  & 2 & 5 &  20\% &  20\%  &  20\% &  20\%  \\
                                     & 3 & 5 &  20\% &  20\%  &  20\% &  20\%  \\
                                     & 4 & 5 &  20\% &  60\%  &  20\% &  60\%  \\
                                     & 5 & 5 &   0\% &   0\%  &   0\% &   0\% 
\end{tabular}
\end{center}
\caption{Results for alternative query methods.  All results are evaluated with the query being the final node only.}
\label{tbl:query-results}
\end{table*}

Tbl.\@ \ref{tbl:query-results} shows the results of using general queries.  Only 3/20 trials succeeded in learning the task with a general target in the original setting (which uses $|V|$=100, Offline Training, and placing $Q$ after $G$) compared 16/20 successful corresponding trials in the new setting (which uses $|V|=|G|$, online Training, and placing $Q$ before $G$).  This indicates that such a finding may be easy to miss.  However, as discussed above there is also the issue of hyperparameter-tuning a model which fails to learn a task.  One can not hyperparameter tune models on the unadulterated path-star task in its original form as it doesn't learn above chance.  {\bf Thus another explanation for why this result might be hard to find, is that, without first finding working models (in our case using the causal-wise shuffling), we may not have properly set the hyperparameters needed for finding successful trials.}

\subsection{Tree Results}\label{appx:tree-results}

Tbl.\@ \ref{tbl:tree-results} shows the results of the tree experiments.  During training, we intermix sampling trees and path-star graphs with the latter being 10\% of the training examples.  This was done due to a length issue where trees can only make strictly shorter paths than the original path task.  

We generate $D$-ary trees by considering branching at probabilities 0.3, 0.4, 0.2, and 0.1 for no branching, branching with 2 children, 3 children, and 4 children respectively.  In any branching case, the remaining nodes are equally divided into each new subtree.  This is repeated recursively until all nodes in $R_t$ are consumed. We generate split trees using a 0.5 split probability and the remaining nodes are equally divided into each new subtree.  The `left' subtree is just a path while the `right' subtree repeats this process recursively. 

\begin{table*}[htp]
\begin{center}
\begin{tabular}{lcc|rr|rr}
                                    &         &     &  \multicolumn{2}{c}{Test-Force $R_t$} & \multicolumn{2}{|c}{Test-Gen $R_t$}  \\
 Experiment Description                           &     $D$ & $M$ &                               SR &ABB & SR & ABB  \\ \hline
    \multirow{ 9}{*}{$D$-ary Trees}  & 2 & 5 & 100\% & 100\%  & 100\% & 100\%  \\
                                             & 3 & 5 &  60\% &  100\%  &  60\% &  100\%  \\
                                             & 4 & 5 &  0\% &  100\%  &  0\% &  100\%  \\
                                             & 5 & 5 &  0\% &  100\%  &  0\% &  100\%  \\ \cline{2-7}
                                             & 2 & 7 &  0\% & 80\%  &  0\% &  80\%  \\
                                             & 3 & 7 &   20\% &   100\%  &   20\% &    100\%  \\
                                             & 4 & 7 &   0\% &   100\%  &   0\% &    100\%  \\
                                             & 5 & 7 &   0\% &   60\%  &   0\% &    60\%  \\ \hline
    \multirow{17}{*}{Split Trees}  & 2 & 5 & 100\% & 100\%  & 100\% & 100\%  \\
                                   & 3 & 5 &  100\% & 100\%  & 100\% & 100\%  \\
                                   & 4 & 5 &  100\% & 100\%  & 100\% & 100\%  \\
                                   & 5 & 5 &  100\% & 100\%  & 100\% & 100\%  \\ \cline{2-7}
                                   & 2 & 7 &  60\% & 100\%  & 60\% & 100\%  \\
                                   & 3 & 7 &  100\% & 100\%  & 100\% & 100\%  \\
                                   & 4 & 7 &  20\% & 100\%  & 20\% & 100\%  \\
                                   & 5 & 7 &  60\% & 100\%  & 60\% & 100\%  \\ \cline{2-7}
                                   & 2 & 9 &  80\% & 100\%  & 80\% & 100\%  \\
                                   & 3 & 9 &  0\% & 40\%  & 0\% & 40\%  \\
                                   & 4 & 9 &  0\% & 20\%  & 0\% & 20\%  \\
                                   & 5 & 9 &  0\% & 0\%  & 0\% & 0\%  \\ \cline{2-7}
                                   & 2 & 12 &  0\% & 20\%  & 0\% & 20\%  \\
                                   & 3 & 12 &  0\% & 20\%  & 0\% & 20\%  \\
                                   & 4 & 12 &  0\% & 0\%  & 0\% & 0\%  \\
                                   & 5 & 12 &  0\% & 0\%  & 0\% & 0\%  \\ \cline{2-7}
                                   & 2 & 15 &  0\% & 20\%  & 0\% & 20\% 
\end{tabular}
\end{center}
\caption{Results for tree methods. }
\label{tbl:tree-results}
\end{table*}

\subsection{Training on Multiple Lengths and/or Degrees Results}\label{appx:multi-results}

Tbl.\@ \ref{tbl:multi-results} shows the results for training using a sampled $M$ and/or sampled $D$.  All values are uniformly sampled.      


\begin{table*}[htp]
\begin{center}
\begin{tabular}{lcc|cc|cc}
                                    &        \multicolumn{2}{c}{Trained On}     &  \multicolumn{2}{c}{Test-Force $R_t$} & \multicolumn{2}{|c}{Test-Gen $R_t$}  \\
 Experiment Description                           &     $D$ & $M$ &                               SR &ABB & SR & ABB  \\ \hline
    \multirow{14}{*}{\shortstack[l]{Multi.\@ $M$}}   & 2 & [2-5] & 100\% & 100\%  & 100\% & 100\%  \\
                                             & 3 & [2-5] & 100\% & 100\%  & 100\% & 100\%  \\
                                             & 4 & [2-5] & 100\% & 100\%  & 100\% & 100\%  \\
                                             & 5 & [2-5] & 100\% & 100\%  & 100\% & 100\%  \\ \cline{2-7}
                                             & 2 & [2-7] &  60\% &  60\%  &  60\% &  60\%  \\
                                             & 3 & [2-7] &  20\% &  80\%  &  20\% &  80\%  \\
                                             & 4 & [2-7] &   0\% &  60\%  &   0\% &  60\%  \\
                                             & 5 & [2-7] &   0\% & 100\%  &   0\% & 100\%  \\ \cline{2-7}
                                             & 2 & [2-9] &  40\% &  40\%  &  40\% &  40\%  \\
                                             & 3 & [2-9] &  20\% &  40\%  &  20\% &  40\%  \\
                                             & 4 & [2-9] &   0\% &  20\%  &   0\% &  20\%  \\
                                             & 5 & [2-9] &   0\% &  20\%  &   0\% &  20\%  \\ \cline{2-7}
                                             & 2 & [2-12] &  0\% &  20\%  &   0\% &  20\%  \\ \hline
    \multirow{3}{*}{\shortstack[l]{Multi.\@ $D$}}   & [2-3] & 5 &   0\% &    NA  &  0\% & NA  \\
                                              & [2-4] & 5 &   0\% &    NA  &  0\% & NA  \\
                                              & [2-5] & 5 &   0\% &   NA  &  0\% & NA  \\ \hline
    \multirow{4}{*}{\shortstack[l]{Multi.\@ $M$ with\\ Multi.\@ $D$}} & [2-3] & [2-9] &   60\% &  NA  &  60\% &  NA  \\
                                             & [2-4] & [2-9] &   20\% &  NA  &   20\% &  NA  \\
                                             & [2-5] & [2-9] &   0\% &  NA  &  0\% &  NA  \\ \cline{2-7}
                                             & [2-3] & [2-12] &  0\% &  NA  &   0\% &  NA  \\ \hline
    \multirow{5}{*}{\shortstack[l]{Multi.\@ $M$ with\\ General Single Target}} & 2 & [2-9] &  60\% &  100\%  &  60\% &  100\%  \\
                                             & 3 & [2-9] &   0\% &  100\%  &  0\% &  100\%  \\
                                             & 4 & [2-9] &   20\% &  80\%  &   20\% &  80\%  \\
                                             & 5 & [2-9] &   80\% &  100\%  &  80\% &  100\%  \\ \cline{2-7}
                                             & 3 & [2-12] &  20\% &  80\%  &   20\% &  80\% 
\end{tabular}
\end{center}
\caption{ All results are evaluated with the query being only the final node in the arm.  We sample both $M$ and $D$ during the evaluation (where applicable).  The above baseline (ABB) statistic does not work when considering multiple $D$ values as it depends on a single $D$ value (hence `NA').} 
\label{tbl:multi-results}
\end{table*}













\section{Related Work}\label{appx:related-work}


There are extensive prior works given that, a), the path-star task questions the fundamental sufficiency of next-token prediction for planning tasks, b), the presented solutions vary widely in terms of methodology, and, c), we provide theoretical insights into the task. As such, this is not an exhaustive review (and still reads like `A House of Leaves' \citep{danielewski2000house})\footnote{See `The \href{https://en.wikipedia.org/wiki/House_of_Leaves\#Format}{\textcolor{blue}{House}} Is Turing Complete Under Assumptions' in Transactions of \textcolor{blue}{House} Mathematics, 2035).  To be slightly more serious, this bloated related works was done for our thesis and it would be nice if more people than just our committee ever saw it, so we included it in case others find it useful.}.  
We also point the reader to the substantial review given in \citet{bachmann2024the}.

\subsection{Large Language Model (LLMs)}\label{appx:LLMs}


LLMs have become the ubiquitous model for solving NLP tasks \citep{NEURIPS2020_1457c0d6, minaee2024large, matarazzo2025survey}.  Their abilities are assessed under various settings and methods.  Zero-shot evaluation queries an LLM with a single direct question. 
This can be enhanced with various prompting methods which prepend additional text to the query \citep{qiao-etal-2023-reasoning, schulhoff2024prompt}.  This leads to few-shot prompting which uses supervised exemplars of question-answer pairs, enabling in-context learning \citep{NEURIPS2020_1457c0d6, dong-etal-2024-survey}.  These methods are training-free as they do not modify the model's parameters.  Alternatively, fine-tuning can be performed \citep{han2024parameter, zhang2023instruction}.     

Methods like chain-of-thought (CoT) and scratchpads elicit LLMs to generate multiple reasoning steps before generating an answer \citep{nye2022show, wei2022chain, chu-etal-2024-navigate}.
This is achieved via additional prompt supervision. 
CoT can be done in the zero-shot setting with the generic prompt `Let’s think step by step' \citep{kojima2022large}.  For graphical tasks, zero-shot prompts include `Let's construct a graph with the nodes and edges first' and  `We can use a Depth-First Search (DFS) algorithm' \citep{wang2023can}. In the few-shot setting, prompts can provide a step-by-step decomposition of the task.  For example `Let's run depth-first search (DFS) step by step. Visit node 0. Neighors of node 0: [3, 6]. Visit node 6. Neighors of node 6: [3, 0]. ...' \citep[sic.\@,][]{luo2024graphinstruct}.\footnote{CoT and scratchpads use similar methods and were introduced simultaneously.  `CoT' is most commonly used in the training-free setting, whereas `scratchpads' generally implies a training setting and supervised decomposition.}

\citet{sprague2025to} found that CoT is most beneficial for symbolic tasks.  \citet{wang2024chainofthought} showed that unprompted LLMs still perform a CoT-like reasoning in non-top scoring beams during beamsearch which implies task decomposition is done by LLMs given a proper search method.

\subsubsection{Reasoning and Planning}\label{appx:LLM-reasoning}

While LLMs were originally designed for use on natural language tasks, it has become common to use LLMs as general predictive computation models and to apply them to reasoning tasks \citep{huang-chang-2023-towards, bubeck2023sparks, zhao2023large, openai-o1}, including math \citep{rabe2021mathematical, zhang2022unveiling}, puzzles \citep{shah2024causal, stechly2025on}, code generation \citep{zan-etal-2023-large, jiang2024survey}, question answering \citep{geva-etal-2021-aristotle, kamalloo-etal-2023-evaluating, ding-etal-2024-knowledge}, abstract pattern matching \citep{
chollet2024arc, openai-o3-arc}, graphs (see Appx. \ref{appx:LLM-graphs}), and planning \citep{zhao2023large, valmeekam2023on, plaat2024reasoning, stechly2025on, kang-etal-2024-empirical}.  Planning and reasoning are closely linked, with planning being a kind of reasoning that achieves a desired goal after a series of actions thus requiring sequential decision-making \citep{kang-etal-2024-empirical}.  

It has been found that LLMs struggle to solve various reasoning tasks \citep{rae2021scaling, han2022human, zhang-etal-2023-causal, ruis2023large, creswell2023selectioninference, balepur-etal-2024-easy, mirzadeh2025gsmsymbolic, jiang-etal-2024-peek, bian-etal-2024-chatgpt} including planning \citep{bubeck2023sparks, valmeekam2023planbench, valmeekam2023on, stechly2025on, plaat2024reasoning, kambhampati2024position}.  \citet{huang2024chasing} found that fine-tuning on planning tasks does not lead to good out-of-distribution (OOD) performance.   These results can be improved using various heuristics and strong search methods \citep{yao2023tree, valmeekam2023on,  creswell2023selectioninference, stechly2025on, plaat2024reasoning, huang2024chasing}.   \citet{hao-etal-2023-reasoning} explored the need for LLMs to represent planning states explicitly. 
They experiment with both easy and hard problems after observing that LLMs can fail on tasks that humans view as easy. 
\citet{kambhampati2024position} argued that LLMs by themselves can not plan, but can when provided with auxiliary models which verify generated plans.  
This poor performance has led to LLMs being pretrained specifically for reasoning tasks \citep{openai-o1}\footnote{Marketed under the name `Large Reasoning Models' 
\citep{valmeekam2024llms, zhao2024marco}.}, which have been shown to outperform other LLMs on reasoning and graphical tasks \citep{valmeekam2024llms, tang2025grapharena}.

For the path-star task, the reasoning task is choosing the correct leading node, and this requires planning to achieve. \citet{bachmann2024the} put forth the argument that LMs failing to learn the path-star task indicates a fundamental inability to learn simple planning tasks via next-token prediction, implying that the poor planning abilities of LLMs may stem from being trained via next-token prediction.  {\bf We find that the core difficulty of the path-star task does not concern planning.}  Note, while we argue that planning is not the core difficulty, planning and reasoning often require multihop reasoning.  This is highly related to task decomposition where each hop is the same operation.  Thus our decomposition findings may be of relevance to other reasoning tasks.     

{\bf We also believe that the kinds of adulteration we have described would have a small impact on the above LLMs.  However, any symbolic tasks where next-tokens can be directly inferred via prior tokens, and are trained to do so, will be at risk of adulteration. This issue may become more common due to the recent interest in pretraining models on reasoning tasks.}  We discuss this further in Appx.\@ \ref{appx:LLM-graphs}.  It is unclear if in-context learning will induce the same kind of shortcuts like CHC as training, however, \citet{khona2024toward} showed a simplicity bias for in-context learning, which they point out is related to shortcut learning (see Appx.\@ \ref{appx:spurious}).

 We use small LMs.  The reasoning abilities of LLMs are considered an emergent property \citep{huang-chang-2023-towards}, though this may be an artifact of using discontinuous evaluations \citep{schaeffer2023are}. \citet{bi2024enhancing} used knowledge distillation to generate chain-of-thought/scratchpad supervision to fine-tune small language models.  \citet{lee-etal-2024-small} did the opposite of this where a small LM was used to guide the generation of a large one.

 \citet{lin2025reasoning} studies the effect of restricting training to just predicting `critical tokens' instead of using full next-token prediction on reasoning tasks.  They find that full next-token prediction works better for pertaining but restricted training can be more efficient for finetuning.  {\bf Interestingly, the training procedure of the path-star task can be viewed as such a restriction since next-token prediction is only performed on the target-side.}  This is because next-token prediction on $G$ and $Q$ is invalid as both must be given information, i.e.\@, you can not predict the next token in the graph without first knowing the graph.  

\subsubsection{LLMs on Graphs}\label{appx:LLM-graphs}

Reasoning tasks have an implicit graphical structure   \citep[][{\em inter alia}]{dziri2023faith, creswell2023selectioninference, Xu_Khalil_Sanner_2023, hao-etal-2023-reasoning, zhao2023large, wu2024can, khona2024toward, zhu2024dyval,  kang-etal-2024-empirical, stechly2025on, han2025reasoning}.  In general, the outputs of any deterministic algorithm decompose into a series of reasoning/computation steps forming a DAG \citep{dziri2023faith, khona2024toward}. 

These tasks can be specified in natural language \citep[][{\em inter alia}]{tandon-etal-2019-wiqa, madaan-etal-2021-give, saha-etal-2021-explagraphs, sakaguchi-etal-2021-proscript-partially, huang2022language, valmeekam2023on, zhang-etal-2023-causal, ding-etal-2024-knowledge, huang2024can}.  
This 
introduces a subtask of mapping language to graph  \citep{wang2023can, fatemi2024talk}.  \citet{madaan-etal-2022-language} found that LLMs that generate reasoning as code instead of natural language are better reasoners, i.e.\@, mapping to a symbolic language may offer better predictive performance.

The implicit graphical nature of reasoning tasks has motivated evaluating LLMs on explicit graphical tasks isolated from various confounding complexities that these reasoning tasks often introduce.
{\bf This assumes that the minimized graphical tasks act as a surrogate to the original reasoning tasks and that this isolates aspects that make the original tasks difficult without introducing new difficulties.}.
To this end, many graph benchmarks and datasets have recently been introduced using synthetic data  \citep{wang2023can, liu2023evaluating, fatemi2024talk, luo2024graphinstruct, 10.1145/3637528.3672010, dai2024revisiting, dai2024large, fan-etal-2024-nphardeval} and real-world data \citep{guo2023gpt4graph, wang-etal-2024-instructgraph, zhang-etal-2024-llm-graph, DBLP:journals/corr/abs-2406-16176, yuan2024gracore, li2024glbench, tang2025grapharena}. \citet{tang2025grapharena, fan-etal-2024-nphardeval} group the task by difficulty according to its complexity class (which relates to expressibility, Appx.\@ \ref{appx:expressivity}).

LLMs struggle to solve graphical tasks \citep{wang2023can, liu2023evaluating, fatemi2024talk, DBLP:journals/corr/abs-2402-07140, guo2023gpt4graph, dai2024revisiting, perozzi2024let, tang2025grapharena}. \citet{zhang-etal-2024-llm-graph} showed poor performance on out-of-domain tasks and that performance on synthetic data does not generalize to real-world data.   

Various things have been attributed to this poor performance. \citet{fatemi2024talk} demonstrated that the way the graph is encoded in natural language for the LLM has a large impact on performance. \citet{DBLP:journals/corr/abs-2402-07140} found that this can be alleviated by pre-processing using some determinist ordering such as depth- or breadth-first-search. \citet{yuan2024gracore} found similar results with a random ordering of the graphs and showed that sorting can help. (i.e.\@, that order matters, Appx.\@ \ref{appx:order-matters})

\citet{dai2024revisiting} showed how task difficulty does not just scale with graph size but also the topology of graphs being evaluated. {\bf The path-star task is a powerful example of this, where the type of graph makes it very difficult even at small sizes, however, this isn't an inherent property of the topology but a pathological relation between topology and training method.}  They also found that LLMs may apply different algorithms to various tasks and that this is sensitive to input, indicating that the LLM may be using shortcuts.  Other works have also identified spurious correlations as an issue \citep[][see Appx.\@ \ref{appx:spurious}]{wang2023can}.   \citet{fatemi2024talk} evaluated the performance of LLMs on various graph tasks on star-shaped graphs.  They found that a) the topology of graph strongly affects performance, and, b) LLMs generally do better on star-shape graphs than other types of graphs. 
Hallucinations have also been found to be an issue that relates to model scale and graph scale \citep{tang2025grapharena}.

Various methods have been proposed to improve graphical reasoning: graph-specific zero-shot CoT prompts \citep[][described in Appx.\@ \ref{appx:LLMs}]{wang2023can}, alternative algorithmic prompts \citep{dai2024revisiting}, self-prompting \citep{guo2023gpt4graph}, soft-prompting \citep{perozzi2024let}, instruction-tuning \citep{10.1145/3637528.3672010, wang-etal-2024-instructgraph} and instruction-tuning in conjunction with masking \citep[][see Appx.\@ \ref{appx:masking}]{luo2024graphinstruct}, preference alignment, \citep{zhang-etal-2024-llm-graph, wang-etal-2024-instructgraph, 10.1145/3637528.3672010}, and re-framing the task as code for code-aware LLMs \citep{zhang-etal-2024-llm-graph, DBLP:journals/corr/abs-2406-16176}, which has been shown to help for other reasoning tasks \citep{madaan-etal-2022-language}.

Another proposed method is to modify the underlying neural architecture by incorporating graph neural nets into the LLM \citep{4700287, 10.1145/3626772.3657775, chai2023graphllm, wu2024can, ren2024survey, jin2024large}. {\bf Given adulterated supervision, the CHC prevents learning about multi-edge relations which require considering more than two nodes at once. This is partially caused by the attention mechanism of the transformer which is limited to pair-wise iterations.  Thus modifications that consider triplet interactions may also be useful for graphical tasks \citep{hussain2024triplet}.}  

As we use synthetic data, we consider this in more detail.
\citet{wang2023can} introduced the NLGraph benchmark which contains 8 graph-based tasks with 29,370 examples, partitioned into three difficulties. They stated that they `employ a general-purpose random graph generator to generate base graphs while using the number of nodes and graph destiny to control for complexity'.\footnote{This process was Erdős–Rényi \citep{fatemi2024talk}.}  {\bf Random graph construction is complex} and one generation process may lack diversity.  As such, \citet{fatemi2024talk} used seven generation process, including Erdős–Rényi \citep{erdds1959random}, scale-free networks \citep{doi:10.1126/science.286.5439.509}, Barab{\'a}si-Albert \citep{albert2002statistical}, and stochastic block model \citep{holland1983stochastic}, and star-shaped graphs. 
{\bf Random tree construction is also complex and we do a poor job of generating trees that would better support the task.  However, we believe this is best left to future work which considers search on general graph structures.}

Out-of-domain evaluation has also been considered. 
\citet{luo2024graphinstruct} introduced GraphInstruct, which contains 21 graph-based tasks with 4 tasks being reserved as out-of-domain tasks that are not included in fine-tuning.  Each in-domain task has 800 training examples. 
They used three different graph generation processes. 
\citet{zhang-etal-2024-llm-graph} introduced NLGIFT, which included out-of-domain testing.  This includes an experimental setup for fine-tuning on synthetic data and testing on real-world data.  They used two different graph generation processes for the syntactic data.  It has been shown that graph construction has a large impact on learnability \citep[][see Appx.\@ \ref{appx:graph-learnability}]{saparov2025transformers}.



{\bf We believe our work has several implications for graph benchmarks of LLMs.}
These works and ours have different goals and hence different research questions; these they are asking {\em`how well do pretrained LLMs perform on a suit of graphical tasks?'} and then often with the secondary questions {\em `why do they struggle to perform well?'} and {\em `how can we improve performance post-hoc?'}, whereas we are asking {\em `why is learning graphical tasks hard?'}.\footnote{This is under the assumption that the path-star task is a minimal example of search on graphs, however, as we found, task-specific issues contribute to its difficulty.}  The former concerns performance while the latter is a question of learnability.  From these stems the question: {\em can the poor performance of LLMs on graphs be attributable to the same difficulties that hinder learning the path-star task?}  {\bf As mentioned above, we believe that adulteration will have a small impact on LLMs.  However, the issues we present will become more applicable as people move to pretraining LLMs for reasoning tasks.  
These issues may also affect finetuning.  Thus our work motivates the careful design of graph tasks when training or finetuning models.}  We leave it to future work to see if our methods can be used to improve the performance of LLMs.    

Because these works concern evaluating LLMs, they used small datasets.  \citet{frydenlund-2024-mystery} found that randomly sampled graphs can easily lead to spurious correlations due to the size of the sample space.  We solved this using an online dataset.  However, such a solution will be less useful for LLMs which are generally not trained on multiple epochs.  Regardless of this, {\bf we strongly urge the move to online datasets, which, for synthetic datasets, should be as easy as exposing the original data generation process. This should be done during both training and evaluation, where data contamination is and will become a bigger issue for evaluating LLMs \citep{zhu2024dyval}}.

\subsection{Learnability of Graphs}\label{appx:graph-learnability}


 Unlike the above works evaluating LLMs, we are concerned with the learnability of graph algorithms on decoder-only transformer language models. \citet{saparov2025transformers} is the most closely related work (outside of \citet{bachmann2024the}). They consider finding the shortest path given a graph.  As with our experimental setup, they provide a query with a start and end node, and the graph is encoded as a list of shuffled edges. The graph is also randomly generated and semanticless.

 Their first finding is that graph topology highly affects performance (especially in out-of-domain evaluation across topologies). This was also observed by \citet{dai2024revisiting}.  They find that a `balanced' graph topology works the best.  These are graphs sampled from a generative process which creates a graph with a uniform distribution over the number of `lookaheads' (path length) required to solve the task.  Note that, while we discussed these as `more general graphs' in the main text, they are very closely related to path-star graphs.

As our work was nearly completed before we became aware of their work, we do not do direct comparisons.  There are several differences: 1) Most of their experiments use encoder-only models. They did not evaluate path-star graphs using decoder-only models (only using encoder-only models, as in  \citet{frydenlund-2024-mystery}).  2) They employed a slight architecture modifications that concatenates the token and position embeddings.  3) They used rotary positional embeddings in their decoder-only experiments (again, only on balanced graphs).  4) They used an approximate second-order optimizer, Sophia (where we used Adam).  5) Their best models were also trained for 883M samples (where we used 100M).  We suspect that all of these may contribute to differences in performance.   

{\bf However, even given these differences, we observe similar scalability issues (and these may have increased scientific value as they were observed independently).} We both find that, as graph size increases, trials become less likely to converge, i.e.\@,  successfully learn the task to high sequence accuracy.  We also both find that there is a high variability in this convergence across seeds.
This is (implicitly) shown in our tables where we show that many trials are unsuccessful but are still learn above the baseline (ABB > SR).  \citet{saparov2025transformers} reported these results in their Fig.\@ 6, which shows the fraction of converged seeds on graphs of various sizes.  This shows a less than 20\% convergence rate for balanced graphs when $|V| > 40$.\footnote{Note that the accuracies reported in their Fig.\@ 2 used a best model and not are the average rates over all trials.}

Finally, they also show that using depth-first-search or section-inference scratchpads which explicitly decompose the task into intermediate steps does not solve these scaling issues.  This leads them to conclude that transformers struggle to learn to search over graphs as the size of the graph grows.


\citet{khona2024toward} studied the behavioral difference of 2-layer LM on graph tasks with and without in-context examples in order to explicitly limit the model to only reason via in-context learning.  They demonstrated a performance gap between the two as well as showed that in-context examplars allow for compositional generalization on OOD data but this does not apply to length generalization. 
\citet{cohen2025spectral} demonstrated that 2-layer decoder-only models can learn shortest-path representations on small graphs where the learnt embeddings correlate with the spectral decomposition of the graph.   

\citet{wu2024can} considered if learning graph tasks leads to improved planning abilities.  They put forth a related argument to the one given by \citet{bachmann2024the} that that next-token prediction is potentially problematic for learning planning graph tasks 
due to learning spurious correlations.  We do not fully appreciate the pertinence of their Theorem 2 to support a broader insufficiency claim, which we feel is being implicitly made.  In particular, they assume that the next-token logits are determined by the target and the current node, however, logits given by real models are formed as a function of the entire graph.  Their Example 1.\@ seems to be empirically contradicted by \citet{saparov2025transformers} and our work.  Indeed, when the CHC causes the logits to become a function of only the current node, they converge to be $1/D$. 
As far as we can tell, there is no empirical investigation into LM's performance being impeded by these specific conjectured spurious correlations.

\subsection{Spurious Correlations and Shortcuts}\label{appx:spurious}

Spurious correlations in LLMs generally concern OOD performance along with related topics like adversarial attacks and fairness \citep{geirhos2020shortcut, du2023shortcut, song2024shortcut, zhou-etal-2024-navigating-shortcut, steinmann2024navigating}.  
\citet{steinmann2024navigating} provided a literature review and taxonomy of shortcut learning where they define a shortcut as `when a model used a spurious correlation as the basis for its decision making'.  They also considered why models learn shortcuts and considered that one reason is that `a model's task is generally not precisely defined' while citing \citet{bachmann2024the} (and hence commenting on the path-star task).  They then followed this with `The broad task definitions do not specify how the task should be solved, thus enabling the model to rely on shortcuts rather than relevant features'.  {\bf This statement is consistent with our description that the original task setup supports learning two different tasks: the desired path-star task and the undesired edge-following task.  
What is also interesting about the path-star task is that the features used in learning the shortcut are not irrelevant features but rather relevant features used in the wrong way.}  

\citet{wang2023can} showed that spurious correlations affect the performance of LLMs on graph tasks.  In particular, they design two special types of graphs; a `chain' which is just a very long path and a `clique' which has a high edge density.  They found that LLMs fail to solve a connectivity task at high rates on these graphs compared to other general graphs.  This implies that the underlying algorithm is not learnt (or being applied consistently across different graph types) and thus a shortcut is being employed. \citet{jiang-etal-2024-peek, mirzadeh2025gsmsymbolic} showed similar results for reasoning tasks where they argue that the model is learning in-domain spurious correlations and thus only learning a superficial pattern matching instead of true reasoning.  \citet{press-etal-2023-measuring} showed that LLMs can often correctly solve multi-hop subtasks without getting the overall or final answer correct, which they attribute to fact memorization which can be considered as a spurious correlation or undesired shortcut.

Addition is a surprisingly hard task for LMs due to the left-to-right ordering of next-token prediction not matching the order of addition carry-overs, thus requiring that models plan $n$-digits ahead.  
\citet{baeumel2025lookahead} showed how LLMs use a single-lookahead shortcut to perform integer addition (for three-digit numbers).  They demonstrated that this shortcut works well -- but not perfectly -- for two operands, but fails as the number of operands increases.  \citet{Lin2025implicit} showed that LLMs use shortcuts for implicit math reasoning and that, while these work well in-domain, they often fail to solve out-of-domain reasoning tasks.  

\citet{liu2023transformers} demonstrated that automata on sequenced of length $T$ can be simulated with transformers of $\log(T)$-depth via algorithmic `shortcuts' and that these are not robust to OOD data (so being true shortcuts in the above sense).  

{\bf The path-star task is unique in that the induced shortcut failure is in-domain where the shortcut actually absorbs supervision and so prevents learning the primary task instead of just compromising performance OOD.}   
\citet{frydenlund-2024-mystery} identified spurious correlation in the original experimental set-up of \citet{bachmann2024the}.  This was partially resolved with structured samples.  We fully resolve the issue by using an online dataset.  We believe that the learnt shortcuts induced by the path-star task are not shortcuts that will appear in natural language -- or at least affect the task so potentially as they do symbolic tasks \citep{tu-etal-2020-empirical, zhou-etal-2024-navigating-shortcut}.

\subsection{Masking}\label{appx:masking}

Masking is often done to avoid spurious correlations and overfitting.
Masks can be crafted or structured depending on the task via inductive biases that mask specifically linked tokens.  \citet{deng-etal-2021-reasonbert} used constructed query-evidence data pairs and a masked spanning objective that masks parts of the query that are supported by evidence, thus inducing the model to learn a connection between the evidence and the query.  Span selection is also needed in other ways of supervised training of reasoning tasks \citep{stacey-etal-2022-logical}. \citet{rabe2021mathematical} used masking for math reasoning by masking specific sub-expressions. 
This used an inductive bias which masks all occurrences of such sub-expression (thus the mask is structured with the task).  \citet{chen-etal-2024-masked} showed that masking tokens within the CoT improved their effectiveness for fine-tuning and that the placement of the masking is important. 

\citet{luo2024graphinstruct} used masking over the fine-tuning instructions for graph-based tasks.  These were selected by choosing `unimportant' words and, hence, employed an inductive bias for selecting the masks.  Given that it only masked unimportant words, we suspect this did not mask graph information and so would not prevent adulteration.     



\subsection{Expressivity and Learnability}\label{appx:expressivity}

Various works have considered the computational limits or expressivity of transformers,  i.e.\@, `can a transformer actually solve this problem' (given a particular capacity in terms of hidden-state size or number of layers etc.) \citep{Yun2020Are}.  Various computational models are used to prove expressibility, such as formal logic \citep{merrill2023a}, formal languages \citep{hao-etal-2022-formal, 10.1162/tacl_a_00663}, massively parallel computation \citep{pmlr-v235-sanford24a}, or declarative programming languages such as RASP (Restricted Access Sequence Programming)\citep{weiss2021thinking}.  

\citet{weiss2021thinking} demonstrated that RASP programs upper-bound the difficulty/complexity of a task (for a transformer) in terms of the number of required layers (and attention heads) required to solve the task.  It employs a limited computational model of transformers that are restricted to performing uniform attention over a subset of queries (i.e., average-hard attention \citep{10.1162/tacl_a_00663}).\footnote{See \citet{yang2024simulating} who consider when soft attention can simulate various kinds of hard attention.}  While this excludes RASP's use on numerical tasks, it does make it easy to model symbolic tasks such as path-star.  \citet{zhou2024what} extended RASP to causal attention and conjecture that short RASP programs lead to length-generalizability.\footnote{They also wrote RASP in Numpy, making it an easy tool for NLP/ML practitioners.}  \citet{huang2025a} formalized this conjecture, showing why certain problems have poor length generalization while also showing that a certain class of tasks have guaranteed length generalization. \citet{strobl2024transformers} extended RASP to model transformers as transducers, which requires accounting for non-length preserving transitions.  RASP programs can be compiled into actual transformers and the reverse \citep{NEURIPS2023_995f693b, NEURIPS2023_771155ab}. 

Transformer can not learn distributions for next-token prediction for some regular and context-sensitive languages and so expressibility does not match the Chomsky hierarchy \citep{10.1162/tacl_a_00663, hu2025between}.
The expressibility of RNNs/state space models and transformers is different \citep{pmlr-v235-sanford24a, bhattamishra2024separations, NEURIPS2023_73bf6924, jelassi2024repeat}. Thus RNNs/Mamba and transformers may not behave the same on the path-star task.

\citet{luca2024simulation} showed that looped transformers 
can express various graph algorithms with a constant number of layers.  They used a modified transformer architecture which allows for encoding a graph as an adjacency matrix, with a special attention mechanism over this matrix.  They made significant note of the need to limit numerical errors through various methods like using hard attention and careful choice of positional embeddings (see Appx. \ref{appx:length-gen}).  
\citet{sanford2024understanding} developed a representational hierarchy of problem classes for transformers on graph problems. {\bf Path-star falls under the `parallelizable tasks' class, in particular, those solvable with logarithmic depth. 
\citet{frydenlund-2024-mystery} showed that transformers can express the path-star task via RASP for encoder- and decoder-only models.} 

Expressibility is not to be confused with learnability, i.e.\@, `can standard learning methods be used to train a transformer to solve this problem' \citep{allen2023physics, deletang2023neural, NEURIPS2023_73bf6924, pmlr-v235-sanford24a}.\footnote{See \citet{svete-cotterell-2024-transformers} and \citet{svete-etal-2024-transformers} for a case study, where the former considered the expressibility of transformers to model $N$-gram language models, and develop various computational models either using $N-1$ layers or $N-1$ attention heads in combination with hard/sparse attention, while the later then considered the learnability of such models.} 

Going back to RASP, \citet{zhou2024what} modified RASP to better model numerical representation and align RASP with empirical results about learnability.  This included only allowing single increment indexing.  \citet{chang2025language} pointed out that transformers fail to count inductively 
and as such, such abilities should not be inherent abilities in the computational model.  This was also an argument stemming from empirical learnability results.  This demonstrates an inherent divide between the models used for expressibility and learnability.  

{\bf Learnability is the core question of this work, i.e.\@, can decoder-only transformers learn the path-star task?} We show this empirically as well as provide theoretical explanations for why adulteration or lack of decomposition causes the task to be unlearnable. 

\citet{luca2024simulation} also considered a small number of learnability experiments using the CLRS dataset. Here they train on 16 node graphs and evaluate on 64 node graphs as a form of length generalization (see Appx. \ref{appx:length-gen}).  They highlighted how learnability is much more difficult than expressibility where `despite demonstrating the existence of parameters capable of [graph] simulation, discovering them through gradient-based training is challenging.' 

\subsection{Sensitivity}\label{appx:sensitivity}

A specific and highly relevant case of transformer expressivity and learnability is for parity due to it being a (maximally) sensitive function \citep{hahn2021sensitivity, bhattamishra-etal-2023-simplicity, hahn2024sensitive}.\footnote{Again, see \citet{hu2025learning}, for a connection between parity and the path-star task.}  The sensitivity of a discrete function on an input sequence $x$ describes the number of disjoint subsets of $x$ which, when changed, cause changes to the output.  Thus functions with low sensitivity contain redundant information across $x$,  whereas functions with high sensitivity have tokens that isolate important information. The path-star task completely changes its output based on a single target token provided in the query.  Another view of sensitivity is as an analog to the smoothness of continuous functions, where path-star is not smooth with respect to a change in target. 

\citet{chiang-cholak-2022-overcoming} showed that small model details (layer normalization) can have a big impact on the empirical results of learning sensitive functions. \citet{hahn2024sensitive} described the interaction of cross-entropy training with transformers on sensitive functions and found that these transformers inhabit only a small volume of parameter space. \citet{vasudeva2025transformers} considered the sensitivity of non-boolean functions and found that lower sensitivity correlates with better robustness and flatter minima in the loss landscape.  

Sensitivity issues can also appear in more complex NLP tasks \citep{hahn2021sensitivity, chen-etal-2023-relation, chakraborty-etal-2023-zero, lu-etal-2024-prompts, vasudeva2025transformers} as well as reasoning tasks, where small changes to the task input can cause large variances in reasoning abilities \citep{pmlr-v202-shi23a, jiang-etal-2024-peek, mirzadeh2025gsmsymbolic}.\footnote{Often sensitivity is not formally defined in these tasks compared to parity.  This is due to the inherent difficulty of formal definitions of sensitivity for complex tasks.}
Such sensitivity issues can often be attributed to learning spurious correlations or shortcuts.

\subsection {Length Generalization, Task Decomposition, and Scratchpads}\label{appx:length-gen} 

The effect of task decomposition on learnability has been studied \citep{wies2023subtask, dziri2023faith, abbe2024how}.  This is often studied in the context of length generalization.  This is a specific kind of OOD generalization of great importance to LMs due to their sequential nature  \citep{anil2022exploring, zhou2024what, zhou2024transformers}.  Length generalization relates to reasoning tasks that scale to the number of required reasoning steps (hops) \citep{dziri2023faith, abbe2024generalization, xiao2024theory, xiao2025generalizing, mirzadeh2025gsmsymbolic}.  One of the main concerns about the difficulty for transformers to learn parity is that, when they do learn the task, this does not generalize to unseen sequence lengths \citep{bhattamishra-etal-2020-ability, hahn2024sensitive}.  This betrays the fact that the underlying algorithm has not been learnt by the model, i.e.\@, 
``the failure in length generalization corroborates the models’ fundamental limitation that they may not genuinely understand the task solving algorithm but may rely on short-cut learning that is only applicable to sequences of trained length'' \citep{cho2024arithmetic}. In addition to parity, integer addition is often used as a test-bed for length generalization \citep{mcleish2024transformers, cho2024arithmetic}. 
\citet{chang2025language} showed that transformers show poor OOD performance for the simple task of counting (including task variants).   \citet{deletang2023neural} showed similar results for a series of more challenging tasks based on formal languages grouped within the Chomsky hierarchy.  

Naively applying LMs to these tasks results in poor length generalization.  This motivated the use of scratchpads \citep{nye2022show} which are necessary to (efficiently) solve parity with transformers \citep{wies2023subtask, hahn2024sensitive, abbe2024how, kim2025transformers}. \citet{wies2023subtask} showed that there is a large gap in learnability between RNN models that are given scratchpad supervision and those that are not.  This class of tasks includes parity.  \citet{kim2025transformers} established similar results for transformer models.   They are also necessary for arithmetic \citep{NEURIPS2023_4e85362c, cho2024arithmetic}.  Scratchpads/CoT are also used for other reasoning tasks \citep{shah2024causal}.

{\bf The reason why scratchpads are critical is because they provide extra computation via autoregressive generation {\em in conjunction with extra training supervision}, allowing tasks to be decomposed into subtasks via intermediate supervision} \citep{wies2023subtask}.\footnote{Note that the extra computation increases expressibility \citep{feng2023towards, merrill2024the, li2024chain}, while the intermediate supervision increases learnability.}  This can also be framed as a simplification of next-token prediction tasks \citep{zhou2024what}. \citet{dziri2023faith} examined the ability of LLMs to decompose tasks into subtasks via framing tasks as computational graphs.  This allowed them to quantize task difficulty/complexity. They showed that OOD performance is poor, attributing such behaviour to learnt shortcuts and that, while scratchpads help, they may not for highly difficult tasks.  \citet{abbe2024how} conjectured that (set-sized) transformers can weakly solve problems that only require `local' information (a small subset of input tokens) but that a `locality barrier' exists which prevents solving problems that require global information.  They then showed how to overcome this via scratchpads.  


{\bf Like the path-star task, integer arithmetic is a simple task with a simple underlying algorithm that LMs fail on.  Also, a trivial reverse solution exists.} 
\citet{zhou2024transformers} shows that the reverse solution to arithmetic is more robust in terms of length generalization.  {\bf Note that both these reverse solutions work because they do not need to think multiple steps ahead.  For the path-star task, this betrays a lack of reasoning, however, for addition, this conforms with how humans perform arithmetic and thus feels like a more valid solution despite not requiring multi-step reasoning.}  The more interesting question is if models can learn to find the trivial order (which we find does not happen for the BoW experiments). See Appx.\@ \ref{appx:order-matters} for order considerations.

Scratchpads require supervised targets.  To avoid this, {\em thinking tokens} have been introduced which are special tokens inserted as input at specified times without any corresponding targets \citep{herel2024thinking, goyal2024think}.  This increases the available sequential computation -- and hence expressibility.  Note this also may affect learnability just via the ability to learn different functions which require additional computation.  \citet{pfau2024lets} showed that increasing the expressive computation capability using thinking tokens does not mean it is easy to learn to use this capacity. 

\citet{yin-etal-2024-semformer} tried thinking tokens for the path-star problem to negative results.\footnote{Note these are done under the original settings that allow for spurious correlations and overfitting, and thus may have failed due to other reasons.}  This is interesting because using $M$ thinking tokens can, in theory,  provide a trivial solution by computing the reverse arm with the thinking token and then the forward arm from the reverse solution (similar to the BoW experiments, just with even less supervision). {\bf We conjecture that thinking tokens do not work for the path-star task as they do not provide additional decomposition supervision.}\footnote{However, these experiments need to be redone for confirmation since they did not use an online dataset.}

\subsection{Positional Embeddings}

With length generalization comes the need to have positional embeddings that allow for exact matching across long lengths and generalize to unseen positions \citep{kiyono-etal-2021-shape, NEURIPS2023_4e85362c, ruoss-etal-2023-randomized, li2024functional, mcleish2024transformers}. \citet{chang2025language} showed that different embedding types generalize differently to different counting tasks.  {\bf Such methods will be important considerations for graph-based tasks when scaling up the size of graphs and considering length generalization.}  This can also lead to some unexpected results like using no positional embeddings (NoPE) being possible for decoder-only models \citep{irie19b_interspeech, tsai-etal-2019-transformer, haviv-etal-2022-transformer, chi-etal-2023-latent, NEURIPS2023_4e85362c, wang-etal-2024-length, irie2024positional, zuo-etal-2025-position} and can lead to better length generalization for symbolic reasoning tasks \citep{NEURIPS2023_4e85362c}. 
However, \citet{wang-etal-2024-length} showed that NoPE fails to generalize due to a collapse in the attention head distribution as the context size increases.  \citet{yang2025rope} followed this up with a hybrid strategy that combines NoPE, for its strong token retrieval and RoPE for its inductive biases.  \citet{frydenlund-2024-mystery} found that the choice of positional embedding mattered for the path-star task and that NoPE worked when using decoder-only models.

Modifying how the task is represented can improve the behaviour of the positional embeddings.  For example, \citet{mcleish2024transformers, cho2024position, cho2024arithmetic} coupled or reused positional embeddings at similar positions for both operands for the task of addition, leading to better length generalization.  This is an example of symbolic tasks being brittle to how the task is represented and requiring a strong task-specific inductive bias to overcome.

\subsection{Order Matters and Reversal Curse}\label{appx:order-matters}

Prior work has explored the impact of ordering source-side information in LLMs for premise order on the complex tasks of reasoning \citep{wang-etal-2023-towards, chenpremise, allen-zhu2024physics, shah2024causal} and proof generation \citep{an2024next}. \citet{liu-etal-2024-lost} has shown that LLMs struggle to retrieve relevant information in long contexts when that information is placed in the middle of the context compared to either the front or back. {\bf \citet{frydenlund-2024-mystery} showed that order matters for the path-star where they considered that the query should proceed the graph.}
 
Order matters on the target-side as well, since the path-star task becomes trivial when asked to generate the arm in reverse order.  An asymmetry in LM predictive abilities coined the {\em reversal curse} is a recent but well-studied phenomenon \citep{berglund2024the, lin2024delving}.  An example of this is for an LM to be able to predict `A is B' but not `B is A'. A common proposed solution is bidirectional training, incorporating bidirectional information, or a bidirectional model modification \citep{ma2023untying, golovneva2024reverse, lv-etal-2024-analysis, guo-etal-2024-mitigating, guo-etal-2024-exploring}.  This is also the underlying idea of the Belief-State Transformer introduced by \citet{hu2025learning} for solving the path-star task. 

\citet{papadopoulos2024arrows} discovered an asymmetry in perplexity between models trained either in the forward or reverse direction, with the forward having consistently lower perplexity.  This is surprising given that both directions give a valid and theoretically equivalent factorization of the sequential probability.

\citet{chen2023positional, fang2025rethinking} consider order invariance for few-shot in-context learning.  The issue here is that the order of the exemplars should not matter.  This requires considering how the attention or model is parameterized as well as the positional embeddings used. \citet{fang2025rethinking} also considered that fully observed question-answer pairs lead to data leakage and shortcuts.  {\bf This `leakage' can be framed as adulterated supervision.}  
{\bf Order invariance is also very important for graphs represented as lists of edges since the order of this list should not matter.  However, it will matter with decoder-only models due to the causal constraint.}

\subsection{Non-AR, Iterative-AR, and Discrete Diffusion Models}\label{appx:iar}

Given the perceived belief that left-to-right autoregressive models were incapable of solving the path-star task, a natural conclusion would be to use non-autoregressive models (NAR)  \citep{gu2018non, gu-kong-2021-fully} or iterative autoregressive models (IAR) \citep{lee-etal-2018-deterministic, ghazvininejad-etal-2019-mask}.  There are two core aspects of NAR/IAR models.  The first is that they use an any-order model parameterization which forgoes enforcing the causal constraint (achieved by not employing a causal mask in the attention mechanism).  The second is that these are trained using a masking loss (MLM) \citep{devlin-etal-2019-bert}.  In the NAR case, the targets are fully masked.  This means each target token is modeled independently (at the classification layer) and all tokens are decoded in a single step during inference.  This can lead to poor performance, motivating the use of IAR models trained using partial masks, thus allowing for partial dependencies. This allows for multiple decoding steps during inference.  Both these aspects come together to allow the model to generate in any-order.

\citet{bachmann2024the} used a `teacher-less' model which masks out all input tokens. \citet{frydenlund-2024-mystery} connected this model to NAR models and also showed that the path-star task was solvable via an encoder-only model with NAR and IAR training. This was based on a modified version of the CMLM model (where the conditional `C' part of the model is removed) \citep{ghazvininejad-etal-2019-mask}.  \citet{frydenlund-2024-mystery} incorrectly implied that the original `teacher-less' model was non-causal when considering it as a NAR model.  The model described by \citet{monea2023pass} is meant to modify an autoregressive model post-hoc and thus is designed to keep the causal constant.  However, in terms of independently modeling and predicting multiple tokens it behaves exactly like a NAR model (i.e.\@, loss, training, and inference procedure).  

{\bf As we know the reverse `solution' works, the any-order aspect of the NAR/IAR models potentially allows these models to learn the reverse solution without direct supervision. Our results show that the masking operation allows for task decomposition for the path-star task.  We expect that this is the more important aspect of these model's successes over the model's parametrization, however, we also believe that parameterizations will matter due to the causal constraint making graph reconstruction more difficult.}

The connection between IAR models and discrete diffusion models was described by \citet{austin2021structured}.
\citet{kitouni2024the} introduced an `MLM-$U$' diffusion model which uses a uniform masking rate and applied it to the path-star task.  They wrote `this approach can be implemented as a denoising process which recovers randomly masked tokens, like BERT, but with uniformly sampled masking rates. This key difference allows training a generative model with masked modeling.'  This key insight was first described by \citet{ghazvininejad-etal-2019-mask} with their IAR CMLM model.  As mentioned, CLMC was used by \citet{frydenlund-2024-mystery}, however, the path-star experiments in \citet{kitouni2024the} were not described in enough detail to do a comparison with \citet{frydenlund-2024-mystery}. 

Different masking strategies have been used; \citet{lee-etal-2018-deterministic} used token replacement from $V$ while \citet{ghazvininejad-etal-2019-mask} used a special masked token.  Other works have explored any-order LM parameterization outside of the NAR/IAR/diffusion framework \citep{yang2019xlnet, liao-etal-2020-probabilistically}

\subsection{ Future Token Prediction}\label{appx:future}

Early works in future prediction designed models which could predict $N$ tokens into the future by creating $N$ separate hidden-states and training on each state with cross-entropy against a single future token  \citep{goodman-etal-2020-teaforn, qi-etal-2020-prophetnet}.  Thus these are not truly belief-states directly, however, a belief-state must be present in the model in order to generate the $N$ separate hidden-states.  The general goal of this was for improved training by explicitly learning to predict future tokens and hence plan for future tokens, and was not for multi-token inference.   \citet{heo2024n} also used multi-state prediction for future n-grams but also introduced a method to explicitly create representations that are compositional into the future. \citet{gloeckle2024better} proposed an efficient training method for $N$ token prediction which repurposed $N$ multi-head attention to create the $N$ hidden-states.\footnote{\citet{qian-etal-2025-beyond} considered the use of  multi-token prediction to alleviate prompt sensitivity for LLMs.  Interestingly, they mentioned that the four future tokens used by \citet{gloeckle2024better} are insufficient for their task.  {\bf This highlights a potential benefit of RITF, which may be able to scale to more distant future tokens since this is decoupled from any model parameterization.}} While they did consider how to use this to increase inference speed, their main goal was to demonstrate that training with multi-token prediction increases performance on downstream tasks even if this ability is removed during inference.  Because of the success of training with multi-token prediction, it has as been adopted into foundational models such as DeepSeek3 \citep{liu2024deepseek}, which used sequential prediction method which requires introducing $N$ multiple parameterized modules for $N$ future tokens.  

\citet{gerontopoulos2025multi} also introduced a method for training a model with multi-token prediction which includes extra {\em register} tokens into the input sequence during training.  These extra tokens induce creating extra hidden-states from which future tokens can be predicted.  Thus, instead of modifying the model's architecture by re-purposing heads to create extra hidden-state, this simply modifies the input sequence.  Since they only use this for training, they also prevent these register tokens from being attended to by regular tokens, which would result in a inference-time bias.  They showed the effectiveness of this method compared to \citet{gloeckle2024better}.\footnote{They also show their method works on the path-star task using a pretrained GPT2 model.}  {\bf Their method is similar to our RITF method, except whereas they make future predictions from extra hidden-states created by modifying the input to the model, we instead reuse hidden-states for multiple future predictions via using a structured loss function. This helps highlight the connection between inputs and targets for multi-token prediction.}

\citet{cai2024medusa} expanded on the multi-head method for multi-token prediction by incorporating it with speculative decoding faster inference \citep{xia2022speculative, chen2023accelerating, leviathan2023fast}.\footnote{As an aside, speculative decoding was also the original purpose of the 'teacher-less' model \citep{monea2023pass} and, as with NAR/IAR, the main motivating factor for speculative decoding is improved inference speed.}.  The parameterized modules of DeepSeek3 can also be retrained during inference and used for speculative decoding \citep{liu2024deepseek}.  

\citet{pal-etal-2023-future} studied the extent to which the hidden-states of LMs contain predictive information about future tokens and hence act as belief states.  They used {\em lens} to show that the hidden-states of models trained solely to predict the next token contain enough to predict up to three tokens into the future between 20-40\% of the type (where, the type of lens and prompting method used had a large effect on the predictive ability \citep{hewitt-manning-2019-structural, nostalgebraist, belrose2023eliciting, yom-din-etal-2024-jump}).
\citet{men-etal-2024-unlocking} also investigates the existence of belief-states in LLMs specifically for planning tasks.  

\citet{wu2024do}  studied the mechanism for LMs to learn future information from a next-token prediction objective. 
They hypothesized that it could be due to two mechanisms; a deliberate {\em pre-cashing} mechanism which computes features earlier than they are needed and an unintentional {\em breadcrumb} mechanism which considers that a LM learns features for predicting the next token and that these just happen to also be good features for predicting future tokens also.  They construct a synthetic dataset and show that pre-cashing is done and necessary for some planning tasks.  However, they also show that pre-cashing is less noticeable in a GPT2 model used for natural language (but also consider this might be less true as LMs scale up). Notably this mechanism will not help for the path-star task as future predictions can ignore both pre-cashing and breadcrumb features due to the CHC. That is, any features used for planning will just be ignored.  This also means that there will be no learning signal to reinforce learning such features.  

{\bf We design future distributions and associated losses to enhance this ability for the path-star task.   
Not only do these create an explicit belief-state that allows for planning, they also avoid adulteration by targeting tokens that require multiple edges or path-reconstruction to predict.}
RITF is also designed to be efficient during train as it only requires a single parallelizable loss.  This is in contrast to other losses on $N$ tokens into the future, which scale linearly with $N$  \citep{goodman-etal-2020-teaforn, qi-etal-2020-prophetnet, gloeckle2024better}.









\end{document}